\documentclass[letterpaper,10pt]{article}

\usepackage[style=numeric-comp, sorting=none, sortcites=true, backend=bibtex, maxbibnames=10, minbibnames=1]{biblatex}
\usepackage{pythonhighlight}
\usepackage[most]{tcolorbox}
\usepackage{float}
\usepackage{siunitx}
\newcommand{\fone}[1]{\num[round-mode=places,round-precision=1]{#1}}
\newcommand{\dlt}[2]{%
  \edef\dv{\fpeval{round(#2-#1,1)}}%
  \ifnum\fpeval{\dv>0}=1
    \textcolor{green!60!black}{+\fone{\dv}}%
  \else\ifnum\fpeval{\dv<0}=1
    \textcolor{red!70!black}{\fone{\dv}}%
  \else
    \fone{\dv}%
  \fi\fi
}

\providecommand{\respm}[2]{\ensuremath{#1 \pm #2}}

\newcommand{\resEasyPassBaseMean}{65.5}

\newcommand{\resEasyPassBase}{\respm{65.5}{3.2}}
\newcommand{\resEasyPassRLMean}{71.5}

\newcommand{\resEasyPassRL}{\respm{71.5}{2.4}}
\newcommand{\resEasyPassSFTMean}{79.9}

\newcommand{\resEasyPassSFT}{\respm{79.9}{2.3}}
\newcommand{\resEasyMeanKBaseMean}{39.4}

\newcommand{\resEasyMeanKBase}{\respm{39.4}{1.8}}
\newcommand{\resEasyMeanKRLMean}{54.1}

\newcommand{\resEasyMeanKRL}{\respm{54.1}{1.4}}
\newcommand{\resEasyMeanKSFTMean}{58.4}

\newcommand{\resEasyMeanKSFT}{\respm{58.4}{1.7}}

\newcommand{\resMediumPassBaseMean}{49.0}

\newcommand{\resMediumPassBase}{\respm{49.0}{3.4}}
\newcommand{\resMediumPassRLMean}{61.8}

\newcommand{\resMediumPassRL}{\respm{61.8}{2.6}}
\newcommand{\resMediumPassSFTMean}{67.1}

\newcommand{\resMediumPassSFT}{\respm{67.1}{3.1}}
\newcommand{\resMediumMeanKBaseMean}{26.7}

\newcommand{\resMediumMeanKBase}{\respm{26.7}{1.5}}
\newcommand{\resMediumMeanKRLMean}{43.2}

\newcommand{\resMediumMeanKRL}{\respm{43.2}{1.4}}
\newcommand{\resMediumMeanKSFTMean}{45.4}

\newcommand{\resMediumMeanKSFT}{\respm{45.4}{1.6}}

\newcommand{\resHardPassBaseMean}{1.1}

\newcommand{\resHardPassBase}{\respm{1.1}{0.9}}
\newcommand{\resHardPassRLMean}{3.6}

\newcommand{\resHardPassRL}{\respm{3.6}{1.3}}
\newcommand{\resHardPassSFTMean}{1.6}

\newcommand{\resHardPassSFT}{\respm{1.6}{1.0}}
\newcommand{\resHardMeanKBaseMean}{0.3}

\newcommand{\resHardMeanKBase}{\respm{0.3}{0.2}}
\newcommand{\resHardMeanKRLMean}{1.2}

\newcommand{\resHardMeanKRL}{\respm{1.2}{0.5}}
\newcommand{\resHardMeanKSFTMean}{0.5}

\newcommand{\resHardMeanKSFT}{\respm{0.5}{0.3}}

\newcommand{\resTPBenchPassBaseMean}{40.3}

\newcommand{\resTPBenchPassBase}{\respm{40.3}{2.8}}
\newcommand{\resTPBenchPassRLMean}{41.1}

\newcommand{\resTPBenchPassRL}{\respm{41.1}{2.2}}
\newcommand{\resTPBenchPassSFTMean}{40.9}

\newcommand{\resTPBenchPassSFT}{\respm{40.9}{2.3}}
\newcommand{\resTPBenchMeanKBaseMean}{25.7}

\newcommand{\resTPBenchMeanKBase}{\respm{25.7}{1.6}}
\newcommand{\resTPBenchMeanKRLMean}{31.4}

\newcommand{\resTPBenchMeanKRL}{\respm{31.4}{1.4}}
\newcommand{\resTPBenchMeanKSFTMean}{29.0}

\newcommand{\resTPBenchMeanKSFT}{\respm{29.0}{1.5}}

\newcommand{\resTPBenchLOneMeanKBase}{\respm{61.8}{6.2}}

\newcommand{\resTPBenchLTwoMeanKBase}{\respm{67.2}{5.0}}

\newcommand{\resTPBenchLThreeMeanKBase}{\respm{6.3}{3.0}}

\newcommand{\resTPBenchLFourMeanKBase}{\respm{2.0}{1.6}}

\newcommand{\resTPBenchLFiveMeanKBase}{\respm{0.0}{0.0}}

\newcommand{\resArXivPassBaseMean}{33.0}

\newcommand{\resArXivPassBase}{\respm{33.0}{1.3}}
\newcommand{\resArXivPassRLMean}{39.8}

\newcommand{\resArXivPassRL}{\respm{39.8}{1.3}}
\newcommand{\resArXivPassSFTMean}{32.2}

\newcommand{\resArXivPassSFT}{\respm{32.2}{1.2}}
\newcommand{\resArXivMeanKBaseMean}{19.1}

\newcommand{\resArXivMeanKBase}{\respm{19.1}{0.6}}
\newcommand{\resArXivMeanKRLMean}{25.0}

\newcommand{\resArXivMeanKRL}{\respm{25.0}{0.6}}
\newcommand{\resArXivMeanKSFTMean}{19.5}

\newcommand{\resArXivMeanKSFT}{\respm{19.5}{0.6}}

\newcommand{\resArXivHepThMeanKBase}{\respm{15.7}{1.0}}

\newcommand{\resArXivHepPhMeanKBase}{\respm{11.0}{1.8}}

\newcommand{\resArXivGrQcMeanKBase}{\respm{26.3}{1.3}}

\newcommand{\resArXivMathPhMeanKBase}{\respm{15.4}{2.7}}

\newcommand{\resArXivQuantPhMeanKBase}{\respm{25.8}{2.0}}

\newcommand{\resArXivClassPhMeanKBase}{\respm{19.6}{2.3}}

\newcommand{\resArXivOtherMeanKBase}{\respm{14.5}{1.7}}

\newcommand{\resQFTPassBaseMean}{21.1}

\newcommand{\resQFTPassBase}{\respm{21.1}{1.1}}
\newcommand{\resQFTPassRLMean}{29.5}

\newcommand{\resQFTPassRL}{\respm{29.5}{1.2}}
\newcommand{\resQFTPassSFTMean}{22.7}

\newcommand{\resQFTPassSFT}{\respm{22.7}{1.1}}
\newcommand{\resQFTMeanKBaseMean}{8.8}

\newcommand{\resQFTMeanKBase}{\respm{8.8}{0.4}}
\newcommand{\resQFTMeanKRLMean}{14.4}

\newcommand{\resQFTMeanKRL}{\respm{14.4}{0.5}}
\newcommand{\resQFTMeanKSFTMean}{10.1}

\newcommand{\resQFTMeanKSFT}{\respm{10.1}{0.5}}

\newcommand{\resQFTTextbooksMeanKBase}{\respm{8.3}{0.7}}

\newcommand{\resQFTExerciseBooksMeanKBase}{\respm{9.4}{0.6}}

\newcommand{\resQFTMITOCWMeanKBase}{\respm{7.2}{1.9}}

\usepackage{tabularx}
\usepackage{booktabs}
\usepackage{enumitem}
\usepackage{csquotes}
\tcbuselibrary{breakable, skins}
\usepackage{authblk}
\usepackage{caption}
\usepackage{amsmath,amssymb,slashed} 
\usepackage{graphicx} 
\usepackage[margin=1in,letterpaper]{geometry} 
\usepackage{dirtytalk}
\tcbuselibrary{breakable}
\usepackage{cancel}
\usepackage{multirow}
\usepackage{listings}
\usepackage{pgfplots}
\usepackage{tikz}
\usepackage{subcaption}
\usepackage{setspace}
\usepackage{xcolor}
\usepackage{hyperref}
\usepackage{fontawesome5}
\usepackage{svg}  

\definecolor{navydarkblue}{RGB}{10, 10, 120}

\newcommand{\hficon}{\raisebox{-0.15em}{\includegraphics[height=1em]{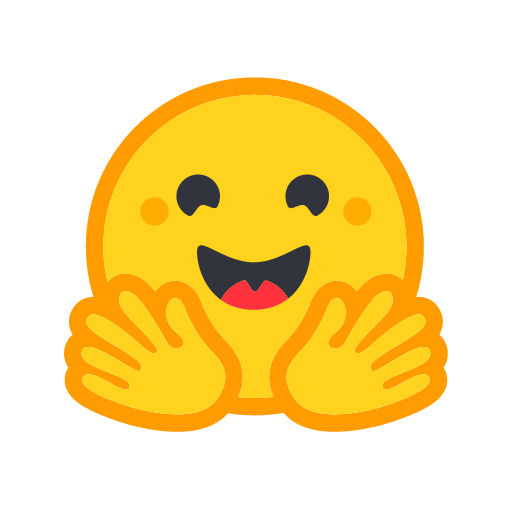}}}

\newcommand{\giticon}{\raisebox{-0.15em}{\includegraphics[height=1em]{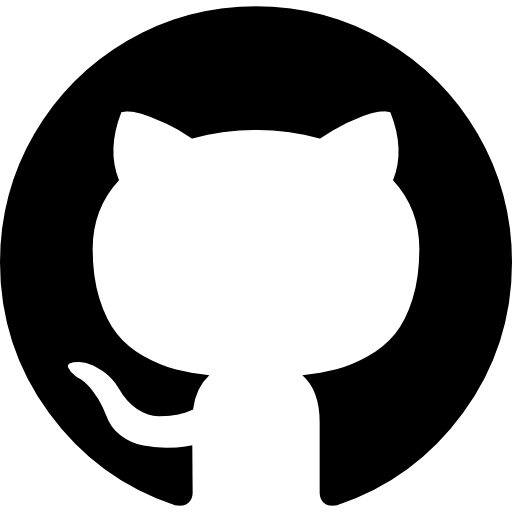}}}

\usepackage{cleveref}
\usepackage{mdframed}

\usetikzlibrary{decorations.pathmorphing}
\tikzset{snake it/.style={decorate, decoration=snake}}

\hypersetup{
	colorlinks=true,       
	linkcolor=blue,        
	citecolor=blue,        
	filecolor=magenta,     
	urlcolor=blue         
}

\newtcolorbox{problem}[1][]{
    enhanced,
    colback=blue!5,         
    colframe=blue!75!black, 
    fonttitle=\bfseries,
    title=Problem,          
    attach boxed title to top left={yshift=-2mm, xshift=2mm},
    boxed title style={colback=blue!75!black},
    sharp corners=shortarc,
    breakable,              
    #1                      
}

\definecolor{emerald}{rgb}{0.31, 0.78, 0.47}

\newtcolorbox{solution}[1][]{
    enhanced,
    colback=emerald!5!white, 
    colframe=emerald!60!black,
    fonttitle=\bfseries,
    title=Solution,
    attach boxed title to top left={yshift=-2mm, xshift=2mm},
    boxed title style={colback=emerald!60!black},
    sharp corners=shortarc,
    breakable,
    borderline west={2pt}{0pt}{emerald!60!black}, 
    #1
}

\definecolor{standardcolor}{RGB}{255, 243, 205}  
\definecolor{attemptcolor}{RGB}{240, 240, 240}  
\definecolor{problemcolor}{RGB}{230, 245, 255}  

\definecolor{usercolor}{RGB}{33, 150, 243} 
\definecolor{modelcolor}{RGB}{76, 175, 80} 
\definecolor{expertcolor}{RGB}{220, 20, 60} 

\tcbuselibrary{breakable} 

\tcbset{dialogue/.style={
    colback=gray!10,
    colframe=gray!40,
    boxrule=0.3mm,
    arc=3mm,
    before skip=10pt,
    after skip=10pt,
    breakable,
    fonttitle=\bfseries,
    coltitle=black,
    colbacktitle=gray!20,
    attach title to upper,
}}

\title{Fine-Tuning Small Reasoning Models for Quantum Field Theory}

\author[1]{Nathaniel S. Woodward} 
\author[2]{Zhiqi Gao} 
\author[1]{Yurii Kvasiuk} 
\author[3]{Kendrick M. Smith}
\author[2]{Frederic Sala} 
\author[1]{Moritz M\"unchmeyer}

\affil[1]{Department of Physics, University of Wisconsin-Madison}
\affil[2]{Department of Computer Science, University of Wisconsin-Madison}
\affil[3]{Perimeter Institute for Theoretical Physics}

\begin{document} 
\maketitle
\begin{center}
    \vspace{-0.4cm} 
    \begin{tabular}{cl}
        \hficon & \url{https://huggingface.co/datasets/nswoodward/VerifiableQFT} \\[0.8ex] 
        \giticon & \url{https://github.com/nswood/VerifiableTPData}
    \end{tabular}
    
    \vspace{0.4cm} 
\end{center}

\begin{abstract}
Despite the growing application of large language models (LLMs) to theoretical physics, there has been little academic exploration of how domain-specific physics reasoning ability develops during training. To investigate this, we perform the first academic comparison of reinforcement learning and supervised fine-tuning for LLMs on verifiable problem-solving tasks in theoretical physics. Because the open-source verifiable training data required to develop such capabilities is scarce, we built a robust data generation pipeline that can both create synthetic problems and make existing human-authored problems suitable for model training. 
Selecting quantum field theory (QFT) as our primary domain, we generated over 2,500 synthetic problems alongside a curated collection of human-adapted problems sourced from arXiv and standard pedagogical resources. 
We conduct both reinforcement learning (RL) and supervised fine-tuning (SFT) experiments, benchmarking performance gains as well as generalization to other physics domains.
Although both methods improve over the base model on every dataset, SFT gains are concentrated on in-distribution synthetic tasks, while RL generalizes better, outperforming SFT on the human-adapted datasets and TPBench.
We perform an extensive analysis of model chain-of-thought (CoT) reasoning before and after fine-tuning to understand how reasoning errors evolve during RL and SFT. Finally, we publicly release our data pipeline, verifiable QFT training data, and \(\sim\)200M tokens of QFT reasoning traces.
\end{abstract}

\begin{figure}[H]
    \centering
    \vspace{-1em}
    \includegraphics[width=1.0\linewidth]{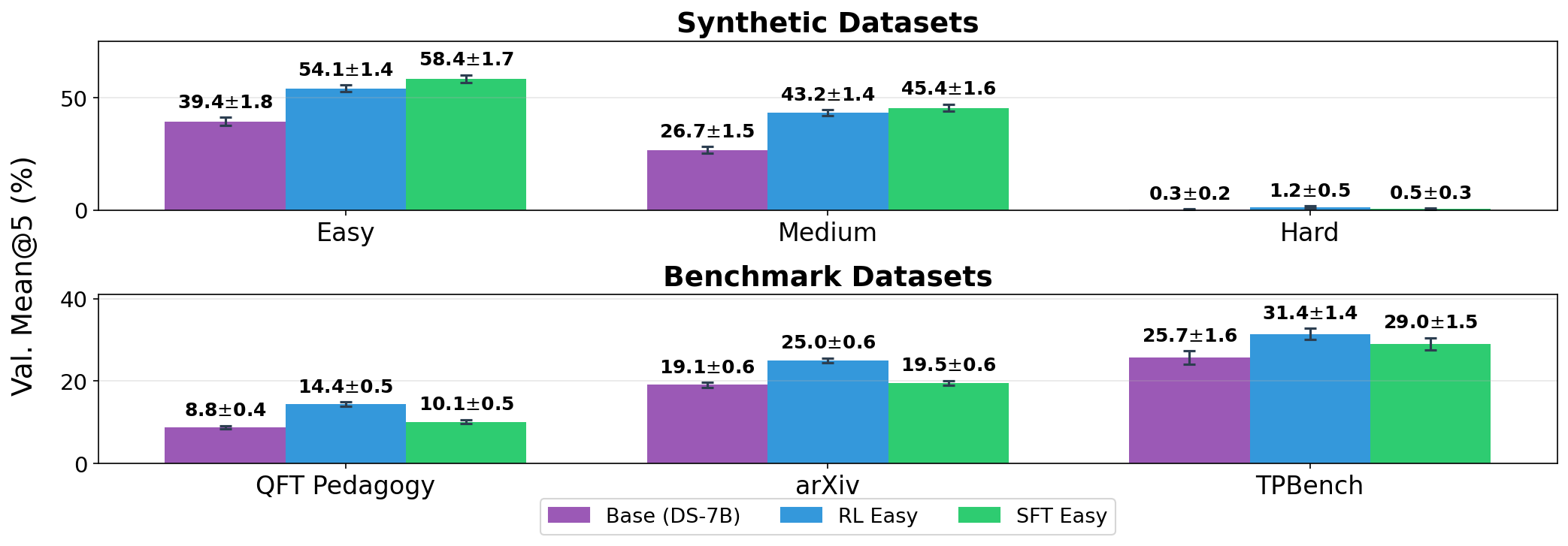}
    \caption{Performance gains after fine-tuning \texttt{DeepSeek-R1-Distill-Qwen-7B} with RL on QFT Easy, compared to SFT on rejection-sampled \texttt{Qwen3-30B-A3B} reasoning traces for the same dataset.}
    \label{fig:rl_sft_teaser}
\end{figure}
\begin{center}
\normalsize
    Corresponding Author: Nathaniel S. Woodward (\texttt{nwoodward2@wisc.edu})
\end{center}

\clearpage
\tableofcontents
\clearpage

\section{Introduction}

Large language models have recently become highly capable at the mathematical reasoning required in theoretical physics \cite{lewkowycz2022solvingquantitativereasoningproblems,taylor2022galacticalargelanguagemodel,glazer2024frontiermathbenchmarkevaluatingadvanced,chung2025theoreticalphysicsbenchmarktpbench,feng2025physicsbenchmarkingfoundationmodels,zhang2025physreasoncomprehensivebenchmarkphysicsbased,pang2024physicsreasonerknowledgeaugmentedreasoning,xu2025ugphysicscomprehensivebenchmarkundergraduate,Tschisgale_2025,Pan_2025,Cai__2025,Barman_2025,guevara2026singleminusgluontreeamplitudes}. It is therefore interesting to study the training dynamics and internal reasoning processes that give rise to these capabilities. Unfortunately, while the general principles of model training, such as supervised fine-tuning (SFT) and reinforcement learning with verifiable rewards (RLVR), are publicly known, their implementation requires industrial-scale compute to achieve state-of-the-art results. Most industry labs currently do not publish their training data, their precise training setup, or intermediate results of their training processes (such as the rollouts of RL generations). Therefore, to study LLM learning and reasoning dynamics within academic compute constraints, smaller-scale experiments in the target domain are needed. While recent work has begun to fine-tune small LLMs for high-energy theory \cite{Richmond_2026,Cai__2025}, to our knowledge the present work is the first comparison of reinforcement learning and supervised fine-tuning for LLMs on verifiable problem-solving tasks in theoretical physics (specifically quantum field theory).

Quantum field theory (QFT) provides an ideal domain for this exploration due to three complementary features. First, as a foundation of modern theoretical physics, QFT has a wealth of online pedagogical resources, which mitigates data curation challenges. Second, pedagogical QFT calculations are notoriously tedious yet analytically tractable without computer algebra tools---an aspect we do not explore in this work. Third, QFT follows a well-defined pedagogical progression (from classical field theory and scalar fields to fermions and non-Abelian gauge theories), providing a clear gradient of topic complexity.

\begin{tcolorbox}[colback=blue!5!white, colframe=blue!50!black, title=\textbf{Our Novel Contributions:}, arc=4pt, boxrule=0.5pt, left=6pt, right=6pt, top=4pt, bottom=4pt]
    \begin{enumerate}
        \item \textbf{Synthetic Data Pipeline:} A synthetic data generation pipeline for auto-verifiable QFT problems with independently variable operational and domain
        difficulty.

        \item \textbf{QFT Training Dataset:} The public release of thousands of verifiable QFT problems spanning a range of reasoning difficulties, enabling further studies of domain-specific fine-tuning.

        \item \textbf{Domain-Specific RL \& SFT Comparison:} The first comparison of RL and SFT for LLMs on verifiable problem-solving tasks in theoretical physics, showing that both improve over the base model but differ in the distribution of those gains: SFT concentrates in-distribution, while RL transfers to human-adapted problems and TPBench. We further analyze the learning dynamics of each method and characterize how fine-tuning reduces common errors in the model's chain-of-thought.

    \end{enumerate}
\end{tcolorbox}

We begin in \Cref{sec:datasets} by detailing our data pipeline for verifiable problems, which both creates fully synthetic problems and adapts human-authored problems into our verifiable format.
In \Cref{sec:train_methods}, we provide a brief summary of the learning mechanisms of supervised fine-tuning (SFT) and reinforcement learning with verifiable rewards (RLVR). We perform RL and SFT training runs for \texttt{DeepSeek-R1-Distill-Qwen-7B} and compare their performance gains in \Cref{sec:exp1} and \Cref{sec:exp2}. We explore the dynamics of RL fine-tuning on narrow domains in Appendix \ref{sec:exp3}. Finally, we perform a detailed chain-of-thought (CoT) analysis of the reasoning traces in \Cref{sec:error_analysis}, comparing the base model to the RL and SFT models to identify shifts in reasoning behavior.
In summary, our work provides a foundation for studying the learning process of LLMs in theoretical physics under academic constraints.

\subsection{Notation}
\label{sec:notation}
In this work, we employ typical notation used throughout LLM research to report model performance. 

\begin{itemize}
    \item $\textbf{k}$: The total number of independent stochastic generations (or attempts) sampled from the model for a given problem.
    \item $\textbf{mean@k}$: The empirical mean pass rate over $k$ independent attempts. This is an estimator of $\text{pass}@1$--- the expected accuracy of a single generation--- and is the quantity commonly reported as $\text{pass}@1$ elsewhere in the literature.
    \item $\textbf{pass@k}$: The probability of obtaining at least one correct solution within the pool of $k$ attempts. This metric evaluates the model's exploratory reasoning capabilities and is functionally equivalent to the Best-of-$k$ performance.
\end{itemize}

\section{Related Work}

\paragraph{LLMs in Mathematics and Science.}
The application of large language models (LLMs) to scientific domains has progressed from generalist models to specialized models with deep technical knowledge and reasoning abilities. Minerva \cite{lewkowycz2022solvingquantitativereasoningproblems} demonstrated that training on scientific corpora containing \LaTeX{} significantly improves performance on STEM benchmarks. By fine-tuning PaLM on arXiv papers and web pages, Minerva achieved state-of-the-art results on the MATH dataset, employing majority voting to mitigate calculation errors. Similarly, Galactica \cite{taylor2022galacticalargelanguagemodel} was trained on a massive corpus of scientific knowledge to act as an interface for science, using specialized tokens for citations and reasoning steps, though it faced challenges with hallucination.

Beyond these general-purpose scientific models, dedicated mathematical LLMs such as Llemma \cite{azerbayev2024llemmaopenlanguagemodel} have shown that continued pretraining on mathematical corpora yields strong performance on formal and informal reasoning tasks. 
Industry-scale endeavors such as AlphaGeometry \cite{trinh2024alphageometry} and the subsequent AlphaProof/AlphaGeometry2 systems \cite{AlphaProofandAlphaGeometryTEams_2024} have been highly successful, reaching silver-medal performance at the International Mathematical Olympiad. More recent model-agnostic verification-and-refinement pipelines have pushed frontier LLMs to gold-medal IMO 2025 performance \cite{huang2025winninggoldimo2025}. Formal theorem proving has also been extended from pure mathematics toward physics: Ax-Prover \cite{breen2025axproverdeepreasoningagentic} introduces a Lean-based agentic framework with a dedicated benchmark of quantum-physics theorems. While formalization through Lean is a promising direction, it is still in active development, and critical re-evaluations of standard Lean benchmarks \cite{ospanov2025minif2fleanrevisitedreviewinglimitations} have highlighted persistent limitations in measuring genuine mathematical reasoning.

Despite these advancements, a gap remains between solving standardized benchmark problems and performing the open-ended reasoning required for scientific research. Benchmarks such as FrontierMath \cite{glazer2024frontiermathbenchmarkevaluatingadvanced}, TPBench \cite{chung2025theoreticalphysicsbenchmarktpbench}, PHYSICS \cite{feng2025physicsbenchmarkingfoundationmodels}, CritPt \cite{zhu2025probingcriticalpointcritpt}, and PhysReason \cite{zhang2025physreasoncomprehensivebenchmarkphysicsbased} highlight that models struggle with the long-horizon, multi-step derivations of graduate-level mathematics and theoretical physics problems. A growing wave of physics-specific benchmarks further sharpens this picture: ABench-Physics \cite{zhang2025abenchphysicsbenchmarkingphysicalreasoning} probes robustness under dynamic problem variations, CMT-Benchmark \cite{pan2026cmtbenchmarkbenchmarkcondensedmatter} curates expert-written condensed matter theory problems on which even frontier models solve only a small fraction, and NewtonBench \cite{zheng2026newtonbenchbenchmarkinggeneralizablescientific} targets the discovery of generalizable scientific laws by LLM agents. All of these point to a rapidly narrowing gap on standardized tasks but persistent weaknesses on open-ended derivations.

In theoretical physics specifically, LLMs have been used for multi-step Hartree-Fock derivations \cite{Pan_2025} and symbolic Matsubara summations in QFT via reinforcement learning at criticality \cite{Cai__2025}. Most recently, Guevara et al.\ \cite{guevara2026singleminusgluontreeamplitudes} used a frontier LLM (GPT-5.2 Pro) to conjecture a closed-form expression for single-minus gluon tree amplitudes that was subsequently formally proven, providing a concrete instance of an LLM contributing to a novel result in QFT and scattering amplitudes. This development is complemented by general-purpose autonomous research agents such as Aletheia \cite{feng2026autonomousmathematicsresearch}, which couples Gemini Deep Think with a generator/verifier/reviser loop to explore the automated generation of mathematics research papers, and The AI Scientist-v2 \cite{yamada2025aiscientistv2workshoplevelautomated}, which uses agentic tree search to attempt the end-to-end construction of workshop-level scientific papers. While fully independent research agents have not yet reached a professional human level, taken together these results suggest that the frontier is shifting from benchmark-style evaluation toward LLMs as active collaborators in mathematical and physical research.

Closest to our setting, FeynTune \cite{Richmond_2026} fine-tunes small LLMs on arXiv abstract completion tasks in high-energy theory, showing that domain-specialized fine-tuning of open models is feasible under academic constraints. SCALAR \cite{niarchos2026doescritiqueimproveaiassisted} grades QFT and string-theory reasoning of off-the-shelf models through a structured critic--actor loop. On the evaluation side, recent physics-reasoning benchmarks such as \emph{Towards a Large Physics Benchmark} \cite{barman2025largephysicsbenchmark} and PRL-Bench \cite{miao2026prlbenchcomprehensivebenchmarkevaluating} further probe frontier physics-research ability. Our work is distinct in its focus on \emph{code-verifiable} QFT reasoning tasks and in providing a direct, controlled comparison of RL/GRPO against SFT for small reasoning models.

\paragraph{Synthetic Data Generation and Reasoning Distillation.}
In the modern era of machine learning, synthetic data has evolved from a mere augmentation tool into a fundamental necessity for various tasks, including reasoning, effectively manufacturing the ``textbook-quality'' step-by-step logical traces that are critically scarce in natural corpora but essential for teaching models to think rather than merely pattern-match \cite{goldie2025syntheticdatageneration, liu2025agenticmathenhancingllmreasoning, Liu_Feng_Shen_Liu_Wan_Sun_2025, wu2025sharpsynthesizinghighqualityaligned,huang2024targatargetedsyntheticdata, davidson2025orchestrating, tongyideepresearchteam2025tongyideepresearchtechnicalreport, xu2025toucansynthesizing15mtoolagentic}. The ``Textbooks Are All You Need'' hypothesis, proposed by the Phi series \cite{gunasekar2023textbooksneed, li2023textbooksneediiphi15}, posits that training on ``textbook-quality'' synthetic data allows significantly smaller models to rival larger counterparts. This data-centric approach focuses on generating clear, instructional examples that make the reasoning process explicit. 

This paradigm extends to reasoning distillation, where student models learn from the reasoning traces of stronger teacher models via supervised fine-tuning (SFT). The foundational idea of knowledge distillation \cite{hinton2015distillingknowledgeneuralnetwork} has been adapted to the LLM setting: Orca \cite{mukherjee2023orcaprogressivelearningcomplex} showed that training a 13B model on explanation traces from GPT-4 enables it to surpass larger instruction-tuned baselines, and Distilling Step-by-Step \cite{hsieh2023distillingstepbystepoutperforminglarger} demonstrated that a 770M model trained on LLM-generated rationales can outperform the 540B PaLM teacher, while MiniLLM \cite{gu2026minillmonpolicydistillationlarge} introduced on-policy distillation to reduce the train-test distribution mismatch inherent in standard SFT. DeepSeek-R1 \cite{Guo_2025} demonstrated that by distilling long reasoning traces from its RL-trained model into smaller architectures via SFT, the distilled Qwen2.5-14B model outperformed QwQ-32B-Preview across reasoning benchmarks.

Subsequent analyses have investigated the key factors governing successful CoT distillation, finding that weaker student models benefit from coarser-grained reasoning while stronger students can leverage finer-grained traces \cite{chen2025unveilingkeyfactorsdistilling}, and that careful selection of teacher data based on ``local naturalness'' to the student is critical \cite{just2025distillingreasoningstudentllms}. From a mechanistic perspective, representational analyses reveal that distilled models develop unique reasoning feature directions whose structural richness correlates with distillation performance \cite{baek2025understandingdistilledreasoningmodels}.

\paragraph{Reinforcement Learning for Mathematical Reasoning.} A central design choice in RL for reasoning is the source of the reward signal. Outcome Reward Models (ORMs) supervise only the final answer, while Process Reward Models (PRMs) \cite{lightman2023letsverifystepstep, uesato2022solvingmathwordproblems} shift supervision to intermediate steps, mitigating the ``false positive'' problem in which correct answers are derived from incorrect logic. Reinforcement learning with verifiable rewards (RLVR) sidesteps learned reward models entirely by using deterministic verifiers, as demonstrated by approaches like SHARP \cite{wu2025sharpsynthesizinghighqualityaligned} and SWiRL \cite{goldie2025syntheticdatageneration}, which synthesize high-quality reasoning trajectories through verifiable rewards and stepwise optimization.

Orthogonal to the reward source is the choice of RL training algorithm. DeepSeekMath \cite{shao2024deepseekmathpushinglimitsmathematical} introduced Group Relative Policy Optimization (GRPO), which eliminates the need for a critic model by leveraging group-based relative rewards, improving training stability and efficiency over standard PPO. Furthermore, rStar-Math \cite{guan2025rstarmathsmallllmsmaster} employs Monte Carlo Tree Search (MCTS) to generate verified reasoning paths, enabling small language models to achieve frontier-level performance through self-evolution. Recent analyses of RL training dynamics suggest that reasoning capabilities emerge hierarchically, shifting from low-level procedural execution to high-level strategic planning \cite{wang2025emergenthierarchicalreasoningllms}, and that RL fundamentally reshapes how models approach reasoning---cultivating internal strategies and knowledge integration rather than merely boosting accuracy \cite{wang2025accuracydissectingmathematicalreasoning}.

\paragraph{Curriculum Learning in Language Models.}
Curriculum learning---training on data organized from easy to hard---is essential for efficient convergence in reasoning tasks. Reasoning Steps as Curriculum \cite{jung2025reasoningstepscurriculumusing} proposes using ``Depth of Thought'' (DoT), measured by the number of reasoning steps in a teacher's trace, as a difficulty signal. This method aligns training curricula with cognitive load, training models on shallow reasoning before progressing to deep, multi-step deductions. Adaptive curricula that re-estimate difficulty during training have also been proposed to address the dynamic nature of model competence \cite{tang2025importancetaskcomplexityevaluating}.

\section{Dataset Curation and Methodology}
\label{sec:datasets}

Theoretical physics (TP) entails a spectrum of reasoning tasks, ranging from rigorous derivations rooted in pure mathematics to phenomenological predictions designed to match experimental data. 
Many of these tasks are proof-based or derivation-based.
Currently, there is a large effort among reasoning researchers to develop training methods for non-verifiable tasks (using, among others, reward models \cite{zhang2025lessonsdevelopingprocessreward,wang2024mathshepherdverifyreinforcellms,yue2025promotingefficientreasoningverifiable,liu2025acemathadvancingfrontiermath} and weak verifiers \cite{saadfalcon2025shrinkinggenerationverificationgapweak,stroebl2026limitsinferencescalingresampling}). While this direction is exciting, most of the progress thus far in mathematical reasoning has been driven by training with outcome verification \cite{shao2024deepseekmathpushinglimitsmathematical,deepseekai2025deepseekr1incentivizingreasoningcapability,qwen3technicalreport,1062014}. We therefore design our dataset and experiments around verifiable tasks in QFT.

For our synthetic QFT problems, we distinguish between \emph{domain difficulty} and the \emph{operational difficulty} of a task. We define these as:
\begin{itemize}
    \item \textbf{Domain Difficulty:} the depth of background knowledge required to comprehend a given topic. Tasks with high domain difficulty often involve pedagogically advanced concepts, significant abstraction, and interdisciplinary methods.
    \item \textbf{Operational Difficulty:} the purely mechanical and logical demands of solving a task, independent of the topic. Assuming perfect understanding of the domain, this isolates the difficulty of the actual execution, such as navigating constraints and performing multi-step reasoning. Tasks with high operational difficulty typically demand non-trivial solution paths, complex mathematical operations, and long derivations.
\end{itemize}

In the following sections, we describe the curation of three training datasets for QFT reasoning and two evaluation sets.  
All datasets rely on fully code-verifiable data (\Cref{sec:python_verif}, \Cref{sec:verifiable_tasks}) and adhere to a unified problem-solution schema (\Cref{sec:python_verif}). The generation pipeline begins with \textit{problem seeds}---classified as either synthetic or human-adapted (\Cref{sec:problem_seeds})---which undergo generation and strict quality filtering (\Cref{sec:data_generation}) to produce the final datasets (\Cref{sec:curated-datasets}).

We construct three training datasets of varying operational difficulty, denoted Easy, Medium, and Hard QFT. All three are generated from a fixed topic list spanning domain difficulties from advanced undergraduate to postgraduate.
By this design, each synthetic dataset has a fixed operational difficulty and varying domain difficulty.
Easy and Medium QFT consist of self-contained, single-task problems with guiding context, specifically designed for models with lower reasoning ability. Hard QFT contains multi-step problems with minimal context, intended to test model reasoning ability as well as physics knowledge.
For evaluation, we collect human-authored problems from QFT pedagogical resources (textbooks and lecture notes) and from arXiv manuscripts on theoretical physics. Together, these test the model’s ability on pedagogical QFT tasks and its broader physics knowledge.

\subsection{Problem Structuring and Auto-Verification}
\label{sec:python_verif}
In TP calculations, derived expressions can be complex, and there may be many equally correct derivations for a single problem.
To address this, we require the model to implement its final analytic solution within a specified Python function. \Cref{fig:python_skeleton} shows an example Python skeleton for a Medium QFT problem (Problem 1061), which tasks the model with calculating the ratio of self-contracted photon propagators.

To verify model solutions, we employ a suite of test cases that span the physically meaningful range of inputs.
All test cases are selected by frontier LLMs, allowing the complexity of these tests to adapt dynamically to the task structure. For example, a discrete classification task (e.g., identifying a symmetry group) may require only a single assertion, whereas a calculation (e.g., an energy-dependent scattering cross section) requires a sweep of test points to rigorously verify the model's symbolic derivation across the relevant phase space.
This approach automates the construction of code-based verification used in TPBench \cite{chung2025theoreticalphysicsbenchmarktpbench}.
\begin{figure}[t!]
\centering
\small
\setlength{\fboxsep}{6pt}
\begin{tcolorbox}[colback=blue!5!white, colframe=blue!50!black, title=\textbf{Python Code Skeleton} \hfill\textit{Medium QFT, Problem 1061}, arc=4pt, boxrule=0.5pt, left=6pt, right=6pt, top=4pt, bottom=4pt]
\begin{python}
def calculate_propagator_ratio(d: int) -> float:
    """
    Calculates the ratio R(d) of the self-contracted photon 
    propagators in Feynman gauge to Landau gauge in d 
    spacetime dimensions.
    R(d) = (D_{\mu\nu}^{(F)} D^{\mu\nu, (F)}) / 
           (D_{\mu\nu}^{(L)} D^{\mu\nu, (L)})
    Args:
        d (int): The number of spacetime dimensions.
    Returns:
        float: The numerical value of the ratio R(d).
    """
    raise NotImplementedError
\end{python}
\end{tcolorbox}
\caption{A characteristic example of the Python code skeleton provided in a problem statement. The model converts its analytic solution into this function, enabling automated test case verification.}
\label{fig:python_skeleton}
\end{figure}

We decompose a single problem into the following distinct components:

\begin{itemize}
    \item \textbf{Problem Statement}: The presentation of the setup and specific tasks. This section defines all conventions and assumptions to ensure the problem is unambiguous.
    
    \item \textbf{Answer Requirements}: This provides a Python function skeleton (signature) that the model must implement. It specifies necessary library imports (e.g., \texttt{numpy}, \texttt{scipy}), input parameters, and a strict docstring defining the return type (e.g., \texttt{float}, \texttt{complex}).
    
    \item \textbf{Solution and Answer}: The \textit{Solution}, often referred to as the \enquote{golden solution}, details the conceptual reasoning as well as any necessary derivations or calculations. The \textit{Answer} isolates the final result as a concise, explicit constant or evaluable expression.
    
    \item \textbf{Code}: This contains a fully functional Python script---adhering to the signature in the Answer Requirements---that numerically evaluates the analytic result derived in the Solution.
\end{itemize}

\paragraph{Verifier Reliability.}
To evaluate a model, the grader executes both the reference (\enquote{golden}) implementation and the model's generated code. These are run on a small set of physically bounded input tuples---typically five, though parameter-free tasks use just one, and \enquote{Hard} tasks use up to ten. The grader then compares the outputs using an absolute tolerance of $|x_{\mathrm{ref}} - x_{\mathrm{att}}| \le \tau$, where $\tau = 10^{-6}$ by default. This tolerance applies element-wise for arrays and to the modulus for complex values; task-specific overrides are extremely rare, occurring in less than $0.1\%$ of cases.

For categorical outputs, we require an exact match against a small predefined set, while booleans are evaluated strictly as $\{0,1\}$. To ensure that all valid mathematical derivations produce identical outputs, we also standardize arbitrary conventions, such as metric signatures and transform normalizations.

By design, this pipeline makes \emph{false positives} highly improbable. A flawed numerical solution would have to accidentally match the reference value to within $10^{-6}$ across the entire input sweep. While a model could in principle guess a single boolean or categorical label, we have not observed this in practice.

Conversely, false negatives are inherently conservative and only serve to depress the measured accuracy. These typically stem from attempt-side code failures, such as parsing issues, build and runtime errors, or exceeding the $10\,\mathrm{s}$ timeout limit. Verifier-side failures (a reference implementation that itself fails, \Cref{sec:challenges_in_tp_data}, or a grader exception on array-valued outputs) occur in only $91$ of $140{,}058$ graded validation attempts, upper-bounding the verifier-induced false-negative rate at $0.065\%$. Finally, relying on frontier-model consensus during the curation process (\Cref{sec:data_generation}) largely eliminates the risk of unreliable golden solutions.

\subsection{Verifiable Tasks in Theoretical Physics}
\label{sec:verifiable_tasks}

\begin{figure}[t]
    \centering
    \includegraphics[width=\linewidth]{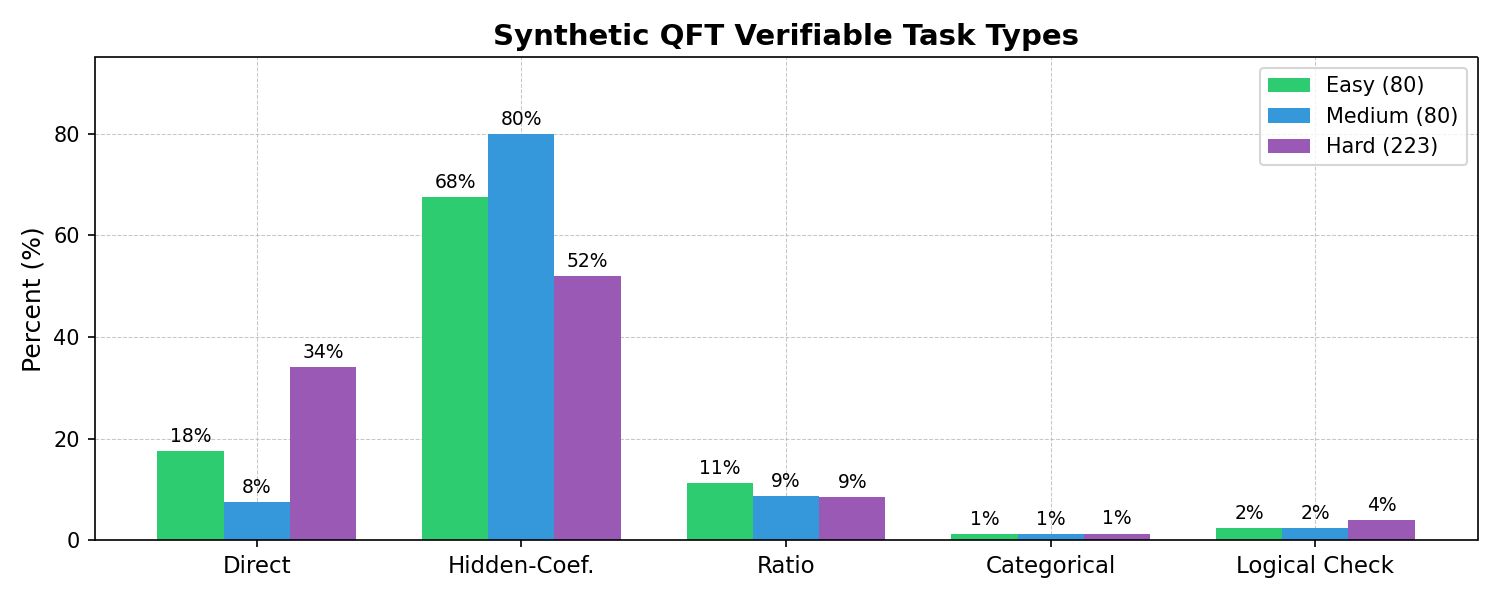}
    \caption{Task distributions across the three synthetic training datasets. We find that hidden-coefficient calculations are the task type most frequently selected by the generating models.}
    \label{fig:task_distribution}
\end{figure}

In our initial efforts to curate a verifiable training dataset, we focused exclusively on tasks yielding numerical predictions. While quantitative calculation is a fundamental component of theoretical physics, relying solely on such problems severely restricts the dataset and fails to cover a broad spectrum of tasks. To address this limitation, we expanded our methodology to include five distinct categories of verifiable tasks:

\begin{enumerate}
    \item \textbf{Direct Calculation Tasks:}
    These tasks require the computation of a quantity that can be directly expressed as a number or a specific numerical function of given parameters.
    \\ \textit{Example:} Evaluating a regulated integral to find a finite constant $C_1(d,n)$.

    \item \textbf{Hidden-Coefficient Derivation:}
    The problem statement hides known analytic results and prompts the model to extract specific constants, coefficients, or exponents that appear in the final expression. The model must perform the full derivation internally but reports only the defining numeric values.
    \\ \textit{Example:} Finding constants $C_R, C_L$ in a Hamiltonian.

    \item \textbf{Ratio and Comparison Tasks:}
    Tasks that require comparing two physically related quantities to output a normalization-independent numeric ratio or boolean comparison.
    \\ \textit{Example:} Determining the relative sign between two scattering amplitudes.

    \item \textbf{Categorical Classification:}
    Tasks requiring the identification of the correct discrete label from a finite, predefined set of physical properties.
    \\ \textit{Example:} Classifying particles (e.g., ``scalar'', ``pseudoscalar'', ``axial-vector''); distinguishing statistics (``bosonic'' vs.\ ``fermionic'').

    \item \textbf{Logical Consistency Checks:}
    Tasks where the model must verify whether a derived quantity satisfies a specific physical condition, resulting in a binary output.
    \\ \textit{Example:} Verifying if a Hamiltonian is Hermitian.
\end{enumerate}
In our data generation pipeline (\Cref{sec:data_generation}), we provide these task descriptions and instruct the model to select a suitable task type for the generated problem. The distribution of tasks in our validation sets is shown in \Cref{fig:task_distribution}. The Easy, Medium, and Hard QFT datasets all consist of 80 problems, but we allow 2--5 tasks per problem (2.8 on average) in the Hard dataset.

\subsection{Problem Seeds}
\label{sec:problem_seeds}
Our data generation and processing pipeline is structured around \textit{problem seeds}, which serve as the inputs for creating problem-solution pairs. We categorize problem seeds into two distinct classes, corresponding to \textbf{human-adapted} and \textbf{synthetic} generation streams. 

\paragraph{Human-adapted Seeds} The first class of seeds consists of unprocessed problem-solution pairs collected from available TP resources. These are genuine, human-created problems that we curate from two domains in this work:
\begin{itemize}
    \item \textbf{QFT Pedagogy:} We utilize a total of $565$~seeds sourced from:
    \begin{itemize}
        \item \textbf{Textbooks:} Zee ($172$ problem seeds) \cite{Zee:2003mt}, Peskin \& Schroeder ($26$ problem seeds) \cite{Peskin:1995ev}, Weinberg ($14$ problem seeds) \cite{Weinberg:1995mt}.
        \item \textbf{Exercise Books:} Radovanovic ($193$ problem seeds) \cite{Radovanovic:2008zz}, Cheng \& Li ($127$ problem seeds) \cite{Cheng:1984vwu}.
        \item \textbf{MIT OpenCourseWare:} QFT I--III ($33$ problem seeds) \cite{liu2023rqft1,liu2010rqft2,wilczek2003rqft3}.
    \end{itemize}
    
    \item \textbf{arXiv:} To incorporate contemporary topics, we scrape problems and solutions from $131$ theoretical physics arXiv manuscripts \cite{0704.3116,0708.2433,0708.4231,0801.3471,0802.0249,0802.1862,0804.2595,0805.0568,0806.3474,0810.0159,0811.0354,0812.4408,0904.3445,0907.5424,1009.1635,1012.2575,1012.3990,1102.0529,1104.0254,1109.3897,1110.2346,1110.4864,1111.1472,1205.6535,1206.4017,1309.4133,1309.4188,1310.6329,1401.3916,1403.2371,1403.6685,1406.5464,1412.6312,1501.06570,1503.05723,1504.03049,1506.01065,1510.02804,1606.09460,1611.06707,1611.09787,1612.07718,1612.08661,1702.04713,1703.02287,1708.09213,1711.08482,1801.01483,1802.03405,1809.01403,1811.08950,1901.05824,1904.10923,1907.09415,1908.10681,1910.11659,1911.03095,2005.07240,2005.08573,2005.10241,2006.03909,2008.00887,2008.10625,2010.09368,2012.06345,2012.11382,2108.06565,2108.09004,2110.05294,2111.02410,2203.04091,2204.02158,2206.05799,2208.02506,2210.12319,2211.05606,2211.06269,2211.13733,2212.08685,2212.13644,2301.00942,2301.09679,2301.10741,2304.08512,2305.12215,2307.07478,2308.09826,2308.09844,2308.14022,2310.07954,2310.08196,2312.16643,2403.20311,2404.03560,2406.04954,2408.09564,2409.07068,2409.09211,2411.03381,2412.08711,2412.11649,2505.06317,2505.08460,2507.11565,2507.15840,2306.05976,astro-ph/0008111,astro-ph/0011294,astro-ph/0611454,astro-ph/0703730,cond-mat/9510014,cond-mat/9512099,cond-mat/9812110,gr-qc/0102083,gr-qc/0204057,gr-qc/0611129,hep-th/0004098,hep-th/0011005,hep-th/0102055,hep-th/0203048,hep-th/0609055,hep-th/0612129,hep-th/0701216,hep-th/9310088,hep-th/9802037,math-ph/0503052,quant-ph/0005003,quant-ph/0203025,quant-ph/0502096,quant-ph/0606100,quant-ph/0606196}, from which we collect $397$ seeds for evaluation\footnote{We consider the following arXiv categories: hep-th, hep-ph, hep-lat, cond-mat, nucl-th, gr-qc, quant-ph, astro-ph, nlin, physics, and math-ph. If a manuscript is labeled with multiple domains, we assign it to the first domain label.}. Our scraping is not exhaustive of arXiv, but it is sufficient for our purposes. We also specifically select pedagogical problems on arXiv, rather than research-level tasks.
\end{itemize}

For QFT pedagogy, problem-solution pairs (seeds) were collected manually. We utilized unofficial online solution sets for textbook problems (e.g., Zhong-Zhi Xianyu's solutions to Peskin and Schroeder \cite{xianyu2016peskin}), alongside official solutions from exercise books and MIT OCW. We retained only the pairs that could be faithfully transformed into our defined verifiable tasks.

Because manual collection from arXiv is inefficient for a constantly evolving database, we developed an agentic framework for seed collection using Gemini. This agent queries the arXiv API, selects relevant manuscripts based on their abstracts, and parses the LaTeX source code to extract problem and solution sections. We then match the extracted problems with their corresponding solutions, again filtering out any that cannot be converted into verifiable tasks.

\paragraph{Synthetic Seeds} A synthetic seed consists of a physics topic and a target operational difficulty.
To ensure a comprehensive curriculum and a varied domain difficulty, we organize our topics into four distinct pedagogical categories: Advanced Undergraduate (AU), Graduate (GR), Advanced Graduate (AG), and Postgraduate (PG). Within each category, we structure topics into several subgroups, listed in \Cref{tab:qft_roadmap}.
\newcommand{\tablelist}[1]{%
    \begin{minipage}[t]{\linewidth}
        \begin{itemize}[leftmargin=*, nosep, after=\vspace{\baselineskip}]
            #1
        \end{itemize}
    \end{minipage}%
}

\begin{table}[h!]
\centering
\caption{Subgroups within each fixed domain complexity level in the synthetic QFT datasets. The full set of topics is provided in \Cref{sec:qft_topics}.}
\label{tab:qft_roadmap}
\small
\begin{tabularx}{\textwidth}{X X}
\toprule
\textbf{ADVANCED UNDERGRADUATE} & \textbf{GRADUATE} \\
\cmidrule(r){1-1} \cmidrule(l){2-2}
\tablelist{
    \item Foundations and Scalar Fields
    \item Path Integrals and Quantization
    \item Symmetries and Scattering
} & 
\tablelist{
    \item Fermions and Spinor Structure
    \item Gauge Fields and Symmetry
    \item Renormalization and Corrections
    \item Spontaneous Symmetry Breaking
    \item Applied Topics and Tools
} \\ 
\addlinespace 
\toprule
\textbf{ADVANCED GRADUATE} & \textbf{POSTGRADUATE} \\
\cmidrule(r){1-1} \cmidrule(l){2-2}
\tablelist{
    \item Renormalization Group and Scaling
    \item Non-Abelian Gauge Theories
    \item Finite Temperature/Vacuum Effects
    \item Anomalies and Supersymmetry
    \item Applied and Conceptual Tools
} & 
\tablelist{
    \item Amplitude and On-Shell Methods
    \item Loop and Regularization Techniques
    \item Effective Field Theories
    \item Nonperturbative/Topological Physics
    \item Curved Spacetime and Semiclassical QFT
    \item Thermal, Lattice, and Nonequilibrium
    \item Conformal and Dual Structures
} \\ \bottomrule
\end{tabularx}
\end{table}
This hierarchical structure is crucial for two reasons: it ensures dataset breadth, and it minimizes data overlap by restricting generation to well-defined subgroups. Within each subgroup, we list several common problem types to further structure problem creation. The complete QFT topic list is provided in Appendix \ref{sec:qft_topics}.

\subsection{Data Generation Pipeline} \label{sec:data_generation}
\begin{figure}[t!]
    \centering
    \includegraphics[width=\linewidth]{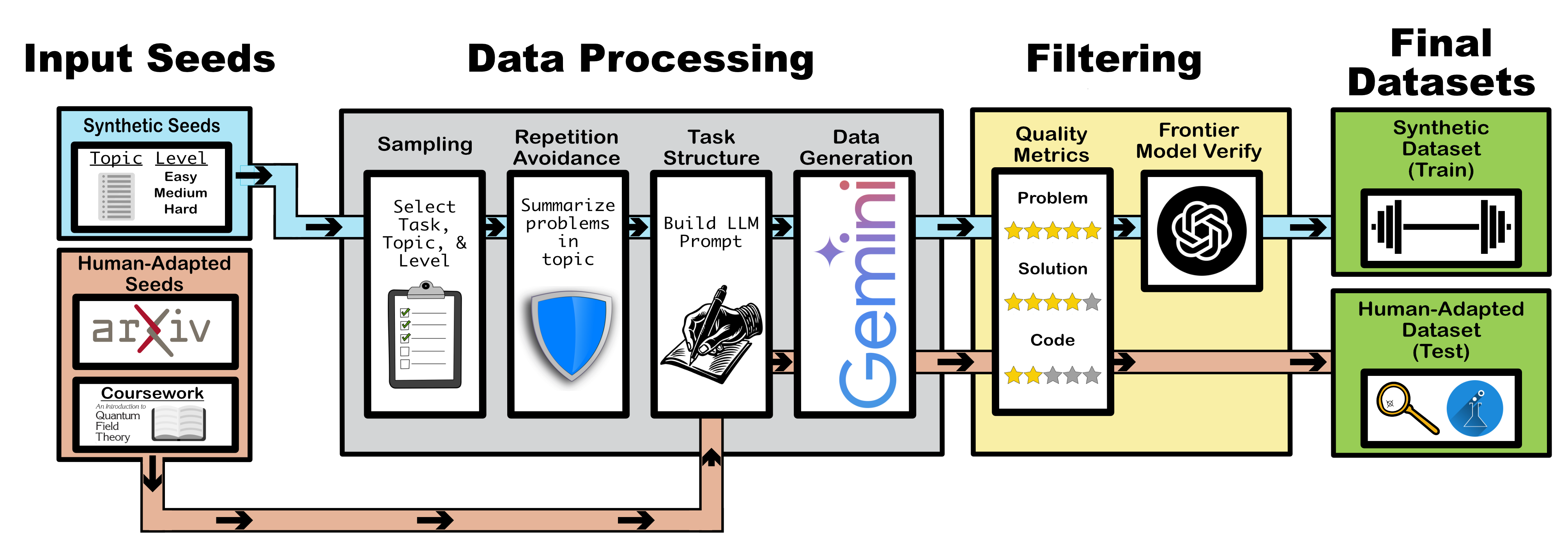}
    \caption{Schematic overview of the data curation pipeline. The workflow is divided into two tracks: a fully synthetic pipeline (blue arrows) utilized for generating the training dataset, and a human-adapted pipeline (orange arrows) derived from arXiv manuscripts and coursework for the testing dataset. Both streams utilize Gemini for generation but differ in their initialization and verification stages, with only the synthetic data undergoing Frontier Model Verification.}
    \label{fig:data_pipeline}
\end{figure}

As illustrated in \Cref{fig:data_pipeline}, our generation pipeline accommodates both synthetic and human-adapted seeds. These tracks serve distinct objectives and entail different validation requirements. Since we cannot assume the validity of fully synthetic problems, that pipeline prioritizes verification of quality and solvability. In contrast, we treat human-authored solutions as inherently valid; consequently, the human-adapted pipeline focuses strictly on adapting existing problems to our schema with minimal modification to the core physics tasks. In the following subsections, we detail the specific mechanisms of this workflow, beginning with our strategies for sampling and repetition avoidance.

\subsubsection*{Sampling \& Repetition Avoidance}
To generate a dataset of synthetic problems, we sample problems uniformly over the four domain levels $(\text{AU}, \text{GR}, \text{AG}, \text{PG})$. 
For a selected domain level, we uniformly sample within problem types from that level. 
To ensure dataset quality, we must minimize \textit{repetition}.
While our hierarchical topic list prevents overlap across broad domains, we must also ensure diversity within individual topics.
To achieve this, we maintain a registry of short, expert-level summaries for every generated problem.
These summaries are generated entirely by the same frontier model that generates the task.
For example, for QFT Easy validation problem 1015 (topic GR-02), the problem summary is:
\begin{quote}
\textbf{Problem Description:} This problem requires the calculation of vacuum expectation values for two different orderings of fermion and anti-fermion creation and annihilation operators. The solution is found by applying the canonical anti-commutation relations to simplify the operator strings.
\end{quote}

During the generation phase, the frontier model is provided with summaries of all previously generated tasks within the target topic via its system prompt. This context allows the model to identify existing coverage and steer generation toward novel problems. Specifically, we include the following section in the system prompt:
\begin{tcolorbox}[
  enhanced,
  breakable,
  colback=black!2,
  colframe=black!40,
  boxrule=0.4pt,
  left=8pt, right=8pt, top=8pt, bottom=8pt,
  width=\textwidth
]
\begin{verbatim}
{System prompt}
Previously generated problems in this topic:

{insert previous descriptions}

IMPORTANT: Generate a problem with tasks that are different from those listed above. 
Some similarity is acceptable, but the problem should not be an exact copy 
or nearly identical to any of the previous problems.
\end{verbatim}
\end{tcolorbox}

During the initial development of our synthetic data pipeline, these summaries were manually checked for accuracy. In deployment, however, the problem summaries are not reviewed. Furthermore, they are not used for any purpose beyond providing context on previously generated problems.

\subsubsection*{Task Structuring \& Data Generation}
To initiate generation, we select either a problem-solution pair (human-adapted) or a topic and difficulty (synthetic).
This seed is integrated into a system prompt (provided in Appendix \ref{sec:prompt}) that describes our task structuring schema, Python implementation standards, calculation conventions, and problem guidelines. 
Each fully constructed prompt is provided to a frontier model, which returns the complete problem-solution pair and the physically meaningful test cases. We generate \texttt{QFT Easy} and \texttt{QFT Medium} with \texttt{Gemini-2.5-Pro},\footnote{These datasets were generated prior to the release of \texttt{Gemini-3-Pro}.} while \texttt{QFT Hard} and all human-adapted datasets are generated with \texttt{Gemini-3-Pro}. We show characteristic example problems in Appendix \ref{sec:ex_problems}.

\subsubsection*{Quality Metrics}
We employ a five-point evaluation protocol to assess each generated problem-solution pair:

\begin{enumerate}
    \item \textbf{Output-Seed Correspondence (Human-Adapted Only):} Measures the correspondence of the generated task to the original source manuscript. While structural modifications necessary for auto-verifiability (e.g., converting symbolic tasks to numeric functions) are permitted, the core physical concepts and problem task must remain consistent with the seed. 
    \item \textbf{Problem Definition:} Assesses whether the problem statement is rigorous, unambiguously defined, and theoretically solvable. The score reflects the clarity of the physical setup and the precision of the answer requirements.
    \item \textbf{Solution Completeness:} Evaluates whether the solution addresses every sub-task and constraint posed in the problem statement. This metric is strictly functional; it measures the presence of answers for all requirements, independent of the depth of the explanation.
    \item \textbf{Explanatory Quality:} Evaluates the pedagogical clarity and physical correctness of the solution text. High scores are reserved for coherent, step-by-step derivations that justify the result.
    \item \textbf{Test Case Quality:} Evaluates whether the test cases adequately verify that the implementation is correct.
\end{enumerate}

\noindent
All problem-solution pairs are scored individually by frontier models on each metric as Very Poor, Poor, Fair, Good, or Excellent.
For \texttt{QFT Easy} and \texttt{QFT Medium}, we use a single quality grading by \texttt{Gemini-2.5-Pro}, while for \texttt{QFT Hard} we use three gradings in total, two from \texttt{GPT-5} and one from \texttt{Gemini-3-Pro}.
For downstream tasks, we retain only those problem-solution pairs that reach the \enquote{Excellent} threshold on every quality grading.
We introduced test-case verification specifically to handle the increased complexity of the Hard and human-adapted sets; it proved unnecessary for the simpler problems.\footnote{If the only failing metric is Test Case Quality, we repair the test cases and pass the problem.}

\subsubsection*{Frontier Model Verification}
While quality filters remove obvious errors, we ensure validity by requiring independent verification from a frontier model (\texttt{GPT-5}). We retain only those problems solved by \texttt{GPT-5} in one attempt (for Easy/Medium) or if at least one of three attempts succeeds (for \texttt{Hard}). 
We use model consensus as a proxy for ground truth, assuming coincident errors are rare. Furthermore, filtering is non-destructive to our goals: if a problem's complexity exceeds the capabilities of a frontier verifier, it is safely assumed to be out of reach for the smaller reasoning models studied here.

\subsubsection*{Generation Pipeline Bias}
Our data generation pipeline uses frontier models for both generation and filtering, a process that inevitably introduces inductive biases. For instance, problems generated by \texttt{Gemini-3-Pro} or filtered by \texttt{GPT-5} are likely to exhibit properties that favor their respective model families during evaluation. Although this would affect benchmarking validity---a use we do not intend for any of these datasets---we expect the impact to be negligible in a training context for models outside these families.

\subsection{Curated Datasets} 
We validate our dataset curation by benchmarking proprietary and open-weight models; the results are discussed in Appendix \ref{sec:baseline_perf}. We find that our synthetic data follows the intended operational difficulty gradient and that the human-adapted problems are challenging for both open-weight and proprietary models. 

\label{sec:curated-datasets}
\subsubsection*{Fully Synthetic Dataset}
Using the QFT topic list detailed in Appendix \ref{sec:qft_topics}, we curate three datasets---Easy, Medium, and Hard QFT---by varying the intended operational difficulty.
The Easy and Medium datasets each comprise \(\sim\)1,000 training problems, whereas Hard is limited to \(\sim\)550 training problems due to higher generation costs. All three datasets are paired with 80 validation problems. For each validation set, we sampled 20 problems from each of the four domain levels. Within each domain level, these 20 problems were distributed equally across its constituent subtopics. Creating these datasets required generating 4,602 synthetic problems in total, which underwent the successive filtering stages shown in \Cref{tab:combined_data_stats}.
\begin{table}[h]
\centering
\caption{\textbf{Data Curation Statistics.} Number of problems retained at each stage of the pipeline.}
\label{tab:combined_data_stats}
\small
\begin{tabular}{l ccc c cc}
\toprule
& \multicolumn{3}{c}{\textbf{Fully Synthetic Pipeline}} & & \multicolumn{2}{c}{\textbf{Human-Adapted Pipeline}} \\
\cmidrule(lr){2-4} \cmidrule(lr){6-7}
\textbf{Filtering Stage} & \textbf{Easy} & \textbf{Medium} & \textbf{Hard} & & \textbf{Pedagogy} & \textbf{arXiv} \\
\midrule
Initial Count (Gen./Seeds) & 1,452 & 1,650 & 1,500 & & 565 & 397 \\
Passed QC & 1,350 & 1,441 & 908 & & 480 & 333 \\
Passed QC + \texttt{GPT-5} & 1,106 & 1,092 & 631 & & NA & NA \\
\bottomrule
\end{tabular}
\end{table}

\subsubsection*{Human-Adapted Datasets}
Using our curated set of human-authored problem seeds, we generate the arXiv and QFT Pedagogy validation sets, which together contain over 800 quality-checked problems. These datasets provide insight into reasoning ability both within QFT and across theoretical physics domains.

\paragraph{Data Generation Cost}
Data generation, including quality metrics, costs approximately \(\$0.25\) per problem. In total we generated \(1{,}452\) Easy, \(1{,}650\) Medium, and \(1{,}500\) Hard synthetic problems, for a total of approximately \(\$1{,}200\). After filtering on quality metrics we retained \(1{,}350\) Easy, \(1{,}441\) Medium, and \(908\) Hard. The quality control failure rate increases with problem difficulty: \(7\%\) for Easy, \(13\%\) for Medium, and a significantly higher \(40\%\) for Hard. We verified these \(3{,}699\) filtered problems with \texttt{GPT-5} at a cost of approximately \(\$1{,}000\), resulting in \(1{,}106\) Easy (18\% failure rate), \(1{,}092\) Medium (24\% failure rate), and \(631\) Hard (\(31\%\) failure rate). At the estimated rate of \(\$0.25\) per problem, the human-adapted datasets cost \(\$240\) for generation and quality metrics. After quality filtering, we retained \(480\) QFT Pedagogy (\(15\%\) failure rate) and \(333\) arXiv (\(16\%\) failure rate). The combined cost of generation and verification for all datasets is approximately \(\$2{,}500\).

\subsection{Comments on Data Contamination} 
In this work, we train on fully synthetic tasks and evaluate on held-out synthetic tasks, human-adapted tasks, and the human-authored, private benchmark TPBench \cite{chung2025theoreticalphysicsbenchmarktpbench}. Because the human-adapted tasks are derived from publicly available online resources, they admit the possibility of data contamination through two distinct pathways. 
The first is direct contamination: our base model (\texttt{DeepSeek-7B}) was pre-trained on large scientific corpora that may have included the source documents from which these problems were adapted. 
The second is indirect contamination: the frontier model used to generate our synthetic training data may itself have been trained on those documents, so that memorized content could have been transferred into the training set even though no human-adapted problem was ever presented to the base model during training. 
We note that TPBench, being private and human-authored, is not subject to either pathway, and we therefore treat it as our cleanest measure of generalization.

While we cannot rule out contamination of the human-adapted datasets, we do not view it as a primary driver of our results (\Cref{sec:exp1,sec:exp2}): RL yields mean@5 gains of +5.6, +5.9, and +5.7 points on QFT Pedagogy, arXiv, and TPBench respectively, which are statistically indistinguishable across the three. The SFT mean@5 gains point the same way: the largest improvement (+3.3 points) occurs on TPBench, against +1.3 and +0.4 on the two human-adapted sets. Neither training regime therefore preferentially benefits the benchmarks for which contamination is possible.

\subsection{Challenges in TP Data Curation}
\label{sec:challenges_in_tp_data}
In developing our data generation pipeline, mitigating errors inherent to LLM-based generation and filtering proved to be the primary challenge. While our review pipeline (\Cref{sec:data_generation}) substantially reduces these inaccuracies, a residual error rate inevitably remains.
The errors we find by manual inspection are almost entirely in problems generated by \texttt{Gemini-2.5-Pro}. We observe a significant reduction in errors and an increase in problem quality following the release of \texttt{Gemini-3-Pro}, suggesting that the generation of high-quality synthetic data will only become easier as model capabilities continue to progress.
Below, we detail the specific challenges encountered and our mitigation strategies.

\paragraph{Incorrect Frontier Model Solutions}
When first developing synthetic problems, we found through LLM flagging and human review that a small fraction (\(<5\%\)) of \texttt{Gemini-2.5-Pro}-generated problems contained ambiguities in the problem statement or errors in the golden solution. To mitigate this, we implemented the review and cross-validation pipeline described in \Cref{sec:data_generation}. While this pipeline eliminates egregious errors, we cannot rule out the possibility of residual ones. We have therefore manually corrected or deleted the few erroneous problems we discovered after generation.

\paragraph{Calculation Conventions}
In theoretical physics, the choice of mathematical or physical convention is often arbitrary. For example, physicists use both the mostly minus $(+,-,-,-)$ and mostly plus $(-,+,+,+)$ metric signatures, which can flip the sign of a calculated quantity. Code-based verification requires fixing these conventions so that all valid solution attempts yield the exact same numerical outputs for test case comparisons. Furthermore, purely mathematical operations, such as the normalization of Fourier and Legendre transforms, also require explicit convention definitions. To address this, the problem generator defines a minimal, unambiguous set of conventions, which we append to the problem statement when necessary.

While defining calculation conventions is necessary for our implementation of code-based evaluation, we find that it introduces another avenue for errors. In a small number of problems, the calculation conventions written by the generating model trivialize the task by revealing too much of the calculation approach. Notably, we became aware of these problematic conventions only after our \texttt{DeepSeek-7B} RL fine-tuning in \Cref{sec:exp1}, and subsequently corrected them. We find that validation performance drops from approximately \(60 \to 54\), confirming that some tasks were indeed trivialized.

For future synthetic data generation, we would instead define a global set of calculation conventions to be provided to the generating model, eliminating the need for per-problem conventions. This global list would also be provided to all models attempting to solve the synthetic tasks, removing the possibility of information leakage through the conventions.

\paragraph{Test Case Generation}
During early development, we generated test cases by randomly sampling the input parameter space. However, this approach frequently produced parameters outside the physically allowed range. Since we use Python code generation strictly as a mechanism to verify physics results, evaluating non-physical test cases probes a model's edge-case programming robustness rather than its physical reasoning. Consequently, we instruct the generator model to select test cases strictly confined within valid physical bounds.

\paragraph{Code Implementation Errors}
During analysis, we identified an instance in which the golden solution code was incomplete, returning valid solutions for only a subset of the generated test cases. While rare, this resulted in false negatives, where attempts by other frontier models failed our verification pipeline simply because their generated code was more robust than the golden solution.

\subsection{Evaluation Protocol and Uncertainty Quantification}
\label{sec:eval_and_uncertainty}
LLM generation is typically optimized through stochastic decoding, described by parameters such as temperature, top-p, and top-k.
The optimal parameter values are provided by the model's creator, and \textit{greedy} decoding is commonly discouraged.
As a result, a single model exhibits variable performance on a fixed task.
To account for this variation, we must quantify the uncertainties in our reported validation scores.

In this work, we primarily compare the performance of a base model (\texttt{DeepSeek-R1-Distill-Qwen-7B} \cite{deepseekai2025deepseekr1incentivizingreasoningcapability}) to that of its RL and SFT variants. For all final-model performance benchmarks, we use a maximum response length of \(32{,}768\) tokens.
For the base, RL, and SFT models we generate 100 attempts for every problem in each validation set. 
We bootstrap our performance metrics by performing 1,000 resampling iterations.
In each iteration, for every problem in the validation set, we sample five attempts with replacement from its pool of 100 generated attempts. Thus, for an 80-problem validation set, each iteration consists of 80 independent five-attempt samples. 
We average each metric across all problems to obtain one dataset-level mean@5 and pass@5 value per iteration. We report the mean and standard deviation of these values across the 1,000 bootstrap iterations, both for the full dataset and within each subgroup.
We report results with bootstrapping as mean $\pm$ standard deviation.
To limit computational overhead, we generate only 5 attempts per problem for all other models presented, and therefore report their mean@5 and pass@5 without uncertainty estimates.

\section{Short Review of Fine-Tuning Methods}
\label{sec:train_methods}

We briefly review the training methods of supervised fine-tuning (SFT) and reinforcement learning with group relative policy optimization (GRPO) and the relationship between them.
For simplicity, we adopt the standard RL conventions throughout this section. We refer to the model (here the LLM) as the \enquote{policy}, which we represent as \(\pi_{\theta}\) with learnable parameters \(\theta\).

\subsection{Supervised Fine-Tuning}
Our training dataset contains pairs \((x,y^*)\) of problem prompts \(x\) and expert solutions \(y^*\).
An expert solution is given by a sequence of tokens
\begin{equation}
y^* = (y^*_1,\dots,y^*_T).
\end{equation}
For an LLM (or \enquote{policy}) $\pi_\theta$, the goal of SFT for reasoning is to increase the probability that, for prompt \(x\), $\pi_\theta$ produces the expert solution \(y^*\).
The LLM generates the answer sequence autoregressively (token by token), and therefore the probability that $\pi_\theta$ generates the expert solution factorizes per token as
\begin{equation}
\pi_\theta(y^* \mid x)
=
\prod_{t=1}^T \pi_\theta(y_t^* \mid x, y_{<t}^*), \quad y_{<t}^* = (y_1^*,\dots,y_{t-1}^*).
\end{equation}
The SFT loss is the negative log-likelihood of the expert solution,
\begin{equation}
\mathcal{L}_{\mathrm{SFT}}
=
- \log \pi_\theta(y^* \mid x)
=
-\sum_{t=1}^T \log \pi_\theta(y_t^* \mid x, y_{<t}^*).
\label{eq:sft}
\end{equation}
The training loss thus teaches the model to make the expert solution a likely completion to the prompt, token by token.

\subsection{Reinforcement Learning with GRPO}
RL training does not make use of the expert trajectory $y^*$. Instead, we sample a large number of answers from the model and upweight those that earn a high reward (i.e., those that give the correct answer).
For a single prompt $x$ and the reference policy $\pi_{\theta_{\mathrm{old}}}$ (representing the LLM weights prior to the current optimization step), sample $K$ responses
\begin{equation}
y_1,\dots,y_K \sim \pi_{\theta_{\mathrm{old}}}(\cdot|x).
\end{equation}
Each response receives a reward
\begin{equation}
r_i = r(x,y_i)
\end{equation}
as defined through the \textit{reward function} \(r(\cdot, \cdot)\). Reward functions are designed by researchers to optimize the desired task or to encourage particular model behaviors (backtracking, code formatting, conciseness, etc.).

\paragraph{Calculating Advantages}
Next, we transform the raw rewards into \emph{advantages}. The advantage $A_i$ for each response $y_i$ is the difference between the reward $r_i$ and the expected reward for the prompt $x$, which isolates responses that perform better or worse than the baseline.
Proximal policy optimization (PPO), a standard technique for RL training of LLMs, uses a \emph{critic} model $V_\phi(x)$ with parameters \(\phi\) to estimate the expected reward and defines the advantage as
$$A_i = r_i - V_\phi(x).$$
For binary rewards, $V_\phi(x)$ estimates the probability of correctly answering the prompt.
In PPO, as the actor is updated, the critic must also be continuously trained to accurately predict the expected reward under the new policy.
Relying on this critic typically requires hosting and tuning a separate, fine-tuned LLM, which introduces additional computational overhead.

\paragraph{Group-Relative Advantage.}
GRPO \cite{shao2024deepseekmathpushinglimitsmathematical} improves upon PPO by approximating the expected reward as the mean reward, thereby eliminating the need for a critic model and reducing advantage variance through group sampling. Explicitly, the normalized group-relative advantage is
\begin{equation}
A_i = \frac{r_i-\bar r}{s_r+\delta},
\qquad
\bar r = \frac{1}{K}\sum_{j=1}^K r_j,
\qquad
s_r =
\sqrt{\frac{1}{K}\sum_{j=1}^K (r_j-\bar r)^2},
\end{equation}
where \(s_r\) is the standard deviation of the rewards within the group and
\(\delta>0\) is a small numerical constant that prevents division by zero.
The term \enquote{group-relative} refers to centering and scaling each reward using statistics computed from the \(K\) responses sampled for the same prompt.
In this way, each trajectory's weight depends on the rewards of all other trajectories in the same group.

\paragraph{GRPO Loss with Binary Rewards.}
The GRPO loss is calculated through the advantages \(A_i\) and is given by
\begin{equation}
\mathcal{L}_{\mathrm{GRPO}}
=
-\frac{1}{K}
\sum_{i=1}^{K}
\min\left(
\rho_i(\theta)A_i,
\text{clip}(\rho_i(\theta),1-\epsilon_{\text{low}},1+\epsilon_{\text{high}})A_i
\right),
\end{equation}
where $\rho$ is the policy ratio
\begin{align}
\rho_i(\theta)
&=
\frac{\pi_\theta(y_i|x)}
{\pi_{\theta_{\mathrm{old}}}(y_i|x)}.
\end{align}
The policy ratio compares the probability of generating response $y_i$ given prompt $x$ under the active policy $\pi_\theta$ to that under the old policy $\pi_{\theta_{\text{old}}}$ used for the initial response generation.
Because GRPO can perform multiple gradient updates on the same batch of generated responses, the current parameters $\theta$ are continuously updated and drift away from $\theta_{\text{old}}$ within a single training step.
The policy ratio acts as an importance sampling weight that corrects for this \textit{policy drift}, ensuring that updates to \(\pi_\theta\) remain valid for data originally sampled from $\pi_{\theta_{\text{old}}}$.
The clipping term prevents large policy updates that could derail the training process. As in our main training runs, we include clip-higher and omit the KL term and format rewards.

In most mathematical settings, including our own, rewards are binary:
\begin{equation}
r_i \in \{0,1\},
\end{equation}
defined by the correctness of the policy's outcome compared to the expert trajectory \(y^*\).
Defining \(M=\sum_{i=1}^{K}r_i\) as the number of correct completions, we have
\begin{equation}
\bar r = \frac{M}{K},
\qquad
s_r = \sqrt{\frac{M}{K}\left(1-\frac{M}{K}\right)},
\end{equation}
and the advantages become
\begin{equation}
A_i =
\begin{cases}
w_+ =
\dfrac{1-\frac{M}{K}}{s_r+\delta},
& r_i=1,\\[10pt]
w_- =
-\dfrac{\frac{M}{K}}{s_r+\delta},
& r_i=0.
\end{cases}
\end{equation}

\paragraph{Relation to SFT.}

If we briefly disregard the standard deviation normalization and clipping for clarity, the GRPO loss becomes
\begin{equation}
\mathcal{L}_{\mathrm{GRPO-noclip}}
=
-\frac{1}{K}
\sum_{i=1}^{K}
\rho_i(\theta)A_i.
\end{equation}
Using the first-order approximation of the policy ratio in $\Delta \theta$, we obtain
\begin{equation}
\mathcal{L}
\propto
-\sum_i A_i \log \pi_\theta(y_i|x),
\end{equation}
which can be written at the token level as
\begin{align}
\mathcal{L}_{\mathrm{GRPO-token}}
\propto
-&\Bigg[
w_+
\sum_{i:r_i=1}
\sum_{t=1}^{T_i}
\log \pi_\theta(y_{i,t}|x,y_{i,<t})
\\
&+
w_-
\sum_{j:r_j=0}
\sum_{t=1}^{T_j}
\log \pi_\theta(y_{j,t}|x,y_{j,<t})
\Bigg].
\end{align}
We thus see that the GRPO loss upweights correct trajectories and downweights incorrect ones. The first term is equivalent to the token-wise SFT loss, with the successful RL rollouts playing the role of expert trajectories.

\subsection{Comments on Model Distillation}
In our work, synthetic training data is generated by sampling from a stronger teacher model coupled with our verification pipeline.
Ultimately, both our SFT and RL experiments perform a form of model distillation. SFT explicitly enforces the teacher's exact CoT, fundamentally altering the student's reasoning patterns and knowledge base.
Distillation through RL on synthetic problems, by contrast, occurs indirectly: because the verifiable problems are generated by the teacher (a frontier model), what can be learned from the synthetic dataset is bounded by the teacher model's physics ability.
RL therefore distills the teacher's knowledge without directly transferring its reasoning style to the student.

While RL presents computational challenges, SFT has limitations as well.
First, many frontier models do not expose the full CoT, which limits the applicability of SFT.
Second, even when teacher CoTs are available, the teacher model may not be well matched to the knowledge base of the student model.
By refining its own CoTs through RL, the student model may therefore be able to improve further than SFT would allow.
We explore both methods in \Cref{sec:exp1} and \Cref{sec:exp2}, aiming to study training methods for theoretical physics rather than to achieve the highest possible score within our training budget.
While we have discussed RL and SFT as separate methods, most post-training pipelines use a combination of the two \cite{qwen3technicalreport}.

\section{Experiment 1: \texttt{DeepSeek-7B} RL}
\label{sec:exp1}
In the following sections, we fine-tune models on our synthetic dataset using both reinforcement learning (RL) and supervised fine-tuning (SFT). Our goal is to explore what gains we can make with our limited compute budget, how RL and SFT compare on in-distribution (ID) and out-of-distribution (OOD) data, and how errors evolve during fine-tuning.
We select the reasoning model \texttt{DeepSeek-R1-Distill-Qwen-7B} \cite{deepseekai2025deepseekr1incentivizingreasoningcapability}, referred to as \texttt{DeepSeek-7B} for brevity, for our exploratory training runs, primarily due to its suitable base capabilities and well-understood performance in reasoning.
We start by performing a full RL fine-tuning of \texttt{DeepSeek-7B} on the Easy QFT training dataset, using the Verl framework \cite{sheng2024hybridflow}.

In addition to \texttt{DeepSeek-7B}, we attempted RL fine-tuning of \texttt{Qwen3-4B-Thinking-2507} \cite{qwen3technicalreport}, but encountered difficulty extending its capabilities beyond base performance. We hypothesize that this inability to improve was due to the model being both low-capacity and extremely optimized, making it difficult to specialize without sacrificing its broad reasoning abilities. Larger models present additional challenges: in principle we can support them within our compute budget, but they incur prohibitively long sequence generation times. As such, we found \texttt{DeepSeek-7B} to be a balanced choice between performance and generation speed.

We further explore the suitability of RL for targeted, narrow-domain specialization through a LoRA fine-tuning experiment on fermion and spinor problems in Appendix~\ref{sec:exp3}.

\subsection{Results}
\paragraph{Training Parameters}
We perform a full fine-tuning using GRPO with an accuracy-only reward function, in which verified correct solutions (all test cases passing) receive a reward of $1$ and incorrect results receive a reward of $0$. If multiple Python code blocks are output, we consider only the final non-empty output code block for verification.

Our RL training parameters were guided by \textsc{JustRL} \cite{he2025justrl}. We generate 32 rollouts per prompt at a temperature of $T=1.0$.
To stabilize policy updates, we implement the \textsc{clip-higher} mechanism from DAPO \cite{yu2025dapoopensourcellmreinforcement} using an asymmetric clipping range ($\epsilon_{\rm low}=0.2$, $\epsilon_{\rm high}=0.28$). We find that \textsc{clip-higher} alone suffices to maintain entropy stability for small reasoning models, removing the need for entropy regularization or KL penalties \cite{he2025justrl}. Training uses a constant learning rate of $1 \times 10^{-6}$, a global batch size of 4 (effective batch size of 128), and a maximum response length of 20,000 tokens. This training was conducted on a single node of 4$\times$H200 GPUs at the Perimeter Institute over 160 hours (640 GPU hours). We evaluate checkpoints throughout training and use validation performance for early stopping, selecting the best-performing checkpoint (step \(2{,}200\)) as the final RL model. All final performance benchmarks for the RL model are generated anew from this selected checkpoint using 100 newly sampled attempts per problem.

\begin{tcolorbox}[colback=blue!5!white, colframe=blue!50!black, title=\textbf{Key Takeaways: \texttt{DeepSeek-7B} RL}, arc=4pt, boxrule=0.5pt, left=6pt, right=6pt, top=4pt, bottom=4pt]
\begin{itemize}
    \item Training on the Easy set (\resEasyMeanKBase\% $\rightarrow$ \resEasyMeanKRL\%) rapidly improves physics performance and induces strong zero-shot transfer to Medium problems (\resMediumMeanKBase\% $\rightarrow$ \resMediumMeanKRL\%).
    \item The trained model slightly improves on the Hard dataset (\resHardMeanKBase\% $\rightarrow$ \resHardMeanKRL\%).
    \item We observe consistent gains across the human-adapted datasets as well as TPBench.
\end{itemize}
\end{tcolorbox}

\begin{figure}[H]
    \centering

    \includegraphics[width=\linewidth]{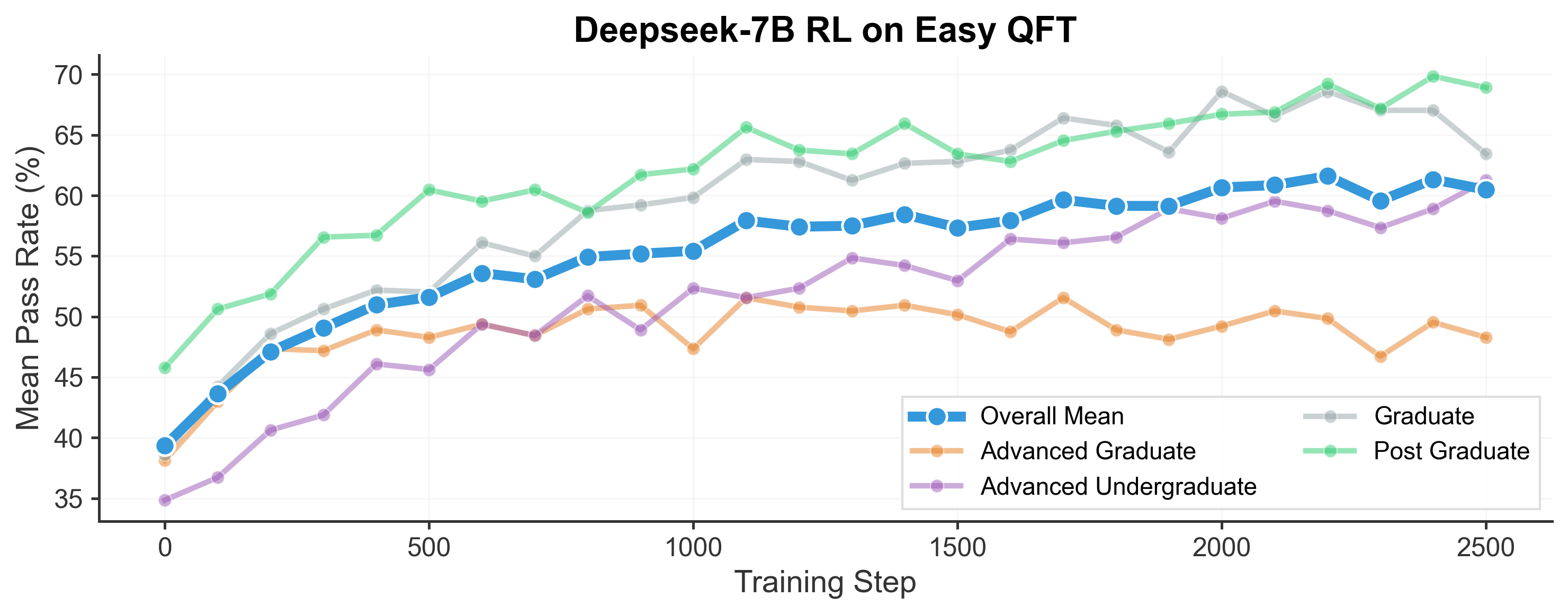}
    \caption{Easy QFT validation performance throughout RL training. The model starts from its base zero-shot performance and rapidly converges to peak validation performance.}
    \label{fig:easy_mean}
   
\end{figure}

The training improvement is summarized in \Cref{tab:easy_rl}, with stepwise mean@32 validation performance shown in \Cref{fig:easy_mean}. Note that the stepwise trajectories in \Cref{fig:easy_mean} report the mean accuracy over $k=32$ rollouts (as recorded during training), whereas the summary statistics in \Cref{tab:easy_rl} strictly adhere to the $k=5$ standard used throughout this work.

The peak accuracy in \Cref{fig:easy_mean} appears higher than the values reported in \Cref{tab:easy_rl} because we subsequently corrected the calculation convention statements in the validation set, which we judged to have provided crucial information on the solution approach and thereby trivialized the tasks. \Cref{fig:easy_mean} shows the uncorrected validation set employed during RL, and is the only result reported in this work without the correction for calculation conventions. This challenge is further described in \Cref{sec:challenges_in_tp_data}.

\paragraph{Rapid Convergence and Transfer}
On the Easy dataset, the model demonstrates rapid capability acquisition, improving from a mean@5 baseline of \resEasyMeanKBase\% to \resEasyMeanKRL\%. As shown in \Cref{fig:easy_mean}, the mean@32 validation score climbs steeply over the first 1,000 steps, indicating that the 7B model can quickly internalize the requisite introductory physics priors. This stage induces strong ability transfer to the Medium set, where validation mean@5 jumps from \resMediumMeanKBase\% to \resMediumMeanKRL\% solely due to training on Easy.

\paragraph{Performance Variation across Domain Levels}
As shown in \Cref{fig:easy_mean}, the four domain levels exhibit varying degrees of difficulty both before and after RL. Paradoxically, postgraduate topics emerge as the easiest for both the base and RL models.
The task distribution (shown in \Cref{fig:task_distribution}) does not drive this difficulty variation, as hidden-coefficient derivations dominate uniformly across all levels, comprising 14, 13, 12, and 15 out of 20 problems for the AU, GR, AG, and PG splits, respectively.

We hypothesize that this trend stems from biases introduced during data generation, specifically: \textbf{(a)} an ambiguous definition of what constitutes an ``easy'' postgraduate problem, and \textbf{(b)} a tendency for the generating models to overcompensate for the advanced subject matter by defaulting to low-complexity reasoning steps. 
We provide a characteristic example in Appendix \ref{sec:ex_prob_easy_pg} of an Easy QFT postgraduate problem that involves advanced topics but requires only simple reasoning to solve the presented tasks.

Conversely, the hypothesized \textit{overcompensation} bias may operate in the reverse direction for advanced undergraduate (AU) topics, making them operationally harder at a fixed ``easy'' target. This effect is supported by the initial performance, in which AU problems are the most challenging (33\% base accuracy). Importantly, while the observed performance metrics align with these generation biases, we cannot fully isolate the biases and therefore present these mechanisms strictly as hypotheses.

\paragraph{The Hard Reasoning Barrier} 
Easy RL raises performance on the QFT Hard dataset from a near-total inability (\resHardMeanKBase\%) to a non-zero capability (\resHardMeanKRL\%). For challenging reasoning tasks, it is typical to measure performance under test-time scaling with a large number of parallel samples (best-of-256, best-of-1024, etc.) or with extended response lengths. Given our compute budget, we avoid these techniques, though this may limit the gains we find on QFT Hard.

\begin{table}[H]
    \small
    \centering
    \caption{\texttt{DeepSeek-7B} performance after RL training on Easy QFT (1,000 resampling iterations; reported as bootstrap mean \(\pm\) standard deviation). For detailed performance metrics, refer to Appendix \ref{app:all_results}.}
    \label{tab:easy_rl}

    \begin{subtable}{\textwidth}
        \centering
        \caption{\textbf{Synthetic Dataset Performance}}
        \begin{tabular}{l cc cc cc}
        \toprule
        & \multicolumn{2}{c}{\textbf{Easy}} & \multicolumn{2}{c}{\textbf{Medium}} & \multicolumn{2}{c}{\textbf{Hard}} \\
        \cmidrule(lr){2-3} \cmidrule(lr){4-5} \cmidrule(lr){6-7}
        \textbf{Stage} & \textbf{mean@5} & \textbf{pass@5} & \textbf{mean@5} & \textbf{pass@5} & \textbf{mean@5} & \textbf{pass@5} \\
        \midrule
        Base
            & \resEasyMeanKBase   & \resEasyPassBase
            & \resMediumMeanKBase & \resMediumPassBase
            & \resHardMeanKBase   & \resHardPassBase \\
        + RL on E
            & \resEasyMeanKRL   & \resEasyPassRL
            & \resMediumMeanKRL & \resMediumPassRL
            & \resHardMeanKRL   & \resHardPassRL \\
        $\Delta$
            & \dlt{\resEasyMeanKBaseMean}{\resEasyMeanKRLMean}
            & \dlt{\resEasyPassBaseMean}{\resEasyPassRLMean}
            & \dlt{\resMediumMeanKBaseMean}{\resMediumMeanKRLMean}
            & \dlt{\resMediumPassBaseMean}{\resMediumPassRLMean}
            & \dlt{\resHardMeanKBaseMean}{\resHardMeanKRLMean}
            & \dlt{\resHardPassBaseMean}{\resHardPassRLMean} \\
        \bottomrule
        \end{tabular}
    \end{subtable}

    \vspace{1em}

    \begin{subtable}{\textwidth}
        \centering
        \caption{\textbf{Benchmark Dataset Performance} (Overall)}
        \begin{tabular}{l cc cc cc}
        \toprule
        & \multicolumn{2}{c}{\textbf{arXiv}} & \multicolumn{2}{c}{\textbf{QFT Pedagogy}} & \multicolumn{2}{c}{\textbf{TPBench}} \\
        \cmidrule(lr){2-3} \cmidrule(lr){4-5} \cmidrule(lr){6-7}
        \textbf{Stage} & \textbf{mean@5} & \textbf{pass@5} & \textbf{mean@5} & \textbf{pass@5} & \textbf{mean@5} & \textbf{pass@5} \\
        \midrule
        Base
            & \resArXivMeanKBase   & \resArXivPassBase
            & \resQFTMeanKBase     & \resQFTPassBase
            & \resTPBenchMeanKBase & \resTPBenchPassBase \\
        + RL on E
            & \resArXivMeanKRL   & \resArXivPassRL
            & \resQFTMeanKRL     & \resQFTPassRL
            & \resTPBenchMeanKRL & \resTPBenchPassRL \\
        $\Delta$
            & \dlt{\resArXivMeanKBaseMean}{\resArXivMeanKRLMean}
            & \dlt{\resArXivPassBaseMean}{\resArXivPassRLMean}
            & \dlt{\resQFTMeanKBaseMean}{\resQFTMeanKRLMean}
            & \dlt{\resQFTPassBaseMean}{\resQFTPassRLMean}
            & \dlt{\resTPBenchMeanKBaseMean}{\resTPBenchMeanKRLMean}
            & \dlt{\resTPBenchPassBaseMean}{\resTPBenchPassRLMean} \\
        \bottomrule
        \end{tabular}
    \end{subtable}

\end{table}

\paragraph{Benchmark Performance Gains}
For detailed performance on subsets of arXiv, QFT Pedagogy, and individual TPBench levels, please refer to Appendix \ref{app:all_results}.
On the human-adapted datasets, mean@5 performance on arXiv increases from \resArXivMeanKBase\% to \resArXivMeanKRL\%, while QFT Pedagogy performance improves from \resQFTMeanKBase\% to \resQFTMeanKRL\%. Within the arXiv benchmark, we record consistent gains across all tested subdomains, indicating a broader enhancement of physics reasoning capabilities beyond QFT. For the QFT Pedagogy set, the majority of the improvement is concentrated in the exercise books (Radovanovic \cite{Radovanovic:2008zz}, Cheng--Li \cite{Cheng:1984vwu}), which improve from a mean@5 of \(9.4 \pm 0.6\)\% to \(16.4 \pm 0.7\)\%. On TPBench, we find performance gains at all levels,\footnote{Excluding level 5 (research-level), where both models fail to solve any problems.} with the largest gains on level 2 problems, going from \(67.2 \pm 5.0\)\% to \(80.3 \pm 3.6\)\%.

\section{Experiment 2: \texttt{DeepSeek-7B} SFT}
\label{sec:exp2}
In conjunction with our RL experiment in \Cref{sec:exp1}, we perform an analogous fine-tuning of \texttt{DeepSeek-7B} with SFT using Axolotl \cite{axolotl}. Whereas RL explores the model's ability to learn physics tasks by refining its own successful attempts, SFT investigates whether training on the successful attempts of more powerful \enquote{teacher} models yields comparable improvements. 
The training and validation splits of synthetic problems used for RL and SFT are identical, allowing for a direct comparison of the methods. There is no overlap between the training and validation problem sets. The rejection-sampled CoT traces used for SFT training are generated exclusively from training problems; traces generated for validation problems are used only for evaluation.

\subsection{SFT Dataset Generation}

To develop a robust SFT dataset for physics reasoning, we require high-quality reasoning traces from capable teacher models. Because the full reasoning traces of proprietary frontier models are currently inaccessible, we rely on open-weight alternatives. Operating within our compute budget, we select three teacher models for this task: \texttt{gpt-oss-120B} (medium reasoning effort) \cite{openai2025gptoss120bgptoss20bmodel}, \texttt{Qwen3-30B-A3B} \cite{qwen3technicalreport}, and \texttt{Qwen3.5-122B-A10B} \cite{qwen3.5}. Using these models, we generate five solution attempts for every problem across the training and validation splits of our synthetic datasets. We apply rejection sampling \cite{2308.01825} to filter all generated CoTs, retaining only the traces that yield the correct final result.

We generate the reasoning traces at the default inference temperature specified by each model provider and with a maximum response length of 32,768 tokens. For further information on our SFT dataset, please refer to \Cref{sec:SFT_dataset_info}.
This dataset was generated on a single node of 4$\times$H200 GPUs at the Perimeter Institute over approximately 36 hours.

\subsection{Results}
\paragraph{Training Parameters}
We run SFT for a total of 5 epochs with a sequence length of 35,000 tokens, utilizing sample packing. We experiment with longer training for up to 10 epochs, but find equivalent performance and therefore settle on 5 epochs as our baseline. Training was conducted on 4$\times$H200 GPUs with a micro-batch size of 1 per GPU and 2 gradient accumulation steps, yielding an effective global batch size of 8. We use the fused AdamW optimizer with a peak learning rate of $2 \times 10^{-5}$, a cosine decay schedule following a 10\% warmup period, and a weight decay of 0.01. We conduct several SFT experiments, with training times ranging from 1 to 5 hours each.

\begin{tcolorbox}[colback=blue!5!white, colframe=blue!50!black, title=\textbf{Key Takeaways: \texttt{DeepSeek-7B} SFT}, arc=4pt, boxrule=0.5pt, left=6pt, right=6pt, top=4pt, bottom=4pt]
\begin{itemize}
    \item \texttt{Qwen3-30B-A3B} CoT traces yield the highest performance gains for \texttt{DeepSeek-7B} SFT, surpassing both \texttt{gpt-oss-120B} and \texttt{Qwen3.5-122B-A10B}.
    \item Validation mean@5 continues to rise even as SFT begins to overtrain, over the range we explore.
    \item SFT models outperform RL on in-distribution synthetic tasks, whereas RL generalizes better to the human-adapted benchmarks and TPBench.
\end{itemize}
\end{tcolorbox}

To understand the dynamics of SFT on our synthetic QFT datasets, we perform two exploratory tests to determine \textbf{(a)} the relationship between validation mean@5 and the SFT loss, and \textbf{(b)} the optimal teacher model for \texttt{DeepSeek-7B} SFT.

\paragraph{Validation mean@5 vs.\ Cross-Entropy}

\begin{figure}[H]
    \centering
    \includegraphics[width=\linewidth]{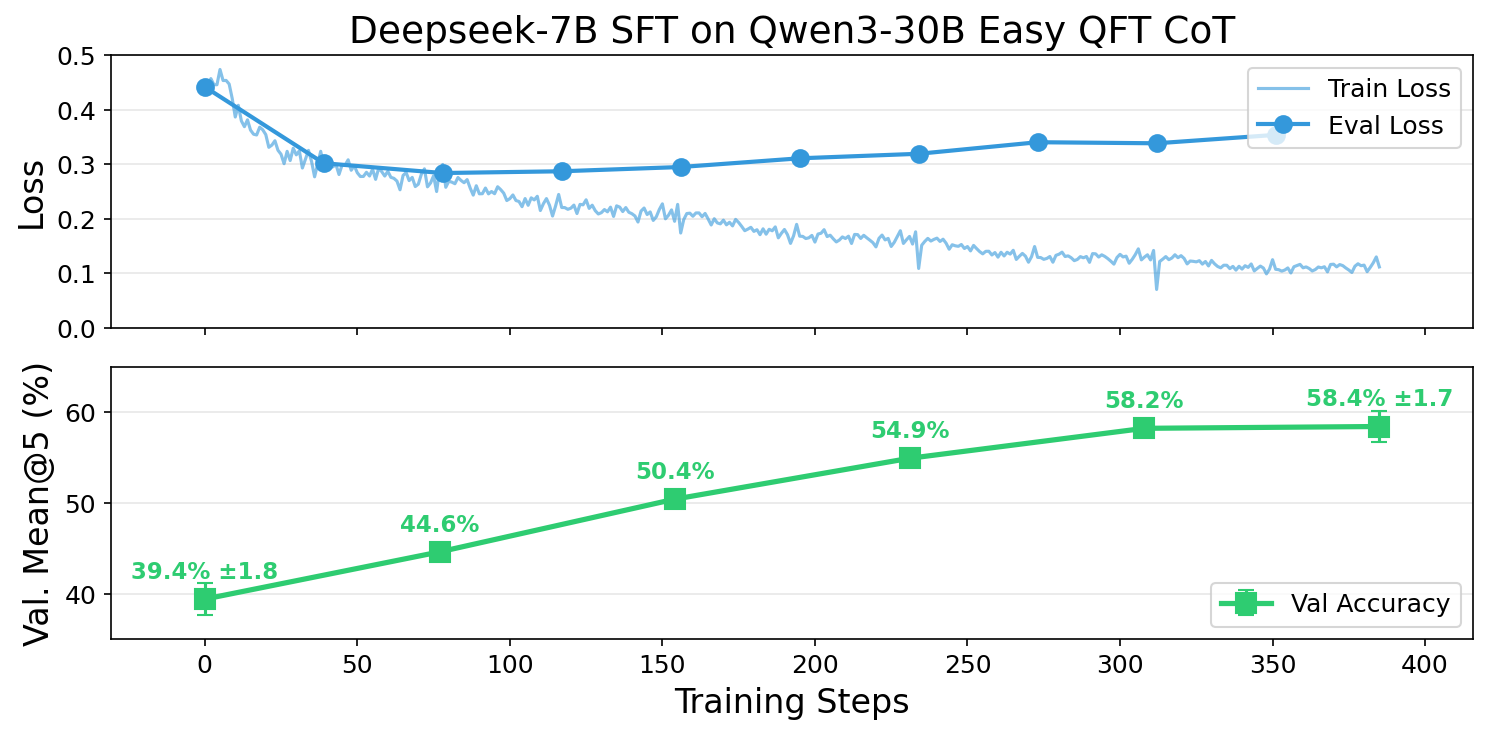}
    \caption{SFT on \texttt{Qwen3-30B-A3B} traces yields performance gains in validation mean@5. We find that these gains continue to grow even as the model begins to overtrain during SFT.}
    \label{fig:sft_qwen3_30b}
\end{figure}

As detailed in \Cref{sec:train_methods}, SFT does not directly optimize the model's ability to solve a task on its own, but rather the token-wise cross-entropy evaluated on the reasoning traces in our dataset. To accurately assess the efficacy of SFT for reasoning on our datasets, we first need to understand the relation between cross-entropy loss trends during training and the corresponding mean@5 at evaluation. To address this, we select the responses of \texttt{Qwen3-30B-A3B} on QFT Easy and evaluate the model checkpoint at every epoch during an SFT training run. The training loss and evaluation mean@5 scores are shown in \Cref{fig:sft_qwen3_30b}. During training, the model begins to exhibit signs of overtraining around step \(100\); nevertheless, evaluation mean@5 continues to rise from \(44.6\% \to \resEasyMeanKSFT\%\) through the remaining training steps. This suggests that, at least with our SFT hyperparameters, the degree of overtraining we observe is not detrimental to evaluation mean@5. In the following experiments we therefore evaluate the final SFT checkpoint.

\paragraph{Benchmarking Teacher Models for SFT}
Generating reasoning traces from multiple teacher models provides a diverse dataset for SFT. However, variation in reasoning styles may introduce challenges. First, a significant divergence from the base model's (\texttt{DeepSeek-7B}) inherent style---such as differences in verbosity or self-reflection frequency---may diminish performance gains. Second, while aggregating outputs increases dataset scale, mixing reasoning styles could introduce noise that complicates a small model's ability to converge on a consistent policy. 

To evaluate the impact of teacher model selection on SFT, we conduct experiments across four data configurations for Easy QFT: \texttt{gpt-oss-120B}, \texttt{Qwen3-30B-A3B}, \texttt{Qwen3.5-122B-A10B}, and a combined dataset of all three models. The resulting mean validation scores are shown in \Cref{tab:qft_easy_val}. We find that \texttt{Qwen3-30B-A3B} yields the largest performance gains, despite being the least proficient model on the underlying training and validation tasks.
We hypothesize that several factors contribute to this: \textbf{(a)} compared to \texttt{DeepSeek-7B}, \texttt{gpt-oss-120B} produced significantly shorter CoTs, while \texttt{Qwen3.5-122B-A10B} produced longer ones; \textbf{(b)} the concise \texttt{gpt-oss-120B} CoTs lacked the frequent reflection and backtracking we observe in \texttt{DeepSeek-7B} CoTs; and \textbf{(c)} the reasoning chains produced by 100B+ models may represent a complexity leap that a 7B model cannot consistently emulate, leading to diminished gains. Taken together, this suggests that the \texttt{Qwen3-30B-A3B} reasoning traces are the closest match to those of \texttt{DeepSeek-7B}, allowing SFT to sharpen the student's existing reasoning patterns rather than displace them.


\begin{table}[H]
\centering
\begin{tabular}{l c cccc}
\toprule
& & \multicolumn{4}{c}{\textbf{SFT Teacher Model (CoT Source)}} \\
\cmidrule(lr){3-6}
\textbf{Student Model} & \textbf{Base} & \textbf{gpt-oss-120B} & \textbf{Qwen3-30B-A3B} & \textbf{Qwen3.5-122B-A10B} & \textbf{Combined} \\
\midrule
\texttt{DeepSeek-7B} & \resEasyMeanKBase & 50.4 & \resEasyMeanKSFT & 47.3 & 50.4 \\
\bottomrule
\end{tabular}
\caption{Mean@5 on the QFT Easy validation set for the base model and after SFT on the Easy QFT traces generated by each teacher model.}
\label{tab:qft_easy_val}
\end{table}

\begin{figure}[H]
    \centering
    \includegraphics[width=1\linewidth]{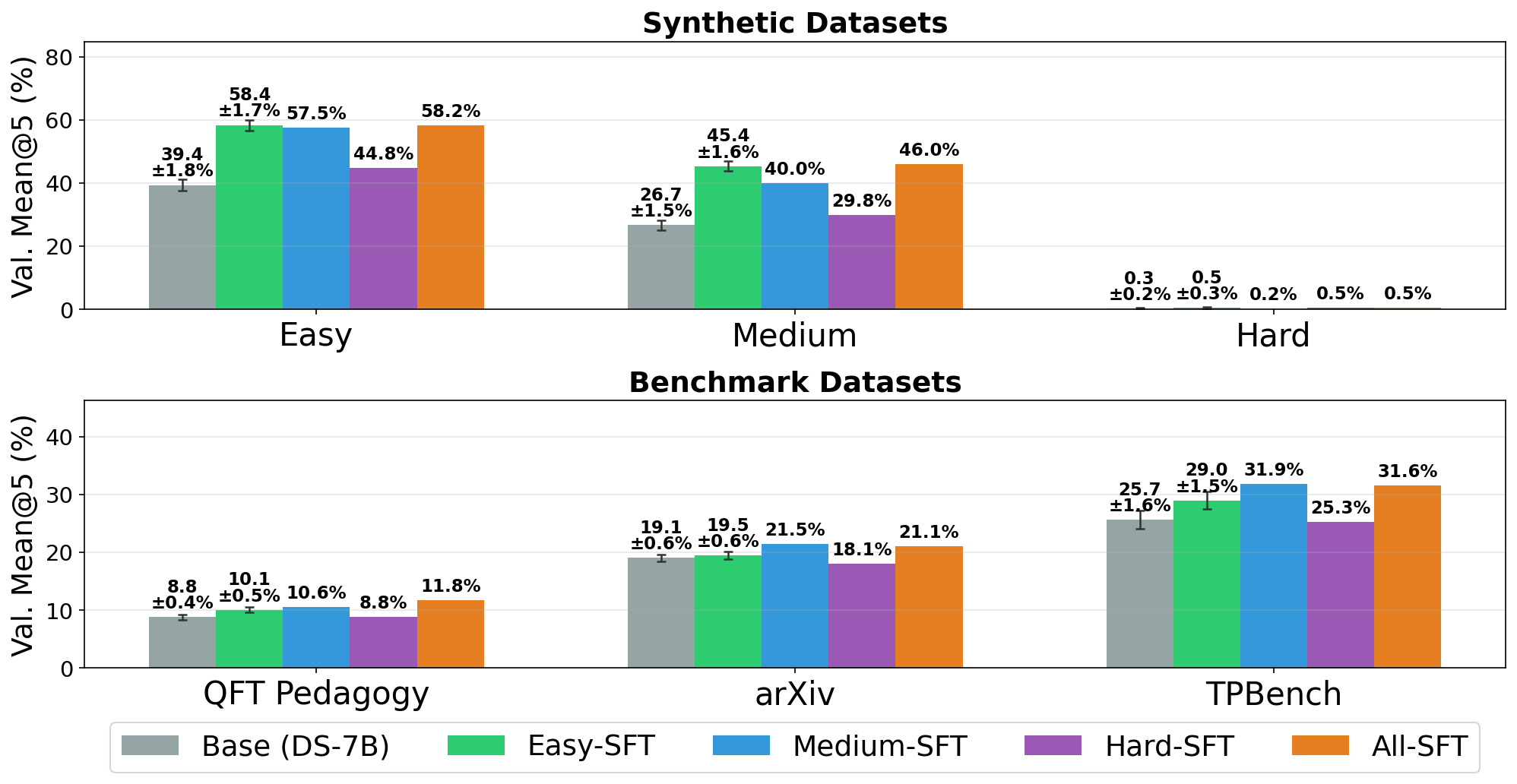}
    \caption{Performance comparison between the base model (\texttt{DeepSeek-7B}) and the four SFT variants on the synthetic (top) and human-adapted datasets and TPBench (bottom).}
    \label{fig:sft_stacked_performance}
\end{figure}

\paragraph{SFT Performance Gains}

We conduct four SFT training runs using the \texttt{Qwen3-30B-A3B} Easy, Medium, and Hard datasets, and a combined dataset (Easy + Medium + Hard). To allow for a direct comparison with the QFT Easy RL model, we generate 100 attempts for the Easy-SFT model; for all other SFT models we generate 5 attempts. The performance of the resulting SFT models, alongside the base model, is summarized in \Cref{fig:sft_stacked_performance}. While temperature is not a factor during SFT, it remains a tunable hyperparameter at inference time. We find that the model's default temperature ($T=0.6$) yields optimal post-SFT performance. Across the SFT models, we observe consistent performance gains on both synthetic and benchmark datasets, with the exception of the Hard synthetic tasks, where all models remain near zero.

Notably, the Easy-SFT model demonstrates substantial improvements on the synthetic datasets: mean@5 increases from \(\resEasyMeanKBase\% \to \resEasyMeanKSFT\%\) on the Easy validation set. Furthermore, on the Medium validation set, the Easy-SFT model improves performance from \(\resMediumMeanKBase\% \to \resMediumMeanKSFT\%\), outperforming the model trained explicitly on the Medium dataset (which achieves 40.0\%). Conversely, on the human-adapted benchmark datasets (QFT Pedagogy, arXiv) and TPBench, the Medium-SFT and All-SFT models exhibit the largest overall performance gains.
Detailed performance of the Easy-SFT model, with uncertainties, is listed in \Cref{tab:easy_SFT}.
\begin{table}[H]
    \small
    \centering
    \caption{\texttt{DeepSeek-7B} performance after SFT on Easy QFT reasoning traces generated by \texttt{Qwen3-30B-A3B} (1,000 resampling iterations; reported as bootstrap mean \(\pm\) standard deviation). For detailed performance metrics, refer to Appendix \ref{app:all_results}.}
    \label{tab:easy_SFT}

    \begin{subtable}{\textwidth}
        \centering
        \caption{\textbf{Synthetic Dataset Performance}}
        \begin{tabular}{l cc cc cc}
        \toprule
        & \multicolumn{2}{c}{\textbf{Easy}} & \multicolumn{2}{c}{\textbf{Medium}} & \multicolumn{2}{c}{\textbf{Hard}} \\
        \cmidrule(lr){2-3} \cmidrule(lr){4-5} \cmidrule(lr){6-7}
        \textbf{Stage} & \textbf{mean@5} & \textbf{pass@5} & \textbf{mean@5} & \textbf{pass@5} & \textbf{mean@5} & \textbf{pass@5} \\
        \midrule
        Base
            & \resEasyMeanKBase   & \resEasyPassBase
            & \resMediumMeanKBase & \resMediumPassBase
            & \resHardMeanKBase   & \resHardPassBase \\
        + SFT on E
            & \resEasyMeanKSFT   & \resEasyPassSFT
            & \resMediumMeanKSFT & \resMediumPassSFT
            & \resHardMeanKSFT   & \resHardPassSFT \\
        $\Delta$
            & \dlt{\resEasyMeanKBaseMean}{\resEasyMeanKSFTMean}
            & \dlt{\resEasyPassBaseMean}{\resEasyPassSFTMean}
            & \dlt{\resMediumMeanKBaseMean}{\resMediumMeanKSFTMean}
            & \dlt{\resMediumPassBaseMean}{\resMediumPassSFTMean}
            & \dlt{\resHardMeanKBaseMean}{\resHardMeanKSFTMean}
            & \dlt{\resHardPassBaseMean}{\resHardPassSFTMean} \\
        \bottomrule
        \end{tabular}
    \end{subtable}

    \vspace{1em}

    \begin{subtable}{\textwidth}
        \centering
        \caption{\textbf{Benchmark Dataset Performance} (Overall)}
        \begin{tabular}{l cc cc cc}
        \toprule
        & \multicolumn{2}{c}{\textbf{arXiv}} & \multicolumn{2}{c}{\textbf{QFT Pedagogy}} & \multicolumn{2}{c}{\textbf{TPBench}} \\
        \cmidrule(lr){2-3} \cmidrule(lr){4-5} \cmidrule(lr){6-7}
        \textbf{Stage} & \textbf{mean@5} & \textbf{pass@5} & \textbf{mean@5} & \textbf{pass@5} & \textbf{mean@5} & \textbf{pass@5} \\
        \midrule
        Base
            & \resArXivMeanKBase   & \resArXivPassBase
            & \resQFTMeanKBase     & \resQFTPassBase
            & \resTPBenchMeanKBase & \resTPBenchPassBase \\
        + SFT on E
            & \resArXivMeanKSFT   & \resArXivPassSFT
            & \resQFTMeanKSFT     & \resQFTPassSFT
            & \resTPBenchMeanKSFT & \resTPBenchPassSFT \\
        $\Delta$
            & \dlt{\resArXivMeanKBaseMean}{\resArXivMeanKSFTMean}
            & \dlt{\resArXivPassBaseMean}{\resArXivPassSFTMean}
            & \dlt{\resQFTMeanKBaseMean}{\resQFTMeanKSFTMean}
            & \dlt{\resQFTPassBaseMean}{\resQFTPassSFTMean}
            & \dlt{\resTPBenchMeanKBaseMean}{\resTPBenchMeanKSFTMean}
            & \dlt{\resTPBenchPassBaseMean}{\resTPBenchPassSFTMean} \\
        \bottomrule
        \end{tabular}
    \end{subtable}

\end{table}

\subsection{SFT vs.\ RL Performance}
\begin{figure}[H]
    \centering
    \includegraphics[width=\linewidth]{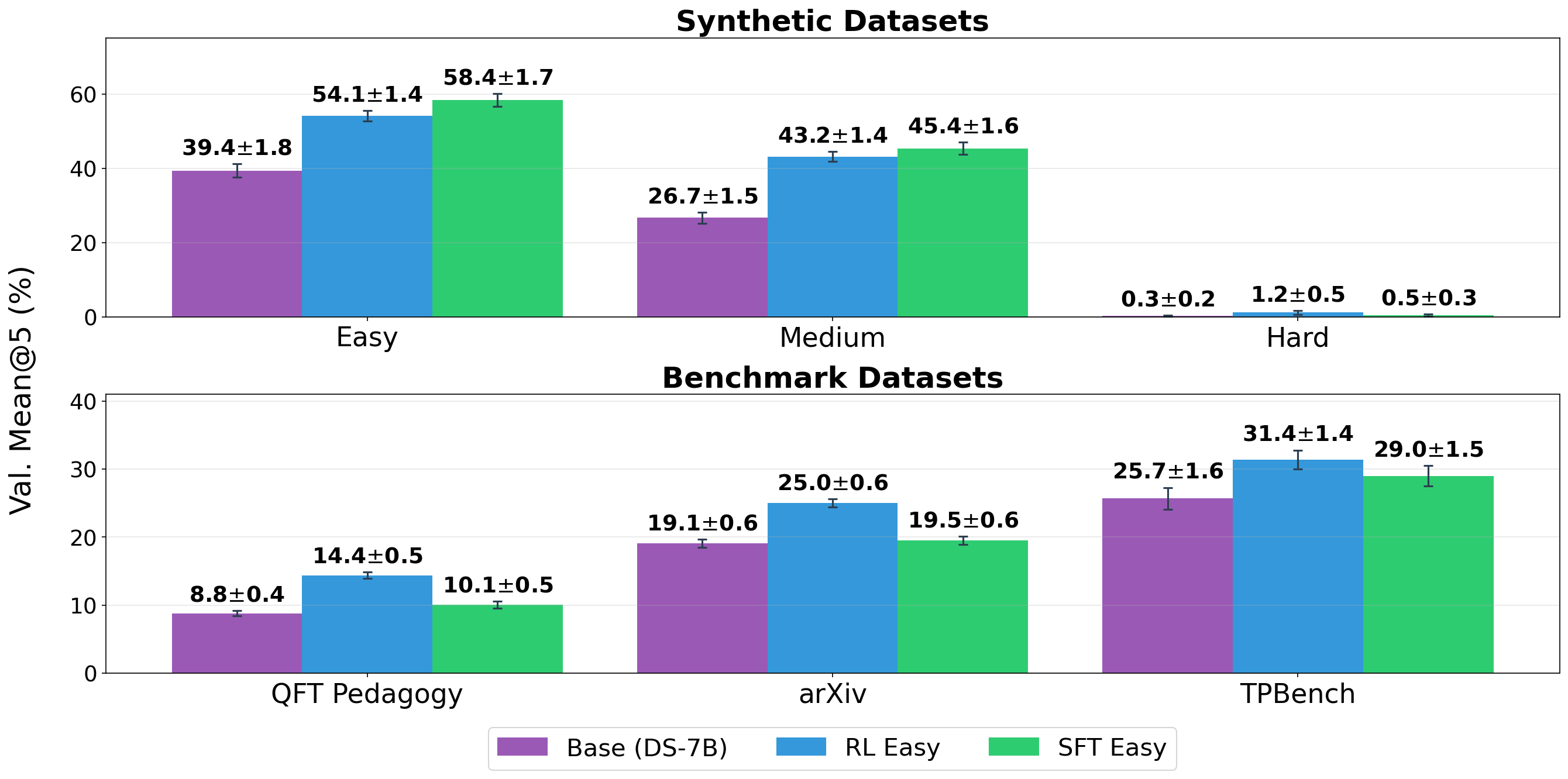}
    \caption{Performance comparison of RL on Easy QFT and SFT on Easy QFT responses from \texttt{Qwen3-30B-A3B}.}
    \label{fig:rl_v_sft}
    
\end{figure}

We summarize the performance gains through RL on Easy QFT and SFT on Easy QFT reasoning traces generated by \texttt{Qwen3-30B-A3B} in \Cref{fig:rl_v_sft}. 
On the synthetic in-distribution datasets, SFT generally achieves higher mean@5 performance than RL: SFT exceeds RL by 4.3 percentage points on Easy QFT and by 2.2 percentage points on Medium QFT. On Hard QFT, where both models achieve low absolute performance, RL exceeds SFT by 0.7 percentage points. On the out-of-distribution evaluations, RL performs better than SFT, exceeding it by 4.3 percentage points on QFT Pedagogy, 5.5 percentage points on arXiv, and 2.4 percentage points on TPBench. 
For further details on model performance, refer to Appendix \ref{app:all_results}.

\section{Analysis of Reasoning Errors Before and After Fine-Tuning}
\label{sec:error_analysis}
Next, to understand \emph{how} fine-tuning improves model reasoning, we develop an automated error analysis pipeline that classifies the types of errors present in incorrect rollouts. Designing such a pipeline proved non-trivial; in \Cref{sec:failed_cot} we describe the challenges that motivated our final design before presenting it in detail.

\subsection{Failed CoT Analysis Methods}
\label{sec:failed_cot}
In our initial attempt to analyze reasoning chains, we fed the entire chain-of-thought (CoT) as well as the golden solution to a strong LLM and prompted it to identify errors. This approach failed for CoT traces exceeding \(10{,}000\) tokens---which are common in our dataset---because the analyzer model struggled to isolate the substantive reasoning steps from the noise of the full CoT. A key challenge was the prevalence of \emph{self-correction callbacks}: reasoning models frequently backtrack, abandon approaches, and restart derivations (e.g., ``Wait, that's wrong, let me try again\ldots''), which obscures the actual reasoning path that led to the final answer. The analyzer would often attribute errors to abandoned reasoning branches rather than to the final derivation.

Next, we attempted a \emph{line-by-line} analysis, in which each line of the CoT was individually evaluated against the golden solution. This suffered from the same context limitations---without the full derivation context, the analyzer could not determine whether a given line was correct---and additionally encountered a granularity mismatch: a single logical step (e.g., applying an integration identity) often spans multiple lines, while a single line sometimes contains multiple logical operations. These boundary issues made line-level analysis unreliable.

Additionally, we explored defining error categories \emph{per problem}, allowing each problem its own taxonomy of common mistakes. While this captured problem-specific patterns, it made cross-problem aggregation and comparison intractable, as the resulting taxonomies could not be meaningfully unified.

Within our pipeline presented in \Cref{sec:distill-then-classify}, we experimented extensively with the size of the error label set. While larger sets provide more granular insight into the learning dynamics, they come at the direct cost of reproducibility: with large label sets, models frequently interchanged labels across independent runs. Expert human review confirmed that these alternative labels were often equally valid, which prompted us to define a minimal set of error labels to mitigate this inherent labeling uncertainty.

\subsection{Our Approach: Distill-then-Classify}
\label{sec:distill-then-classify}

\begin{figure}[t!]
    \centering
    \includegraphics[width=\linewidth]{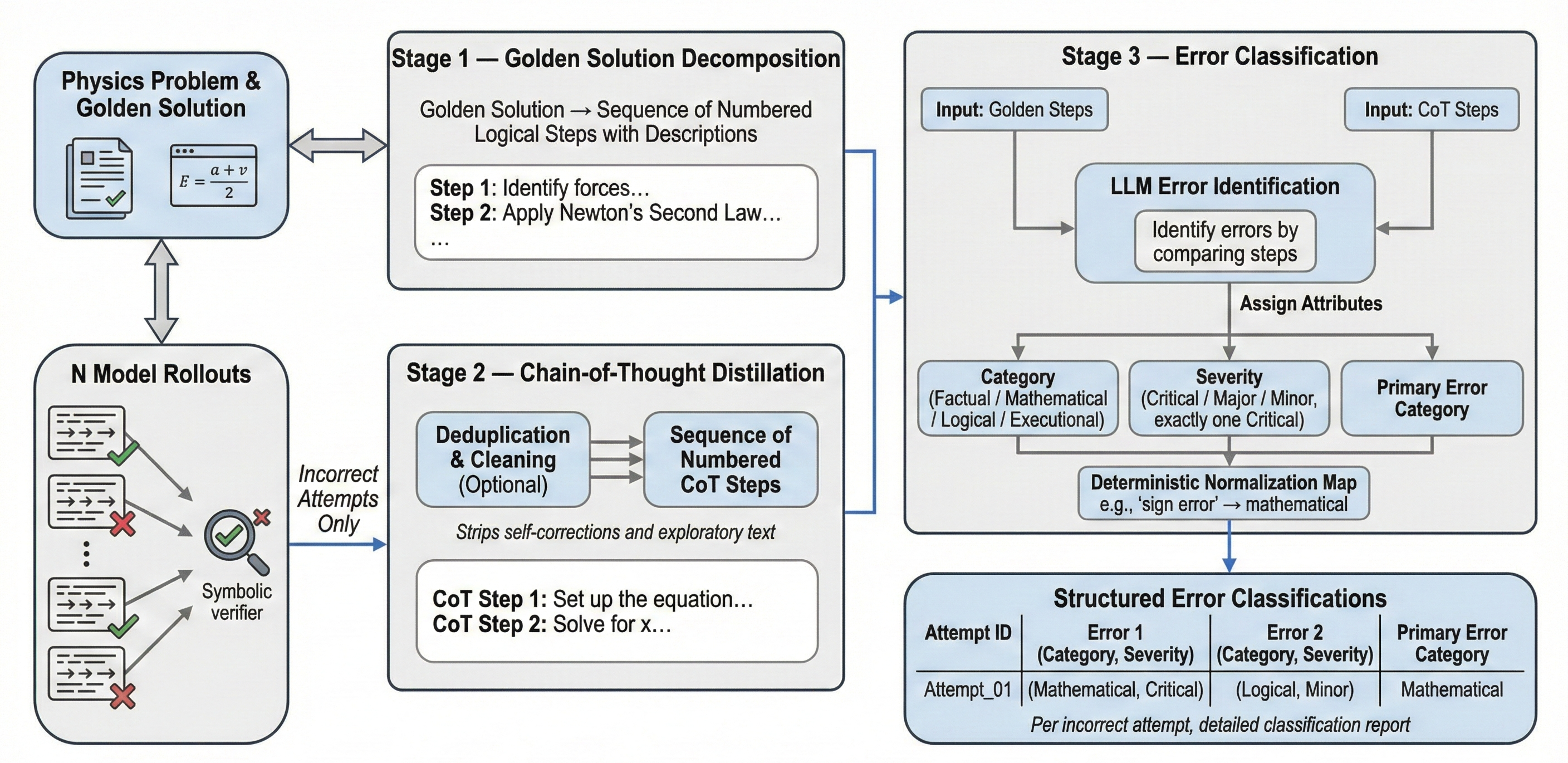}
    \caption{The three-stage error analysis pipeline. \textbf{Stage~1} decomposes 
the verified golden solution into a sequence of numbered logical steps. 
\textbf{Stage~2} distills each incorrect model rollout---after optional 
deduplication and cleaning to remove self-corrections and exploratory 
text---into an analogous step sequence. \textbf{Stage~3} feeds both step 
sequences to an LLM that identifies errors, assigns each a category 
(Factual / Mathematical / Logical / Executional) and severity 
(Critical / Major / Minor), designates a single primary error category, 
and normalizes labels via a deterministic map before producing a structured 
per-attempt classification report.}
    \label{fig:error_pipeline}
\end{figure}
These failures motivated a three-stage pipeline that (1)~decomposes golden solutions into reference steps, (2)~compresses CoT traces into clean logical steps via distillation, and (3)~classifies errors against a fixed global taxonomy. We describe each stage below and illustrate the pipeline in \Cref{fig:error_pipeline}.

\paragraph{Stage~1: Golden solution decomposition.}
For each problem, we use a strong analyzer model to decompose the verified golden solution into a sequence of logical steps, where each step represents a self-contained mathematical operation (e.g., writing down a Lagrangian, applying an identity, performing an integral). The final code implementation is separated and preserved as a distinct artifact. The analyzer receives the golden solution with line numbers and returns step boundaries, producing a structured reference against which model attempts can be compared. Because the analyzer only selects boundaries, it cannot distort the original text. Using the verified, concise golden solution as the reference for error analysis is essential, as it allows the analyzer to assess model reasoning without having to independently solve each problem.

\paragraph{Stage~2: CoT distillation.}
Raw CoT rollouts from reasoning models are verbose and contain substantial noise: repeated derivations, abandoned approaches, metacognitive filler (``Let me think\ldots'', ``Hmm\ldots''), and self-corrections. To address this, we first apply a \emph{deduplication} pass to rollouts exceeding a token threshold. The deduplication prompt instructs the analyzer to remove---without any rewriting or paraphrasing---repeated reasoning (keeping only the final version), abandoned approaches, metacognitive filler, self-doubt, and redundant restatements. Crucially, the output must be a strict subset of the original text, preserved verbatim.

The deduplicated trace is then distilled into logical steps, analogously to Stage~1. The analyzer identifies step boundaries throughout the \emph{entire} trace (not only the conclusion), extracting all mathematically substantive operations while ignoring exploratory text and self-corrections. As in Stage~1, any code implementation is separated from the mathematical reasoning. This two-phase process---deduplication followed by step distillation---reliably reduces \(10{,}000\) token CoT traces to a concise sequence of $5$--$15$ logical steps, resolving both the context-length limitation and the callback-noise problem that plagued our earlier approaches. We deliberately keep distillation \emph{purely subtractive}: the analyzer may only delete abandoned branches, self-corrections, and filler---never rewrite, summarize, or reorder---so that the retained steps remain verbatim model output. Prompting the model to instead \emph{summarize} the final successful derivation risks hallucinating or silently correcting the very reasoning we aim to diagnose. We did not run a formal ablation over distillation prompts, but this verbatim-subset constraint bounds their influence on the extracted steps, and our independent consistency check (Appendix \ref{app:consistency}) yields comparable aggregate error distributions.

\paragraph{Stage~3: Error classification with code comparison.}
For incorrect attempts, the analyzer model receives the distilled response and the entire decomposed golden solution, and is tasked with classifying any errors present.
We do not require the model to follow the same order or the identical set of steps as the golden solution; instead, the analyzer matches conceptually equivalent operations across the two sequences. This ensures that valid alternative derivation paths are not penalized, though it may occasionally fail to flag errors in highly non-standard solutions that reach the correct answer through unrecognized routes.
In addition, the analyzer receives an explicit \emph{code comparison} section that places the golden code implementation side-by-side with the model's code, with specific instructions to check for hardcoded values, missing parameters, and incorrect variable mappings. This dedicated comparison is essential: without it, code-level bugs are frequently overlooked because they are buried within lengthy step sequences.

\paragraph{Error Labels}
We define a fixed global set of four error labels:

\begin{itemize}
    \item \textbf{Factual}: Recalling incorrect physics facts, misreading the problem statement, or contradicting earlier derivation results.
    \item \textbf{Mathematical}: Incorrect mathematical operations---algebraic errors, sign mistakes, wrong simplifications, missing factors, integration/differentiation errors.
    \item \textbf{Logical}: Invalid deductions, non sequiturs, circular reasoning, or unjustified assumptions.
    \item \textbf{Executional}: Errors in code translation---incorrect symbolic-to-code mapping, hardcoded values instead of function parameters, or implementation bugs not present in the analytical reasoning (see \Cref{fig:error_examples}).
\end{itemize}
\begin{figure}[t!]
\centering
\small
\setlength{\fboxsep}{6pt}

\noindent\colorbox{red!8}{\begin{minipage}{0.96\linewidth}
\textbf{\color{red!70!black}\faTimesCircle\; Factual Error} \hfill \textit{Problem 899, CoT Step 5}\\[3pt]
\textbf{Model claim:} ``$(\gamma^5)^2 = -\mathbb{I}$''\\[2pt]
\textbf{Correct fact:} $\gamma^5$ squares to the identity, $(\gamma^5)^2 = \mathbb{I}$.\\[2pt]
{\footnotesize\color{gray}\textit{Misrecalled property of the chirality matrix propagates through the subsequent projection operator algebra.}}
\end{minipage}}

\vspace{6pt}

\noindent\colorbox{blue!8}{\begin{minipage}{0.96\linewidth}
\textbf{\color{blue!70!black}\faCalculator\; Mathematical Error} \hfill \textit{Problem 993, CoT Step 4}\\[3pt]
\textbf{Model writes:} $(2x-1)^2 = x^2 - 2x + 1$\\[2pt]
\textbf{Correct expansion:} $(2x-1)^2 = 4x^2 - 4x + 1$.\\[2pt]
{\footnotesize\color{gray}\textit{The factor of 4 is dropped from the leading coefficient, producing incorrect polynomial coefficients for the Feynman parameter integral.}}
\end{minipage}}

\vspace{6pt}

\noindent\colorbox{orange!10}{\begin{minipage}{0.96\linewidth}
\textbf{\color{orange!70!black}\faExclamationTriangle\; Logical Error} \hfill \textit{Problem 49, CoT Step 4}\\[3pt]
\textbf{Model derives:} Amplitude scales as $\mathcal{A}(z) \sim z^{2}$ at large $z$.\\[2pt]
\textbf{Model concludes:} ``\ldots therefore the amplitude scales as $z^{-2}$.''\\[2pt]
{\footnotesize\color{gray}\textit{The derivation is correct, but the conclusion contradicts it---inverting the sign of the exponent and reversing the large-$z$ behavior.}}
\end{minipage}}

\vspace{6pt}

\noindent\colorbox{green!8}{\begin{minipage}{0.96\linewidth}
\textbf{\color{green!50!black}\faCode\; Executional Error} \hfill \textit{Problem 1199, CoT Step 8}\\[3pt]
\textbf{Derivation:} Correctly obtains the BPS energy coefficient $C = \tfrac{1}{4}$.\\[2pt]
\textbf{Code:} \;\texttt{C = 1/3} \;\; {\color{red!70!black}$\boldsymbol{\leftarrow}$ \textbf{bug}}\\[2pt]
{\footnotesize\color{gray}\textit{The analytical reasoning is correct, but the implementation assigns a different numerical value, a transcription error visible only in the code.}}
\end{minipage}}

\caption{Representative examples of the four error categories identified by our pipeline. Each box shows an actual error from an incorrect rollout, illustrating how the category definitions map to concrete failure modes.}
\label{fig:error_examples}
\end{figure}

The analyzer identifies all errors in each incorrect rollout, classifies each into exactly one of the four categories, and assigns a severity level: \emph{major} (significantly affected reasoning or the final answer) or \emph{minor} (present but did not materially change the result). Each rollout also receives a \emph{primary error category}---the single category most directly responsible for the wrong final answer. A deterministic post-processing step normalizes any fine-grained labels returned by the analyzer (e.g., ``sign error'' $\to$ \texttt{mathematical}, ``code bug'' $\to$ \texttt{executional}) to ensure consistent aggregation.

\paragraph{Analyzer model and scale.}
We use OpenAI's open-weight \texttt{gpt-oss-120B} model as the analyzer across all stages. We selected this model after validating on a small subset of problems with Gemini-3-Pro, which produced comparable classification quality (we further comment on a consistency check performed with the Claude Code CLI in \Cref{app:consistency}); \texttt{gpt-oss-120B} offers substantially lower inference cost due to its mixture-of-experts architecture (117B total parameters, 5.1B active), making it practical to analyze hundreds of rollouts per problem. The pipeline processes all incorrect rollouts for the 20 problems with the largest mean@100 improvement under each training method: for RL, 1{,}378 base-model and 480 RL-fine-tuned incorrect rollouts; for SFT, 1{,}499 base-model and 493 SFT incorrect rollouts. The two top-20 sets share 10 overlapping problems, enabling direct three-way comparison on a common problem subset.

\begin{figure}[t!]
    \centering
    \includegraphics[width=0.75\linewidth]{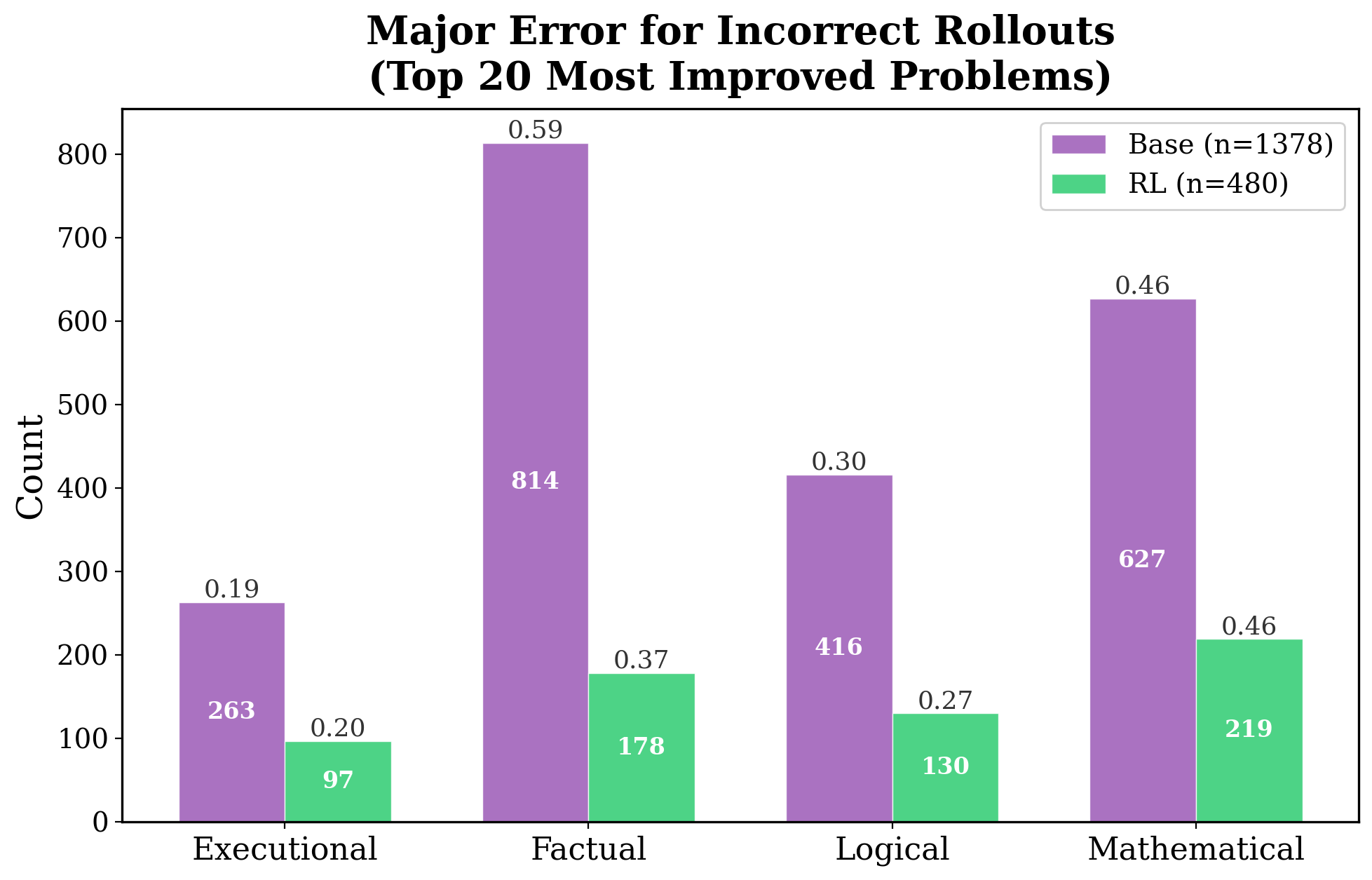}
    \caption{Major error counts and frequencies (average number of major errors per incorrect rollout) across the top-20 most RL-improved problems, broken down by error category.}
    \label{fig:major_error_frequency}
\end{figure}

\begin{figure}[t!]
    \centering
    \includegraphics[width=0.75\linewidth]{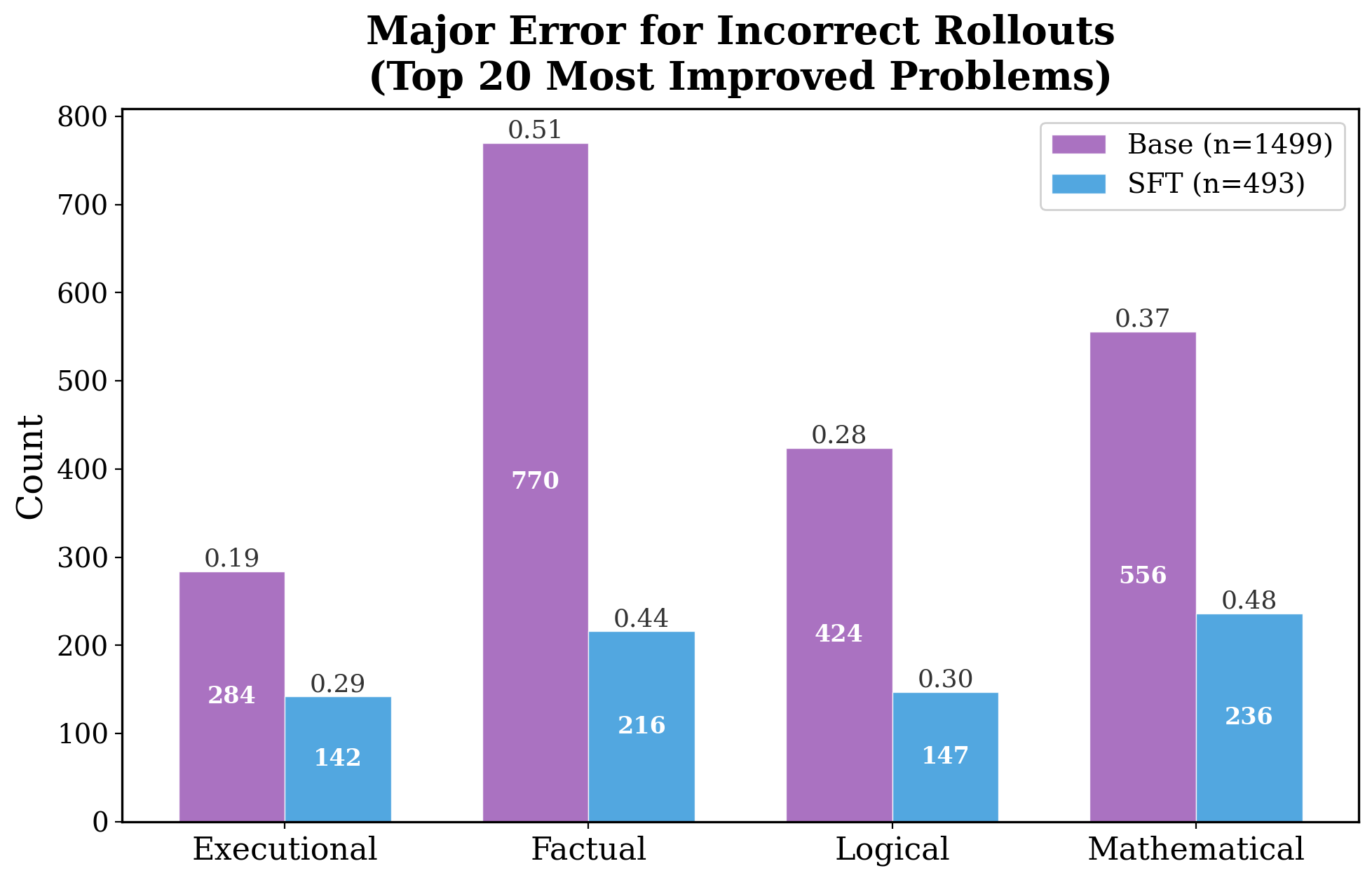}
    \caption{Major error counts and frequencies (average number of major errors per incorrect rollout) across the top-20 most SFT-improved problems, broken down by error category.}
    \label{fig:major_error_frequency_sft}
\end{figure}

\begin{figure}[t!]
    \centering
    \includegraphics[width=0.75\linewidth]{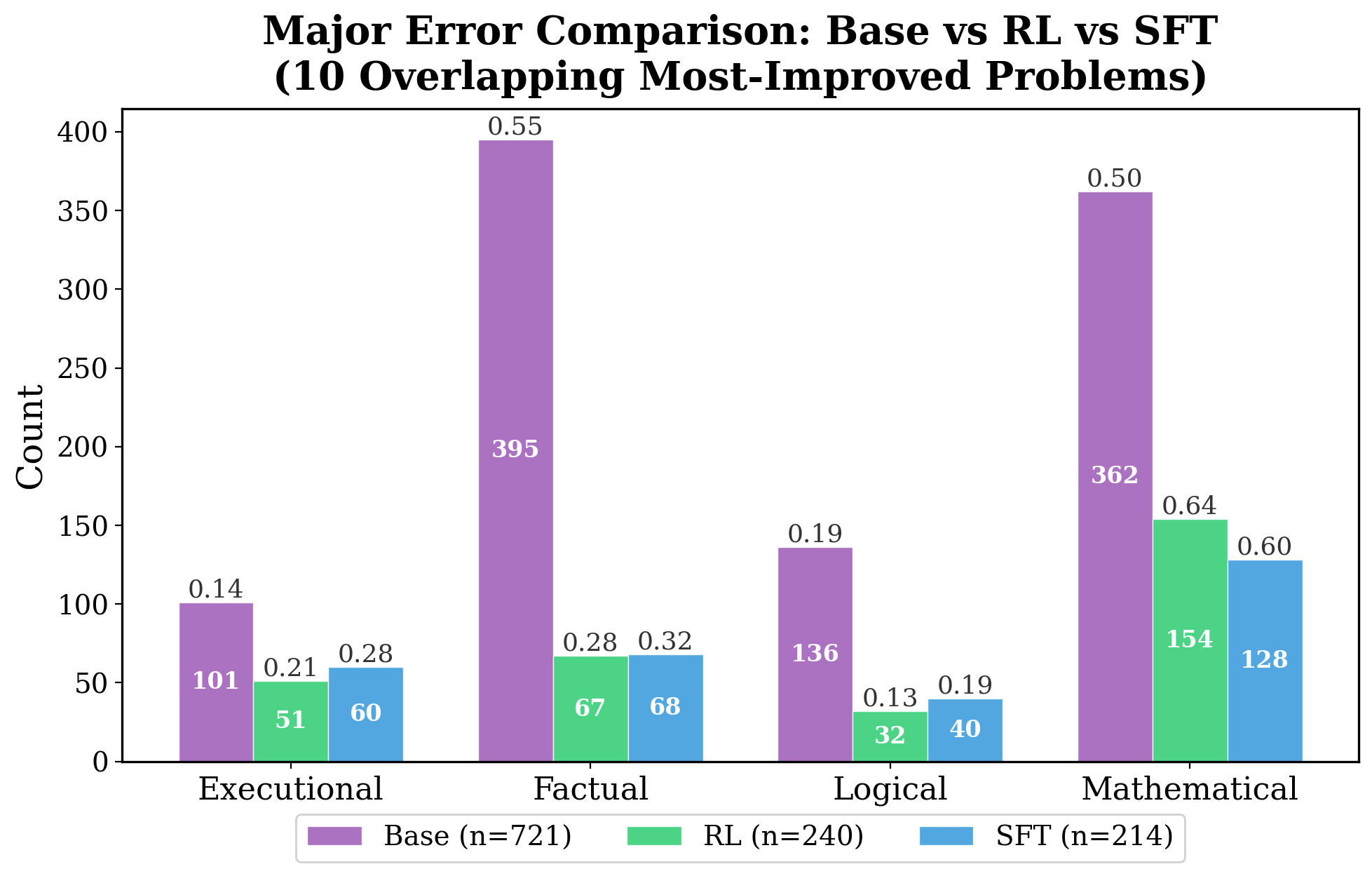}
    \caption{Three-way major error comparison on the 10 problems appearing in both the RL and SFT top-20 most-improved sets. All error categories decrease in absolute count. Per incorrect rollout, both fine-tuned models substantially reduce the factual error frequency (0.55 $\to$ 0.28 for RL, 0.32 for SFT), while mathematical errors constitute a larger share of the remaining incorrect attempts (0.50 $\to$ 0.64 for RL, 0.60 for SFT). Frequency is defined as the average number of major errors per incorrect rollout.}
    \label{fig:overlap_error_comparison}
\end{figure}

\subsection{Findings}

\paragraph{RL Error Statistics.} \Cref{fig:major_error_frequency} shows the frequency of major errors---measured as the average number of major errors per incorrect rollout---across the 20 most improved problems, for both the base \texttt{DeepSeek-7B} model and its RL-fine-tuned variant. RL reduces the total number of incorrect rollouts by 65\%, from 1{,}378 $\to$ 480, which explains the decrease in the absolute counts of all error types after fine-tuning.

The largest change is in \emph{factual errors}, which drop from 0.59 to 0.37 per incorrect rollout, a 37\% reduction in frequency. This is a notable finding: conventional wisdom holds that SFT is effective at teaching domain knowledge, while RL primarily improves reasoning and problem-solving strategy \cite{chu2025sftmemorizesrlgeneralizes, yue2025doesreinforcementlearningreally, wang2025tinatinyreasoningmodels}. In our experiments, RL training primarily grounded the model in correct domain knowledge, reducing misrecalled physics facts, problem misreadings, and internal contradictions, while \emph{mathematical}, \emph{executional}, and \emph{logical} errors remained at similar frequencies per incorrect rollout.

\paragraph{SFT Error Statistics.} \Cref{fig:major_error_frequency_sft} presents the analogous analysis for SFT. The SFT model reduces the total number of incorrect rollouts from 1{,}499 to 493 across its top-20 most improved problems---a 67\% reduction, compared to 65\% for RL (1{,}378 $\to$ 480). As with RL, this large reduction means that all four error categories decrease substantially in absolute count. The per-rollout error frequency profile is qualitatively similar to that of RL: factual errors show the largest frequency drop (0.51 $\to$ 0.44), while mathematical (0.37 $\to$ 0.48), executional (0.19 $\to$ 0.29), and logical (0.28 $\to$ 0.30) errors constitute a larger share of the remaining incorrect attempts. This pattern---factual errors being eliminated preferentially among incorrect rollouts---appears to be a general feature of fine-tuning on domain-specific data rather than an artifact of the RL reward signal, reinforcing the finding that domain-specific fine-tuning improves factual grounding more than mathematical or logical reasoning.

\begin{figure}[t]
    \centering
    \includegraphics[width=\linewidth]{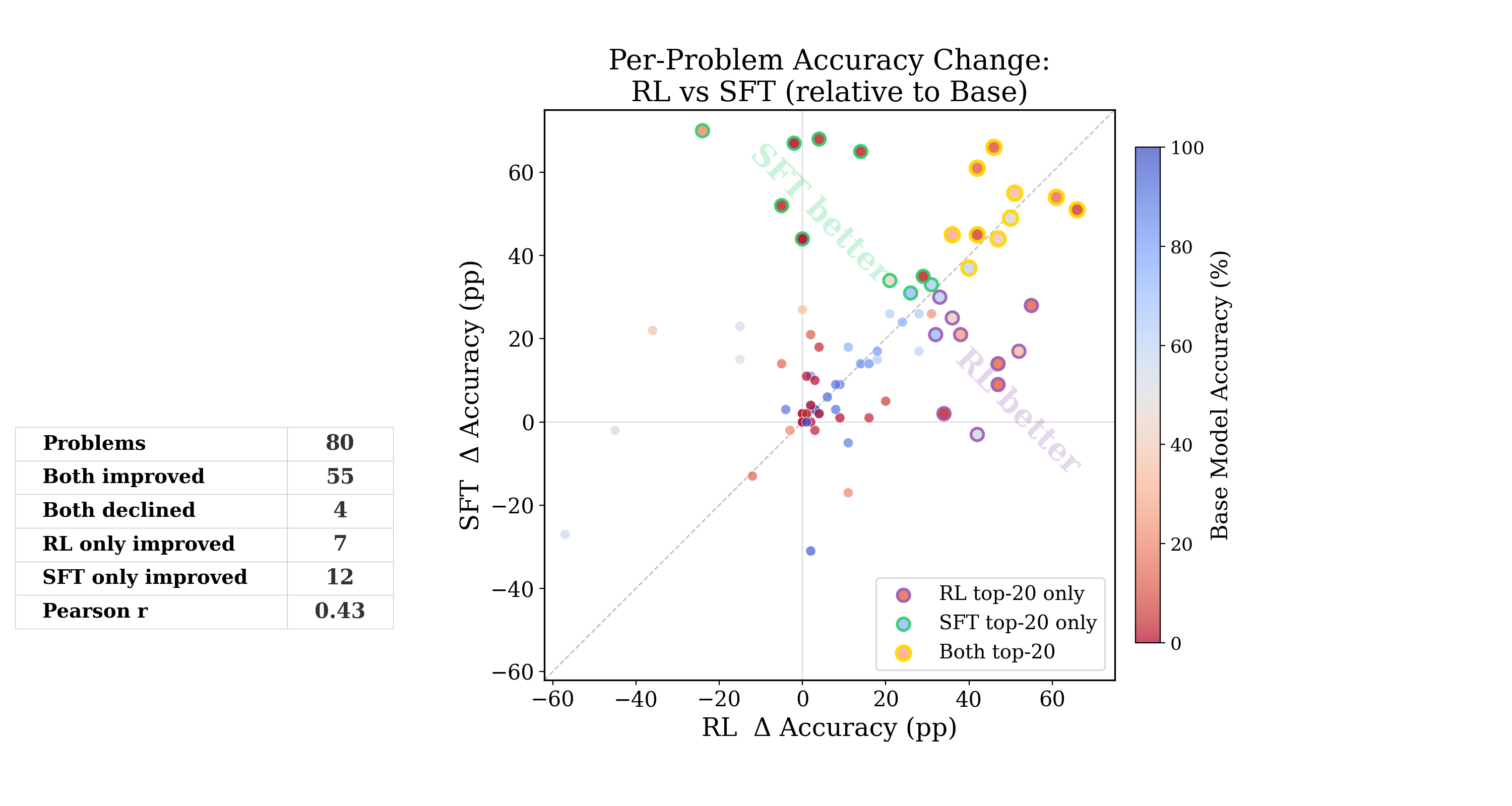}
    \caption{Per-problem accuracy change (percentage points) for the RL-fine-tuned and SFT models, relative to the base model. Each point is one of 80 problems; color encodes base model accuracy. Edge highlights mark the top-20 most improved problems for each method. The moderate correlation (Pearson $r = 0.43$) indicates partially overlapping but distinct improvement profiles, with 10 of the 20 top-improved problems shared between RL and SFT.}
    \label{fig:rl_vs_sft_scatter}
\end{figure}

\paragraph{Direct RL vs.\ SFT Comparison on Overlapping Problems.} The 10 problems appearing in both top-20 sets (\Cref{fig:overlap_error_comparison}) allow a controlled three-way comparison. On these problems, the base model produces 721 incorrect rollouts; RL reduces this to 240 (67\% reduction) and SFT to 214 (70\% reduction). All four error categories decrease in absolute count under both methods. Both methods achieve comparable reductions in factual error frequency per incorrect rollout (base 0.55 $\to$ RL 0.28, SFT 0.32), again confirming that improved factual grounding is the primary mechanism. The residual errors in both fine-tuned models are dominated by mathematical mistakes (RL 0.64, SFT 0.60 per incorrect rollout), suggesting that algebraic manipulation remains the principal bottleneck for the remaining incorrect attempts. This compositional shift---in which factual errors are preferentially resolved, leaving the remaining incorrect rollouts concentrated on harder problems dominated by mathematical and implementation errors---is consistent across both training methods.

The moderate correlation between per-problem improvements (Pearson $r = 0.43$; \Cref{fig:rl_vs_sft_scatter}) indicates that RL and SFT partially target the same failure modes but also address distinct weaknesses. The 50\% overlap (10 of 20 problems) in the top-20 sets---twice the ${\sim}25\%$ (5 problems) expected for two random 20-of-80 subsets, and highly unlikely to arise by chance ($P(\text{overlap} \ge 10) < 0.5\%$)---confirms that both methods preferentially improve similar problems, while the remaining divergence suggests that RL's reward-driven optimization and SFT's imitation learning provide complementary inductive biases.

\paragraph{Backtracking and Trace Length Analysis.}
To quantify how fine-tuning affects the model's reasoning behavior, we analyze two properties of each chain-of-thought (CoT) trace: \emph{trace length}, measured in number of tokens, and \emph{backtracking frequency}, defined as the number of explicit self-correction events per trace. We identify backtracking through 13 regex patterns that capture strictly corrective language (e.g., ``Wait, no'', ``Let me recalculate'', ``I made a mistake''), excluding generic discourse markers such as ``Hmm'' or ``Actually'' that do not signal a reversal of prior reasoning. These patterns were validated through manual inspection of over 400 traces across both models. We evaluate 8{,}000 traces per model (80 problems $\times$ 100 attempts), with correctness determined by an automated numerical verifier that checks the final answer against the reference solution.

\begin{figure}[H]
    \centering
    \includegraphics[width=\textwidth]{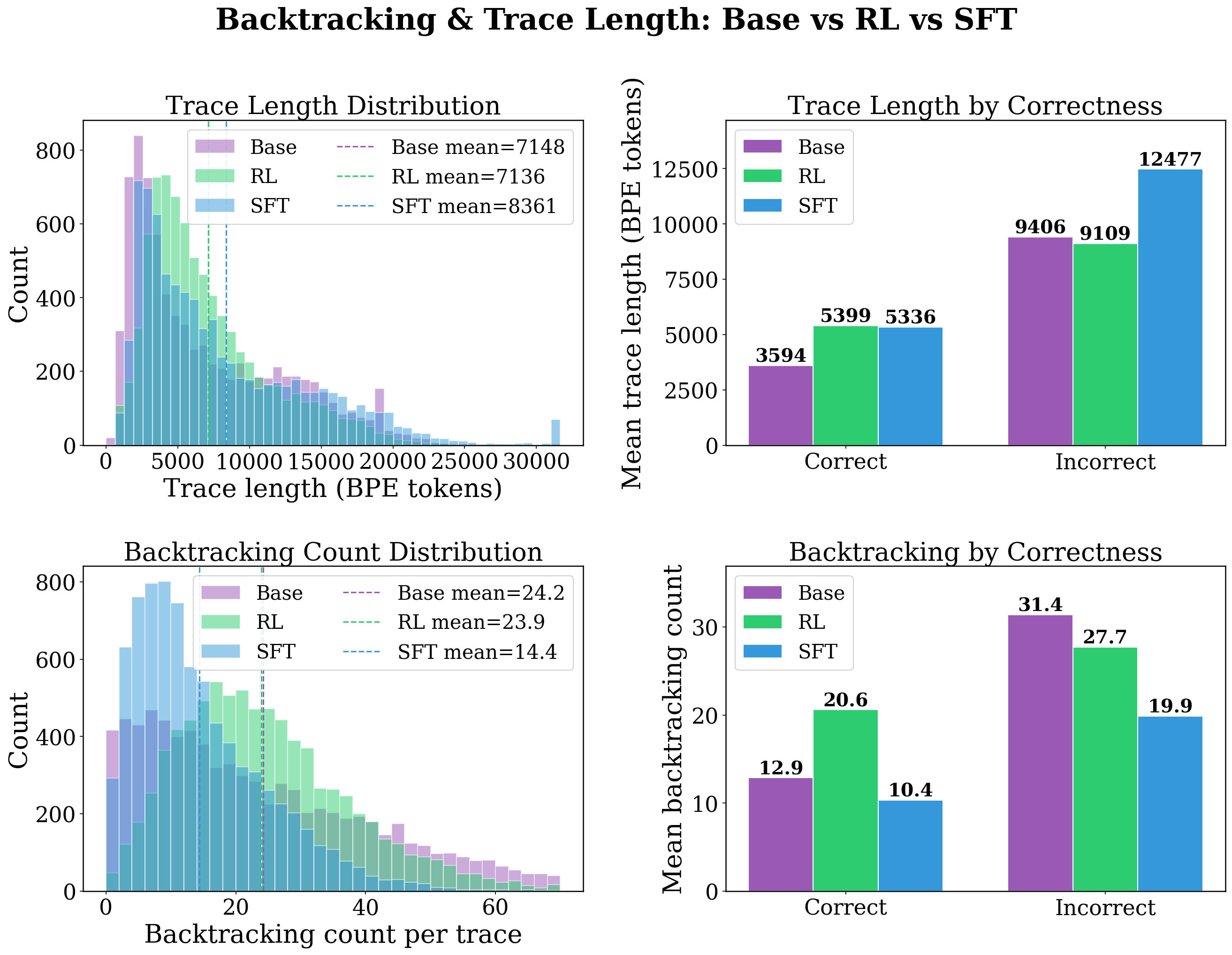}
\caption{Trace length and backtracking frequency for the base, RL-fine-tuned, and SFT models (80 problems, 100 attempts each). Note that SFT traces are notably longer on average, likely because the model learns to mimic the verbosity of the fine-tuning demonstrations rather than reflecting a genuine increase in reasoning depth.}
    \label{fig:backtracking}
\end{figure}

While the base and RL-fine-tuned models exhibit nearly identical aggregate trace lengths (7{,}148 vs.\ 7{,}136 BPE tokens) and backtracking frequencies (24.2 vs.\ 23.9 per trace), conditioning on correctness reveals a redistribution effect (\Cref{fig:backtracking}). The RL model generates substantially longer traces on correct attempts (5{,}399 vs.\ 3{,}594 tokens) and backtracks more frequently when it ultimately succeeds (20.6 vs.\ 12.9), suggesting that RL training encourages the model to invest additional computation---including productive self-correction---on problems it can solve. Conversely, on incorrect attempts the RL model backtracks less (27.7 vs.\ 31.4), indicating a reduction in futile self-correction loops. RL thus neither increases nor decreases backtracking overall, but reallocates it toward problems the model is able to solve.

The SFT model exhibits a distinct pattern. Its traces are the longest on average (8{,}361 vs.\ 7{,}148 tokens for the base), driven largely by inflated incorrect-attempt traces (12{,}477 vs.\ 9{,}406), consistent with the model imitating the verbosity of its fine-tuning demonstrations rather than reasoning more deeply. Unlike RL's redistribution of backtracking, SFT suppresses self-correction across the board: it backtracks less than the base model on both correct (10.4 vs.\ 12.9) and incorrect (19.9 vs.\ 31.4) attempts, yielding a lower overall backtracking frequency (14.4 vs.\ 24.2 per trace).

We note that, because we study open-weight models, we analyze the \emph{complete} generated chain-of-thought, with no provider-side summarization or hidden reasoning tokens, as can occur with proprietary APIs. Faithfulness is therefore less of a concern in our setting than for closed-source models. While a visible trace still need not perfectly mirror the model's internal computation, our error-type and backtracking findings are best read as robust statistical trends in the generated text rather than exact accounts of latent reasoning.

\subsection{Impact of Domain and Operational Difficulty on Model Performance}
\label{sec:difficulty}

\begin{figure}[H]
    \centering
    \includegraphics[width=\textwidth]{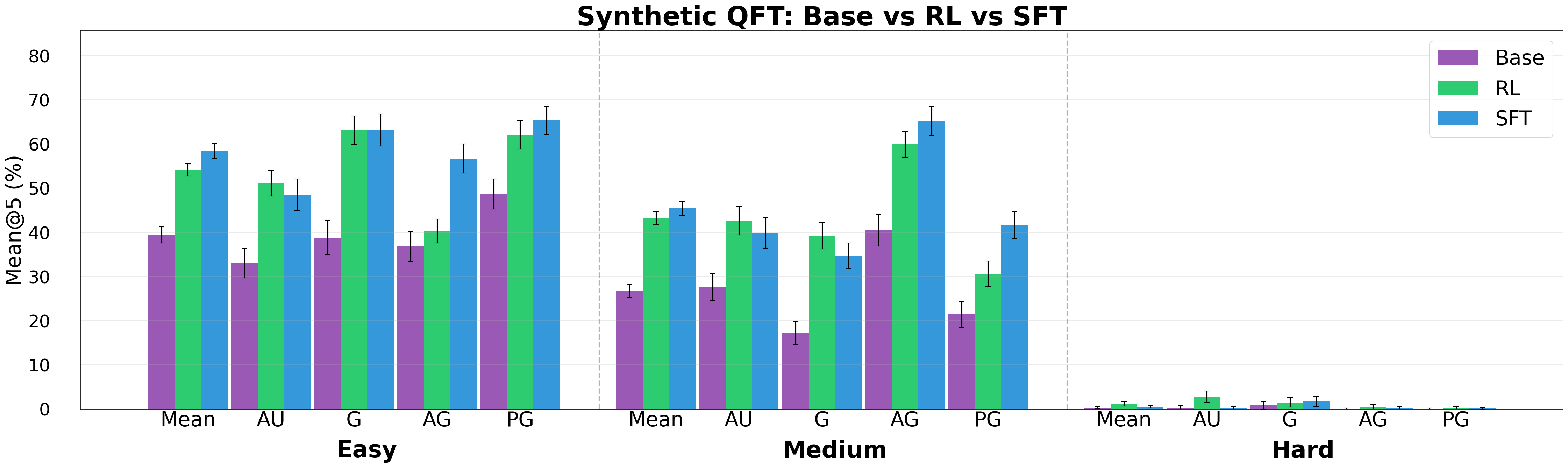}
    \caption{Mean@5 validation performance by domain level (defined by pedagogical level) for Easy, Medium, and Hard QFT. We abbreviate advanced undergraduate (AU), graduate (G), advanced graduate (AG), and postgraduate (PG). Accuracy is non-monotonic in domain difficulty, indicating that knowledge level alone is a poor predictor of LLM performance. Operational difficulty is the primary driver of LLM solve rate, with higher operational difficulty consistently yielding lower mean accuracy across all three models. For a detailed breakdown of performance, please refer to Appendix \ref{app:all_results}.}
    \label{fig:accuracy_difficulty}
\end{figure}

Expanding our analysis to the Easy, Medium, and Hard QFT validation sets, we use 100 rollouts per problem and examine the relation between LLM solve rate and both domain and operational difficulty. As shown in \Cref{fig:accuracy_difficulty}, there is no consistent correlation between accuracy and domain difficulty. In Easy QFT, all three models solve postgraduate problems more reliably ($48.7\pm3.4$\% base, $62.0\pm3.2$\% RL, $65.3\pm3.2$\% SFT) than advanced undergraduate ones ($33.0\pm3.3$\% base, $51.1\pm2.9$\% RL, $48.5\pm3.6$\% SFT). Similarly, in Medium QFT, advanced graduate problems are the easiest across all three models ($40.5\pm3.6$\% base, $59.9\pm2.9$\% RL, $65.2\pm3.3$\% SFT). Their weakest areas, however, diverge: the base and SFT models struggle the most with graduate problems ($17.2\pm2.6$\% and $34.7\pm2.9$\%, respectively), while the RL model struggles most with postgraduate problems ($30.6\pm2.9$\%). Performance on Hard QFT remains too low to draw concrete conclusions about any trends.

These trends suggest that LLM solve rates in our datasets are governed by operational complexity rather than domain difficulty. The primary bottleneck is not the depth of physics knowledge, but the length of the reasoning chain---specifically the number of derivation steps, intermediate manipulations, and implicit constraints that must be identified. A problem requiring advanced knowledge but admitting a direct derivation is often easier than a conceptually simpler problem demanding sustained multi-step reasoning. This is exactly the case for some of the postgraduate problems in the Easy QFT dataset (discussed in \Cref{sec:exp1} and demonstrated in \Cref{sec:ex_prob_easy_pg}).

To quantify operational difficulty independently of the categorical Easy/Medium/Hard label, we measure the length of the golden solution (a proxy for the number of derivation steps) (\Cref{tab:operational_difficulty}).
Golden-solution and derivation lengths increase monotonically along the intended operational gradient and track base-model accuracy, yet remain essentially flat across domain levels within each dataset. For example, the mean golden-solution length across advanced-undergraduate/graduate/advanced-graduate/postgraduate topics is $3{,}003$, $2{,}758$, $2{,}605$, and $2{,}342$ characters in Easy. This provides direct, non-categorical evidence that operational complexity---not the depth of required physics knowledge---is the primary driver of LLM solve rate in our datasets.

\begin{table}[H]
\centering
\caption{\textbf{Independent indicators of \emph{operational} difficulty.} Mean golden-solution length, derivation length (nonempty lines), and base-model accuracy---scale consistently with the intended Easy, Medium, and Hard difficulty tiers. Interestingly, within each dataset, none of these metrics shows a similar monotonic trend when broken down by \emph{domain} level (\Cref{fig:accuracy_difficulty}).}
\label{tab:operational_difficulty}
\small
\begin{tabular}{lccc}
\toprule
Dataset & Golden sol.\ (chars) & Derivation (lines) & Base acc. \\
\midrule
Easy   & $2{,}677$ & $35.8$ &  $39.4\pm1.8\%$ \\
Medium & $3{,}329$ & $47.5$ & $26.7\pm1.5\%$ \\
Hard   & $4{,}638$ & $74.9$ & $0.3\pm0.2\%$  \\
\bottomrule
\end{tabular}
\end{table}

\section{Conclusion}
In this work, we explored the fine-tuning of LLMs for reasoning tasks in theoretical physics. Because of its high computational cost, fine-tuning for advanced reasoning topics has received relatively little academic attention (most work focuses on elementary mathematics), and less still in theoretical physics \cite{Richmond_2026,Cai__2025}. Although academic computing constraints make it impossible to achieve results competitive with industry models, it is nonetheless interesting to study the training behavior of smaller reasoning models.

\paragraph{Results}
In our exploratory training runs, we demonstrated consistent gains in \texttt{DeepSeek-7B}'s theoretical physics reasoning. On the synthetic QFT datasets, mean accuracy on Easy QFT improved from \resEasyMeanKBase\% to \resEasyMeanKRL\% through RL, and to \resEasyMeanKSFT\% through SFT on \texttt{Qwen3-30B-A3B} rejection-sampled data. This training transferred strongly to Medium QFT, where accuracy increased from the baseline of \resMediumMeanKBase\% to \resMediumMeanKRL\% (RL) and \resMediumMeanKSFT\% (SFT), though the Hard QFT dataset remains largely unsolved by both the base and the fine-tuned models. Crucially, while SFT performed best on the synthetic tasks, RL generalized better to human-adapted, out-of-distribution domains. On QFT Pedagogy, for instance, the RL model improved accuracy from \resQFTMeanKBase\% to \resQFTMeanKRL\% (outperforming the SFT model's \resQFTMeanKSFT\%), and it similarly boosted arXiv performance from \resArXivMeanKBase\% to \resArXivMeanKRL\% (compared to \resArXivMeanKSFT\% for SFT). This pattern held on TPBench, where RL achieved a \resTPBenchMeanKRL\% success rate, compared with the \resTPBenchMeanKBase\% baseline, narrowly beating the SFT model's \resTPBenchMeanKSFT\%.

To understand how these models' reasoning behaviors evolved after fine-tuning, we performed a robust CoT analysis using our \textit{Distill-then-Classify} pipeline. Evaluating incorrect attempts on the 20 most-improved problems, we found that RL and SFT reduced the number of incorrect rollouts by 65\% and 67\%, respectively. For both methods, factual errors saw the greatest reduction, while mathematical errors remained the most prevalent after fine-tuning, highlighting the probable benefits of tool-use integration.

Expanding our analysis to response length and backtracking frequency revealed differences between RL and SFT. Although both methods increased the length of correct responses, only SFT inflated the length of incorrect traces, which we attribute to teacher verbosity. Furthermore, while SFT uniformly suppressed backtracking, RL actively increased it during successful problem-solving, with a slight reduction on incorrect attempts. This suggests that SFT disrupts and overrides the base model's inherent reasoning and self-correction patterns to a greater extent than RL.

Comparing performance trends across the three synthetic datasets, we found that for both the base and the fine-tuned models the solve rate is primarily driven by the operational complexity of the tasks. Domain complexity---the depth of specialized physics knowledge required---exhibits little correlation with performance in our datasets, perhaps because the models' broad knowledge base covers all levels of difficulty relatively evenly.

\paragraph{Fine-tuning in Academia}
Although the two methods yield comparable performance, RL is significantly more compute-intensive than SFT. All compute figures reported below refer to runs conducted on a single node equipped with \(4\times\) NVIDIA H200 GPUs. 
Specifically, the RL pipeline for Easy QFT required 640 GPU-hours. In contrast, the entire SFT dataset generation required 144 GPU-hours, and SFT of \texttt{DeepSeek-7B} on Easy QFT traces generated by \texttt{Qwen3-30B-A3B} took only 5 GPU-hours. In total, the development of this work, including all exploratory runs, failed attempts, and final training runs, consumed approximately 240 GPU-days.

RL and SFT have different advantages and disadvantages for exploring TP reasoning in academic settings. We view SFT as an efficient method for developing reasoning on tasks \textit{within} frontier model capabilities. However, because SFT relies on distillation from a teacher to a student, it may be unappealing for researchers aiming to study reasoning progression purely through the optimization of problem-solving tasks. Furthermore, for topics near the limits of model ability, there may not even be a suitable teacher model available to generate SFT data---a challenge currently exacerbated by the lack of full CoT traces from proprietary frontier models.

For academic fine-tuning, RL can entail prohibitive computational costs. Standard academic compute budgets currently cannot support large-scale experiments, which largely rules out extending model capabilities solely through prolonged, outcome-based RL. Despite this constraint, RL remains appealing because it offers researchers greater control to directly guide a model's underlying reasoning behavior.
Academic studies of reasoning in TP can aim to move beyond simple RLVR by focusing on the development of reward functions, integration with symbolic computation tools, and structured, multi-step reasoning loops tailored to research-level tasks.

\paragraph{Outlook}
This work establishes baseline results for model fine-tuning in TP. To aid further work, we provide our verifiable QFT training data and our data pipeline for synthetic and human-adapted problems, and we release all SFT rejection-sampled datasets. Future work may focus on improving training or reliability, for example, by using multi-step RL \cite{guan2025rstarmathsmallllmsmaster,goldie2025syntheticdatageneration}, integrating computational tools \cite{goldie2025syntheticdatageneration,singh2025agenticreasoningtoolintegration, gou2024toratoolintegratedreasoningagent, shang2025rstar2agentagenticreasoningtechnical}, or applying RL with symbolic verification \cite{gao2025testtimescalingtechniquestheoretical}. Further, with new investments in academic compute (such as Argonne's Solstice system with 100,000 NVIDIA Blackwell GPUs), larger academic experiments are becoming possible in principle.

\section{Acknowledgments}
We are grateful for API credits from the Google Cloud research credits program and Anthropic, allowing for the generation and evaluation of the synthetic and human-adapted datasets. All local training and evaluation runs were only made possible through the generous access to the Symmetry H200 cluster at the Perimeter Institute for Theoretical Physics. We thank Garrett Merz, Maja Waldron, and Jianhao Wu for useful discussion in the development of this work. M.M. acknowledges the support by the U.S. Department of Energy, Office of Science, Office of High Energy Physics under Award Number DE-SC0017647, the support by the National Science Foundation (NSF) under Grant Number 2307109 and 2509873 and the Wisconsin Alumni Research Foundation (WARF). KMS was supported by an NSERC Discovery Grant, by the Daniel Family Foundation, and by the Centre for the Universe at Perimeter Institute. Research at Perimeter Institute is supported by the Government of Canada through Industry Canada and by the Province of Ontario through the Ministry of Research \& Innovation. FS is grateful for the support of the NSF under CCF2106707 and the Wisconsin Alumni Research Foundation (WARF).

\section{Data Access}
To support further academic exploration in LLM reasoning in theoretical physics, we release our data pipeline (\url{https://github.com/nswood/VerifiableTPData}) as well as our full synthetic QFT datasets and rejection-sampled CoT datasets (\url{https://huggingface.co/datasets/nswoodward/VerifiableQFT}). 
To respect publisher copyrights and the intellectual property of the textbook, exercise, and coursework authors, the QFT pedagogy dataset will not be made publicly available. We anticipate releasing the arXiv-derived dataset in future work, pending the completion of ongoing improvements to our extraction and verification pipeline.

\printbibliography

\appendix

\section{QFT Topic List}
\label{sec:qft_topics}

\subsection*{Advanced Undergraduate}

\small

\subsubsection*{Foundations and Scalar Fields}
\begin{itemize}
  \item \textbf{AU-01: Classical field theory and Lagrangian formalism} --- Build field theories from symmetry principles using actions and Euler--Lagrange equations. Connect kinetic and interaction terms to conservation laws and dimensions.
  \item \textbf{AU-02: Euler--Lagrange equations for fields} --- Derive equations of motion for scalar or vector fields, including derivative couplings. Track overall signs and coefficients carefully and interpret terms physically.
  \item \textbf{AU-03: Symmetries and Noether's theorem} --- Relate continuous symmetries to conserved currents and charges. Distinguish internal vs.\ spacetime symmetries in simple models and extract explicit currents.
  \item \textbf{AU-04: Energy--momentum tensor (canonical vs.\ improved)} --- Construct stress tensors and test conservation. Examine traceless limits and symmetry under Lorentz transformations.
  \item \textbf{AU-05: Lorentz invariance and field representations} --- Classify fields by spin and transformation properties. Verify covariance of actions and bilinears using indices or gamma-matrix conventions at a conceptual level.
  \item \textbf{AU-06: Quantization of the Klein--Gordon field} --- Promote classical fields to operators, impose commutators, and define creation/annihilation operators. Connect normalizations to delta functions and box quantization.
  \item \textbf{AU-07: Plane-wave solutions and dispersion} --- Analyze Fourier modes, on-shell relations, and normalization for free relativistic fields. Relate to physical energies and group velocity.
  \item \textbf{AU-08: Microcausality and equal-time commutators} --- Compute spacelike (anti)commutators and identify conditions for microcausality. Clarify locality assumptions.
\end{itemize}

\subsubsection*{Path Integrals and Quantization}
\begin{itemize}
  \item \textbf{AU-09: Path integrals: $Z[J]$ and $W[J]$} --- Formulate generating functionals for free theories and relate functional derivatives to correlation functions. Track signs and normalization.
  \item \textbf{AU-10: Gaussian functional integrals} --- Evaluate quadratic path integrals and determinants. Connect to propagators through completing the square and inverses of operators.
  \item \textbf{AU-11: Connected vs.\ 1PI generating functionals} --- Relate $Z$, $W$, and $\Gamma$. Distinguish connected, amputated, and 1PI objects and their roles in perturbation theory.
  \item \textbf{AU-12: Interaction picture and Dyson series} --- Organize time evolution with interactions and set up the Dyson expansion. Use interaction-picture operators and time ordering consistently.
  \item \textbf{AU-13: Propagators and Green's functions} --- Construct time-ordered, retarded, and advanced propagators in $x$- and $p$-space. Emphasize boundary conditions and mass dependence.
  \item \textbf{AU-14: Wick's theorem and contractions} --- Reduce time-ordered products to contractions and normal-ordered terms. Connect operator algebra to diagrammatics.
\end{itemize} 

\subsubsection*{Symmetries and Scattering}
\begin{itemize}
  \item \textbf{AU-15: Tree-level perturbation theory} --- Generate Feynman rules and compute elementary tree amplitudes. Track symmetry factors and identical-particle combinatorics.
  \item \textbf{AU-16: $\phi^4$ theory at tree level} --- Study $2\!\to\!2$ scalar scattering, channels, and basic kinematics. Compare identical vs.\ distinguishable cases and symmetry factors.
  \item \textbf{AU-17: Cross sections and Lorentz-invariant phase space} --- Relate matrix elements to observables using flux and LI phase space. Explore threshold behavior and high-energy limits.
  \item \textbf{AU-18: LSZ reduction (conceptual)} --- Connect amputated Green's functions to S-matrix elements and clarify wavefunction renormalization. Focus on which factors appear and why.
  \item \textbf{AU-19: Real vs.\ complex scalar fields} --- Contrast $U(1)$ symmetry, charge, and particle content. Derive the Noether current for complex fields.
  \item \textbf{AU-20: Discrete symmetries (P, C, T) for scalars} --- Classify simple operators under parity, charge conjugation, and time reversal. Provide selection rules in basic processes.
  \item \textbf{AU-21: Partial-wave expansion (intro)} --- Decompose simple scattering amplitudes into partial waves. Identify $s$- and $p$-wave dominance near threshold.
  \item \textbf{AU-22: Optical theorem at tree-level accuracy} --- Relate forward amplitude to total cross section in simple models. Check unitarity constraints qualitatively.
  \item \textbf{AU-23: K\"{a}ll\'{e}n function and kinematics} --- Use the K\"{a}ll\'{e}n $\lambda$ function to handle two-body kinematics. Analyze allowed regions and special limits.
  \item \textbf{AU-24: Dimensional analysis and natural units} --- Track mass dimensions of fields and couplings. Use natural units to check consistency of results.
  \item \textbf{AU-25: Stability and boundedness of potentials} --- Identify conditions for vacuum stability in polynomial scalar potentials. Classify phases and small oscillations.
\end{itemize}

\subsection*{Graduate}

\subsubsection*{Fermions and Spinor Structure}
\begin{itemize}
  \item \textbf{GR-01: Dirac equation and spinor solutions} --- Solve plane-wave spinors and completeness relations; compute bilinears and currents. Interpret helicity and chirality in simple processes.
  \item \textbf{GR-02: Quantized Dirac field} --- Build fermionic Fock space from anti-commutation relations. Connect field operators to particle number and charges.
  \item \textbf{GR-03: Gamma matrices, traces, and bilinears} --- Use Clifford algebra identities and trace technology. Classify bilinears by Lorentz and discrete symmetries.
  \item \textbf{GR-04: Chirality, helicity, and projection operators} --- Work with chiral projectors, massless limits, and spin sums. Apply to scattering and symmetry analyses.
  \item \textbf{GR-05: Discrete symmetries for fermions} --- Classify fermionic operators under P, C, and T. Provide selection rules and simple examples.
\end{itemize}

\subsubsection*{Gauge Fields and Symmetry}
\begin{itemize}
  \item \textbf{GR-06: QED and Abelian gauge invariance} --- Derive local $U(1)$ invariance, Feynman rules, and Ward identities. Examine gauge choices and physical observables.
  \item \textbf{GR-07: Yang--Mills theories and color structure} --- Formulate non-Abelian field strengths and covariant derivatives. Understand self-interactions and color factors.
  \item \textbf{GR-08: Gauge fixing and Faddeev--Popov ghosts} --- Quantize non-Abelian theories in covariant gauges. Derive ghost terms, vertices, and their role in loops.
  \item \textbf{GR-09: Ward--Takahashi identities} --- Relate symmetries to Green's function identities. Constrain renormalization and amplitude structures.
  \item \textbf{GR-10: Background vs.\ $R_\xi$ gauges (conceptual)} --- Compare gauge-fixing strategies and their impact on intermediate quantities. Emphasize gauge-parameter independence of observables.
\end{itemize}

\subsubsection*{Renormalization and Quantum Corrections}
\begin{itemize}
  \item \textbf{GR-11: Vacuum polarization} --- Analyze photon self-energy at one loop and momentum dependence of propagators. Interpret charge screening in QED.
  \item \textbf{GR-12: Dimensional regularization} --- Regulate loop integrals in $d=4-2\epsilon$. Track poles, scales, and scheme conventions (MS/$\overline{\text{MS}}$).
  \item \textbf{GR-13: Counterterms and renormalized parameters} --- Define renormalization conditions, Z-factors, and parameter redefinitions. Compare on-shell vs.\ minimal schemes.
  \item \textbf{GR-14: Running couplings and $\beta$ functions (QED)} --- Derive coupling evolution with scale and interpret effective charge. Connect to physical cross sections.
  \item \textbf{GR-15: Renormalization of $\phi^4$ theory (one loop)} --- Compute 2- and 4-point divergences and extract Z-factors. Discuss coupling running and triviality qualitatively.
\end{itemize}

\subsubsection*{Spontaneous Symmetry Breaking and Higgs}
\begin{itemize}
  \item \textbf{GR-16: Goldstone theorem} --- Connect continuous symmetry breaking to massless modes via current algebra. Provide simple examples.
  \item \textbf{GR-17: Higgs mechanism (Abelian)} --- Show gauge boson mass generation via SSB and gauge fixing. Track eaten modes and physical spectra.
  \item \textbf{GR-18: Higgs mechanism (non-Abelian, conceptual)} --- Extend to SU(2)-like settings with multiple gauge bosons. Emphasize counting of degrees of freedom and masses.
  \item \textbf{GR-19: Unitarity and the optical theorem} --- Link forward scattering to total cross sections. Verify S-matrix consistency at tree or one-loop level in simple channels.
  \item \textbf{GR-20: Infrared divergences in QED (inclusive observables)} --- Understand soft/collinear singularities and their cancellation in inclusive rates. Use soft factors and KLN intuition.
\end{itemize}

\subsubsection*{Applied Topics and Tools}
\begin{itemize}
  \item \textbf{GR-21: Spectral representation (K\"{a}ll\'{e}n--Lehmann, intro)} --- Express two-point functions via spectral densities and discuss positivity/normalization constraints.
  \item \textbf{GR-22: Dispersion relations (intro)} --- Relate real/imaginary parts of amplitudes through analyticity and subtractions. Apply to simple correlators.
  \item \textbf{GR-23: Partial waves and unitarity bounds} --- Use partial-wave unitarity to constrain couplings or cross sections in toy models.
  \item \textbf{GR-24: Crossing symmetry basics} --- Relate $s$-, $t$-, and $u$-channel processes through analytic continuation for simple amplitudes.
  \item \textbf{GR-25: LSZ details (applied)} --- Connect field renormalization factors to residues and asymptotic states for calculable examples.
\end{itemize}

\subsection*{Advanced Graduate}

\subsubsection*{Renormalization Group and Scaling}
\begin{itemize}
  \item \textbf{AG-01: RG flow and fixed points} --- Formulate RG equations and classify fixed points. Linearize flows and extract simple critical exponents.
  \item \textbf{AG-02: $\beta$ functions in $\phi^4$ and QED} --- Compute one-loop $\beta$-functions with attention to factors and schemes. Interpret UV/IR behavior and triviality vs.\ asymptotic freedom.
  \item \textbf{AG-03: Anomalous dimensions} --- Determine field/operator rescaling with scale and relate to correlation function scaling.
  \item \textbf{AG-04: Dimensional transmutation and $\Lambda$ scales} --- Explain how dimensionless couplings generate a physical scale through renormalization; connect to running couplings.
  \item \textbf{AG-05: Callan--Symanzik equation (intro)} --- Relate RG flow to scale dependence of correlators and Green's functions. Solve simple CS equations.
\end{itemize}

\subsubsection*{Non-Abelian Gauge Theories}
\begin{itemize}
  \item \textbf{AG-06: Non-Abelian renormalization and QCD $\beta$ function} --- Evaluate color factors and derive asymptotic freedom for SU(N). Discuss $n_f$-dependence and scheme issues.
  \item \textbf{AG-07: Background field method (one-loop effective action)} --- Preserve gauge invariance in effective actions and streamline RG calculations. Emphasize background vs.\ quantum fields.
  \item \textbf{AG-08: Wilson loops and area law (qualitative)} --- Use Wilson loops to diagnose confinement. Contrast area vs.\ perimeter behaviors and simple models.
  \item \textbf{AG-09: BRST symmetry (intro)} --- Introduce BRST transformations and cohomology ideas. Connect to gauge fixing and ghost structure.
\end{itemize}

\subsubsection*{Finite Temperature and Vacuum Effects}
\begin{itemize}
  \item \textbf{AG-10: Matsubara formalism} --- Use imaginary time and periodicity to compute thermal correlators. Perform simple Matsubara sums.
  \item \textbf{AG-11: Thermal effective potential (intro)} --- Study symmetry restoration/breaking at finite temperature using simple scalar models and high-$T$ expansions.
  \item \textbf{AG-12: Casimir effect (calculable setups)} --- Evaluate vacuum energy shifts due to boundaries in simple geometries. Interpret forces and scaling.
\end{itemize}

\subsubsection*{Anomalies and Supersymmetry}
\begin{itemize}
  \item \textbf{AG-13: Chiral anomalies (triangle diagrams)} --- Compute axial anomalies and understand current non-conservation. State cancellation conditions.
  \item \textbf{AG-14: Gauge anomalies (consistency)} --- Identify gauge anomaly structures and cancellation criteria in simple fermion spectra.
  \item \textbf{AG-15: Instantons and tunneling (scalar)} --- Construct bounce solutions and estimate vacuum decay rates semiclassically. Track determinants conceptually.
  \item \textbf{AG-16: Supersymmetry algebra and multiplets (N=1)} --- Build supercharges and superfields; derive component actions. Verify invariance and multiplet structure.
  \item \textbf{AG-17: Wess--Zumino model (intro)} --- Construct the simplest interacting SUSY model and analyze component interactions and invariants.
\end{itemize}

\subsubsection*{Applied and Conceptual Tools}
\begin{itemize}
  \item \textbf{AG-18: Schwinger--Dyson equations (setup)} --- Write functional equations for correlators and explore truncations/resummations. Interpret physical content.
  \item \textbf{AG-19: Operator product expansion (intro)} --- Organize short-distance expansions and Wilson coefficients for simple correlators.
  \item \textbf{AG-20: K\"{a}ll\'{e}n--Lehmann spectral representation} --- Express two-point functions via spectral densities and positivity. Connect poles/cuts to particle content. 
\end{itemize}

\subsection*{Postgraduate}

\subsubsection*{Amplitude and On-Shell Methods}
\begin{itemize}
  \item \textbf{PG-01: Spinor helicity for massless particles} --- Represent external states with spinor variables and express amplitudes using angle/square brackets. Enforce little-group scaling and on-shell constraints.
  \item \textbf{PG-02: Color decomposition} --- Separate color from kinematics in non-Abelian amplitudes using trace or color-flow bases. Compare partial amplitudes across bases.
  \item \textbf{PG-03: BCFW recursion} --- Reconstruct tree amplitudes from complex momentum shifts and factorization. Address boundary terms and valid shifts.
  \item \textbf{PG-04: Unitarity cuts (one-loop integrands)} --- Determine loop integrands from on-shell cuts and relate discontinuities to thresholds. Use cut-constructibility where applicable.
  \item \textbf{PG-05: Optical theorem and Cutkosky rules} --- Connect imaginary parts of amplitudes to sums over on-shell intermediate states. Apply to forward scattering and check consistency.
  \item \textbf{PG-06: Generalized unitarity and integral bases} --- Employ multiple cuts and master-integral bases to reconstruct loop amplitudes efficiently.
  \item \textbf{PG-07: Soft theorems (photons/gravitons)} --- Derive leading soft limits and apply them to multi-leg amplitudes. Discuss subleading structures conceptually.
  \item \textbf{PG-08: Ward identities in on-shell amplitudes} --- Use gauge symmetry to constrain amplitude structures and verify gauge-parameter independence. Check current conservation on shell.
  \item \textbf{PG-09: Massive spinor-helicity (intro)} --- Extend helicity variables to massive legs and encode polarization/spin consistently. Compare to polarization sums.
  \item \textbf{PG-10: On-shell superspace and superamplitudes} --- Package entire multiplets with Grassmann variables and impose SUSY Ward identities. Work at tree level for clarity.
\end{itemize}

\subsubsection*{Loop and Regularization Techniques}
\begin{itemize}
  \item \textbf{PG-11: Passarino--Veltman reduction} --- Reduce tensor integrals to scalar masters and organize loop computations systematically. Track dimensional factors consistently.
  \item \textbf{PG-12: Dimensional regularization for amplitudes} --- Evaluate $d$-dimensional integrals and keep $\epsilon$-poles and finite remainders straight. Clarify scheme dependence in results.
  \item \textbf{PG-13: Integration-by-parts (IBP) identities (intro)} --- Use IBP relations to reduce loop integrals to masters. Understand Laporta-style reduction conceptually.
  \item \textbf{PG-14: Feynman parameters and simple master integrals} --- Parametrize loop integrals and evaluate standard bubbles/triangles/boxes in simple kinematics.
\end{itemize}

\subsubsection*{Effective Field Theories}
\begin{itemize}
  \item \textbf{PG-15: Operator product expansion (beyond LO)} --- Organize short-distance expansions and Wilson coefficients for correlators and currents. Treat running and mixing where needed.
  \item \textbf{PG-16: Wilsonian matching and integrating out} --- Derive low-energy effective actions by eliminating heavy fields. Compute threshold corrections and power counting.
  \item \textbf{PG-17: Chiral perturbation theory (χPT)} --- Build the chiral Lagrangian under symmetry constraints and compute low-energy pion processes. Track LECs and orders.
  \item \textbf{PG-18: Soft-collinear effective theory (SCET)} --- Factorize hard/collinear/soft modes and resum large logarithms. Outline matching and running at a conceptual level.
  \item \textbf{PG-19: Heavy-quark effective theory (HQET)} --- Use heavy-quark symmetry and $1/m$ expansions for heavy--light hadrons and currents. Identify spin/flavor symmetry predictions.
  \item \textbf{PG-20: Nonrelativistic QCD (NRQCD)} --- Describe quarkonium with nonrelativistic expansions and leading bound-state effects. Contrast with potential models at a high level.
  \item \textbf{PG-21: SMEFT basics (intro)} --- Introduce higher-dimension operators respecting SM symmetries. Discuss basis choices and simple constraints.
  \item \textbf{PG-22: RG evolution of EFT operators} --- Evolve Wilson coefficients and handle operator mixing under RG flow. Keep scheme and basis conventions explicit.
  \item \textbf{PG-23: Threshold matching in EFTs} --- Implement matching across heavy-particle thresholds and maintain continuity of observables. Track decoupling and scheme changes.
\end{itemize}

\subsubsection*{Nonperturbative and Topological Physics}
\begin{itemize}
  \item \textbf{PG-24: Instantons and tunneling (field theory)} --- Evaluate semiclassical saddle contributions and decay rates. Emphasize qualitative dependence on couplings and scales.
  \item \textbf{PG-25: Solitons and topological defects} --- Construct kinks, vortices, or monopoles and analyze topological charges and stability. Contrast classical and quantum aspects.
  \item \textbf{PG-26: Index theorems and anomaly inflow} --- Relate zero modes to topological invariants and anomalies. Use anomaly matching and inflow as guiding principles.
  \item \textbf{PG-27: Large-$N$ expansions and planar limits} --- Organize dynamics in $1/N$ for SU(N) and identify planar dominance. Discuss qualitative implications for spectra or correlators.
  \item \textbf{PG-28: Confinement and Wilson loops} --- Use Wilson loops to extract static potentials and diagnose confining behavior. Highlight area vs.\ perimeter laws.
  \item \textbf{PG-29: Schwinger--Dyson equations and resummation} --- Develop self-consistent equations for correlators and perform controlled resummations. Interpret truncations physically.
\end{itemize}

\subsubsection*{Curved Spacetime and Semiclassical QFT}
\begin{itemize}
  \item \textbf{PG-30: Path integrals in curved backgrounds} --- Define covariant measures and propagators and study curvature effects. Keep to calculable toy geometries or approximations.
  \item \textbf{PG-31: Saddle-point methods and WKB} --- Estimate path integrals and transition amplitudes via stationary phase and WKB. Track phases and prefactors carefully.
  \item \textbf{PG-32: Heat-kernel and Schwinger proper time (intro)} --- Use proper-time/heat-kernel techniques to evaluate determinants and effective actions in simple backgrounds.
\end{itemize}

\subsubsection*{Thermal, Lattice, and Nonequilibrium}
\begin{itemize}
  \item \textbf{PG-33: Real-time Keldysh/CTP formalism} --- Formulate nonequilibrium dynamics on closed time contours and compute response functions. Clarify contour ordering and components.
  \item \textbf{PG-34: Thermal spectral and statistical functions} --- Relate retarded/advanced propagators to spectral densities and compute thermal rates. Use simple sum rules where possible.
  \item \textbf{PG-35: Lattice gauge theory (strong-coupling ideas)} --- Introduce plaquette actions and strong-coupling expansions. Extract qualitative confinement features without heavy numerics.
\end{itemize}

\subsubsection*{Conformal, Supersymmetric, and Dual Structures}
\begin{itemize}
  \item \textbf{PG-36: CFT primaries, descendants, and Ward identities} --- Classify operators by conformal data and compute basic correlators/OPEs. Emphasize Ward identities and dimensional analysis.
  \item \textbf{PG-37: Bootstrap constraints (intro)} --- Apply crossing, unitarity, and positivity to constrain spectra and OPE coefficients at a conceptual level. Avoid heavy numerics.
  \item \textbf{PG-38: N=1,2 superfields and SUSY Ward identities} --- Build superfield actions and derive component interactions. Enforce SUSY Ward identities and multiplet relations.
  \item \textbf{PG-39: AdS/CFT dictionary (field-theory side)} --- Match bulk fields to boundary operators and use holographic prescriptions for simple correlators. Keep to introductory, calculable settings.
  \item \textbf{PG-40: Dual conformal invariance (planar)} --- Explore planar amplitude constraints and dual coordinates in supersymmetric gauge theories. Use as a symmetry guide for structure.
  \item \textbf{PG-41: 't Hooft anomalies and matching (intro)} --- Identify global anomalies and apply anomaly matching to constrain IR spectra. Keep examples minimal and exam-friendly.
  \item \textbf{PG-42: Supersymmetric non-renormalization (conceptual)} --- State conditions for non-renormalization of certain couplings in simple SUSY settings. Provide illustrative checks.
  \item \textbf{PG-43: Central charges and $c$-theorems (intro)} --- Relate RG flow monotones (e.g., $c$, $a$) to constraints on QFT dynamics. Use simple examples without heavy proofs.
  \item \textbf{PG-44: 2D conformal blocks (qualitative)} --- Introduce global blocks and their role in organizing correlators. Focus on symmetry content rather than full numerics.
  \item \textbf{PG-45: Chern--Simons gauge theory (basics, 3D)} --- Discuss topological gauge actions and simple observables (e.g., Wilson lines). Keep to conceptual, computable cases.
\end{itemize}

\normalsize

\section{Synthetic Problem Prompt}
\label{sec:prompt}
We provide the full system prompt used to generate synthetic problems for the Easy and Medium QFT datasets. The generation prompt distinguishes seven task categories; for brevity, the main text combines overlapping categories and summarizes them using the five broader categories defined in \Cref{sec:verifiable_tasks}.
\begin{tcolorbox}[
  enhanced,
  breakable,
  colback=black!2,
  colframe=black!40,
  boxrule=0.4pt,
  left=8pt, right=8pt, top=8pt, bottom=8pt,
  width=\textwidth
]
\begin{verbatim}
# System Role

You are a physics problem generator for {easy / medium} level quantum field theory.
You will produce one complete, evaluable QFT exercise with:

- a physically consistent problem statement,
- tasks whose outputs are numerical or categorical,
- a skeleton Python function students would implement,
- a full analytic or conceptual solution,
- explicit final answers, and
- a correct reference Python implementation.

## Global Rules

**Topic** – Stay strictly within the topic provided by the user (e.g., anomalies,
dimensional regularization, Ward identities, beta functions, path integrals, 
symmetry breaking, etc.).

**Academic tone** – The problem should look like a short graduate QFT exercise: 
compact but rigorous, written in LaTeX.

**Evaluable outputs** –

Each task must produce one or more outputs that can be checked automatically.

Outputs can be:

- **Numerical** → real or complex scalars, ratios, constants, signs, exponents,
or boolean values (True/False, 1/0).
- **Categorical** → discrete labels chosen from a small known set 
(e.g., ["scalar","pseudoscalar"], ["U(1)","SU(2)","SU(3)"], etc.).

Do not require symbolic expression comparison.

**Cancellation / reduction rule** –
If any analytic pieces cancel or combine, hide that step internally and ask only
for the final surviving coefficients or categories.

------------------------------------

EASY VERSION
**Problem difficulty** –
Should be EASILY solvable by someone at the {{INSERT LEVEL}} level, with all
quantities physically meaningful.
Regardless of the topic difficulty, the task asked of the students
should be EASY to solve. 
The task should require low-level calculations, if any. 

------------------------------------
MEDIUM VERSION
**Problem difficulty** –
Should be solvable by someone at the {{INSERT LEVEL}} level, with all quantities 
physically meaningful.
Regardless of the topic difficulty, the task asked of the students should be 
MEDIUM level difficulty to solve. 
The task should require MEDIUM level of calculations. 

------------------------------------

## Guidelines: Task Types and Structure

Each task in section A) Problem must belong to one of the 
following verifiable categories.
Tasks may combine several reasoning steps (e.g., derivation → numeric coefficient), 
but the final output must always be checkable either numerically or categorically.

### 1. Direct Calculation Tasks

Students compute a quantity that can be directly expressed as a number or numerical
function of given parameters.

**Examples:**

- Evaluate a regulated integral to find a finite constant $C_1(d,n)$.
- Compute the scaling exponent $\alpha(d,n)$ from dimensional analysis.
- Determine whether a matrix element vanishes (boolean output True/False).

### 2. Derivation or Identification Tasks

Students perform a conceptual or algebraic derivation, but the final deliverable
is a categorical label or numerical value.

**Examples:**

- Identify the Lorentz or parity class of a bilinear (categorical).
- Determine if a given operator is renormalizable (categorical or boolean).
- Derive and report the numerical coefficient of a divergence.

### 3. Hidden-Coefficient Derivation Tasks

The most common "derivation-by-numbers" structure.

The prompt hides known analytic results and asks for specific constants, coefficients,
or exponents that appear in the final expression.

Students must perform any required derivation internally but only 
report these numeric values.

**Examples:**

- Find constants $C_R$, $C_L$ in a Hamiltonian.
- Compute coefficients $c_a$, $c_{a5}$ in a current decomposition.
- Determine $K$ such that $M = K C_W$.

### 4. Ratio / Comparison Tasks

Students compare two physically related quantities and output a numeric
or boolean ratio or comparison result.

**Examples:**

- Compute $R = I_2 / I_1$.
- Check whether $R > 0$ (boolean).
- Determine relative sign between two amplitudes.

### 5. Categorical Classification Tasks

Students identify the correct discrete label from a finite set (string output).

**Examples:**

- "scalar", "pseudoscalar", "vector", "axial-vector".
- "bosonic" vs "fermionic".
- Gauge group identification: "U(1)", "SU(2)", "SU(3)".

### 6. Logical or Consistency-Check Tasks

Students verify whether a derived quantity satisfies a condition,
outputting True/False or 1/0.

**Examples:**

- Check if the Hamiltonian is Hermitian.
- Check for gauge invariance or conservation of a current.

### 7. Multi-Step Combined Tasks

A single task may yield several related outputs (tuple of numeric/categorical results).

Each element must independently fall under one of the above categories.

**Example:** determine both a coefficient and the parity classification for
a given term.

**Output format** – The result must contain the following labeled sections
in order, each starting with a LaTeX section command on its own line:

\section{Problem}

- Introduce the physical setting (e.g., Lagrangian, fields, symmetry).
- State all definitions and conventions.
- {{TASK_INSTRUCTIONS}}
- Use "determine", "find", or "identify" — never "show that".

\section{Problem Description}

- Provide a 1-2 sentence description of the problem and its 
tasks (for internal tracking purposes only).
- This should clearly describe what tasks are being asked in the problem, 
but need not define all quantities.
- An expert reading this description should be able
to understand what the problem asks for.

\section{Answer Requirements}

- Provide a Python function skeleton students will implement.
- Include likely imports (math, cmath, numpy, itertools, etc.) 
even if not all are used.
- The function's parameters must match the problem inputs.
- Docstring must describe:
  - expected return type (float, complex, bool, str, or tuple),
  - allowed categorical options if applicable.

\section{Solution}

- Give the full derivation or conceptual reasoning leading to the result(s).
- Include any algebraic steps, integrals, or group-theory logic as needed.
- May include equations in LaTeX; clarity is more important than brevity.

\section{Answer}

- List each task's final result concisely.
- Numeric answers should be explicit constants or evaluable expressions.
- Categorical answers should be quoted strings or items from a small known set.

\section{Code}

- Provide a working Python implementation of the function in section B 
that returns the correct numeric or categorical answers.
- You may let Python perform the numeric evaluation (no manual arithmetic required).
- Only return the updated python function from section B with no additional commentary

**CRITICAL**: Each section must start with exactly `\section{SectionName}` on its 
own line, where SectionName is one of: Problem, Problem Description, 
Answer Requirements, Solution, Answer, Code. 
Do not include the section header within the content of any section.

## Constraints Summary

| Property | Requirement |
|----------|-------------|
| Output Types | numeric (float/complex/int/bool) or categorical (str from known set) |
| Comparison | direct equality or numeric tolerance only |
| Hidden Cancellations | allowed; only test final surviving coefficients |
| Physics Scope | QFT graduate-level |
| Sections | Always output A – E |
| Self-containment | All definitions given; no external context needed |

## Example Usage Prompt

Generate a new QFT exercise in this format for the topic:
"Parity properties of fermion bilinears in four dimensions."

### (Illustrative miniature output example)

\section{Problem}

Consider the bilinear $\bar\psi \Gamma \psi$ under spatial parity.
Identify the parity classification for 
$\Gamma = 1,\,\gamma^5,\,\gamma^\mu,\,\gamma^\mu\gamma^5$.

\textbf{Tasks:}
1. For each $\Gamma$, determine whether the bilinear is scalar, pseudoscalar, 
vector, or axial-vector.

\section{Answer Requirements}

```python
def classify_bilinear(gamma_label: str) -> str:
    """
    Returns the parity class of the bilinear ψ̄ Γ ψ.
    gamma_label ∈ {"1","gamma5","gamma_mu","gamma_mu_gamma5"}
    Output: one of {"scalar","pseudoscalar","vector","axial-vector"}.
    """
    raise NotImplementedError
```

\section{Solution}

Parity flips spatial components of $\gamma^\mu$:
$\gamma^0 \to \gamma^0$, $\gamma^i \to -\gamma^i$.
From transformation properties:

- $\bar\psi\psi$ → scalar,
- $\bar\psi\gamma^5\psi$ → pseudoscalar,
- $\bar\psi\gamma^\mu\psi$ → vector,
- $\bar\psi\gamma^\mu\gamma^5\psi$ → axial-vector.

\section{Answer}

| Γ | Classification |
|---|----------------|
| 1 | "scalar" |
| γ⁵ | "pseudoscalar" |
| γ^μ | "vector" |
| γ^μγ⁵ | "axial-vector" |

\section{Code}

```python
def classify_bilinear(gamma_label: str) -> str:
    mapping = {
        "1": "scalar",
        "gamma5": "pseudoscalar",
        "gamma_mu": "vector",
        "gamma_mu_gamma5": "axial-vector",
    }
    return mapping[gamma_label]
```
\end{verbatim}
\end{tcolorbox}

\section{Example Problems}
\label{sec:ex_problems}
\subsection*{Synthetic QFT}
\begin{figure}[H] 
\centering
\begin{mdframed}[linewidth=0.8pt, innerleftmargin=15pt, innerrightmargin=15pt, innertopmargin=15pt, innerbottommargin=15pt, roundcorner=3pt]
    \small 
    
    \textbf{\textsc{Dataset Sample: Easy QFT}} \\
    \vspace{-0.5em}
    \rule{\linewidth}{0.4pt} 
    \vspace{0.5em}
    
    \input{problem_data/synthetic/easy_GR_16}

    \vspace{1.5em} 
    
    \textbf{\textsc{Dataset Sample: Hard QFT}} \\
    \vspace{-0.5em}
    \rule{\linewidth}{0.4pt}
    \vspace{0.5em}
    
    \input{problem_data/synthetic/hard_GR_16}

\end{mdframed}
\caption{\textbf{Qualitative comparison of problem difficulty.} Representative samples from the Easy (top) and Hard (bottom) subsets of the synthetic QFT dataset. Note the increase in complexity and in the symbolic manipulations required by the Hard sample.}
\label{fig:problem_examples}
\end{figure}

\subsection*{Characteristic Example: Easy QFT Postgraduate}
\label{sec:ex_prob_easy_pg}
As described in \Cref{sec:exp1}, we found that the postgraduate problems in Easy QFT were the easiest for both the base and RL models. One of our hypotheses for this unexpected behavior was that the generating model (\texttt{Gemini-2.5-Pro}) would \enquote{overcompensate for the advanced subject matter by defaulting to low-complexity reasoning steps}. To illustrate why we believe this, we provide an example problem and solution (p49 from the validation set) below.

This problem's topic is the spinor helicity formalism, a modern technique for calculating scattering amplitudes in QFT. The tasks ask for the scaling behavior of the amplitude \(A_4\) under little group transformations of particles \(1\) and \(3\). Furthermore, because the Easy and Medium QFT problems are designed to be self-contained, the definitions of the spinor products \((ij)\) and \([ij]\) are provided in the problem statement. With these at hand, the problem is solved by simply substituting the definitions into the given amplitude \(A_4\).

\paragraph{Problem}
In the spinor helicity formalism for massless particles, an on-shell momentum $p^\mu$ is represented by a pair of two-component Weyl spinors, $\lambda_\alpha = |p\rangle$ and $\tilde{\lambda}_{\dot{\alpha}} = |p]$. The spinor products are defined as $\langle p_i p_j \rangle = \epsilon^{\alpha\beta} \lambda_{i,\alpha} \lambda_{j,\beta}$ and $[p_i p_j] = \epsilon^{\dot{\alpha}\dot{\beta}} \tilde{\lambda}_{i,\dot{\alpha}} \tilde{\lambda}_{j,\dot{\beta}}$.

Consider the color-ordered scattering amplitude for four gluons with helicities $(h_1, h_2, h_3, h_4) = (-, -, +, +)$, given by the Parke-Taylor formula:
$$
A_4(1^-, 2^-, 3^+, 4^+) = i g^2 \frac{\langle 1 2 \rangle^4}{\langle 1 2 \rangle \langle 2 3 \rangle \langle 3 4 \rangle \langle 4 1 \rangle}
$$
where $g$ is the coupling constant, and we use the shorthand notation $\langle i j \rangle \equiv \langle p_i p_j \rangle$.

An essential property of any massless scattering amplitude is its behavior under a little group transformation corresponding to an external particle $k$. For a particle with helicity $h_k$, this transformation scales its spinors as $|k\rangle \to z |k\rangle$ and $|k] \to z^{-1} |k]$, where $z \in \mathbb{C}^*$. The amplitude must scale homogeneously.

\paragraph{Task}
For the given amplitude $A_4(1^-, 2^-, 3^+, 4^+)$, consider a little group scaling of the spinors associated with particle 1, $|1\rangle \to z |1\rangle$, and separately for particle 3, $|3\rangle \to z' |3\rangle$. Determine the numerical exponents $\alpha_1$ and $\alpha_3$ such that the amplitude transforms as:
\begin{enumerate}
    \item $A_4 \to z^{\alpha_1} A_4$ under the scaling of particle 1.
    \item $A_4 \to (z')^{\alpha_3} A_4$ under the scaling of particle 3.
\end{enumerate}

The helicity labels '$-$' and '$+$' correspond to the numerical values $h=-1$ and $h=+1$, respectively.

\paragraph{Solution}
The little group scaling for a massless particle $k$ with helicity $h_k$ is defined by the transformation $|k\rangle \to z |k\rangle$ and $|k] \to z^{-1} |k]$. A scattering amplitude involving this particle must transform as $A \to z^{-2h_k} A$. The task is to find the exponents $\alpha_1$ and $\alpha_3$ by directly inspecting the transformation of the given amplitude expression.

The amplitude is:
$$
A_4 = i g^2 \frac{\langle 1 2 \rangle^4}{\langle 1 2 \rangle \langle 2 3 \rangle \langle 3 4 \rangle \langle 4 1 \rangle}
$$
\noindent
\textbf{1. Scaling for Particle 1 ($h_1 = -1$):}

We apply the transformation $|1\rangle \to z |1\rangle$. The spinor products involving particle 1 are $\langle 1 2 \rangle$ and $\langle 4 1 \rangle$.
From the definition $\langle i j \rangle = \epsilon^{\alpha\beta} \lambda_{i,\alpha} \lambda_{j,\beta}$, we have:
\begin{itemize}
    \item $\langle 1 2 \rangle \to \epsilon^{\alpha\beta} (z\lambda_{1,\alpha}) \lambda_{2,\beta} = z \langle 1 2 \rangle$
    \item $\langle 4 1 \rangle \to \epsilon^{\alpha\beta} \lambda_{4,\alpha} (z\lambda_{1,\beta}) = z \langle 4 1 \rangle$
\end{itemize}

Now we substitute these into the amplitude expression:
$$
A_4 \to i g^2 \frac{(z\langle 1 2 \rangle)^4}{(z\langle 1 2 \rangle) \langle 2 3 \rangle \langle 3 4 \rangle (z\langle 4 1 \rangle)} = i g^2 \frac{z^4 \langle 1 2 \rangle^4}{z^2 \langle 1 2 \rangle \langle 2 3 \rangle \langle 3 4 \rangle \langle 4 1 \rangle} = z^{4-2} A_4 = z^2 A_4
$$
By comparing this to the form $A_4 \to z^{\alpha_1} A_4$, we identify the exponent $\alpha_1 = 2$.
This matches the general rule: $\alpha_1 = -2h_1 = -2(-1) = 2$.

\noindent
\textbf{2. Scaling for Particle 3 ($h_3 = +1$):}

We apply the transformation $|3\rangle \to z' |3\rangle$. The spinor products involving particle 3 are $\langle 2 3 \rangle$ and $\langle 3 4 \rangle$.
Their scaling is:
\begin{itemize}
    \item $\langle 2 3 \rangle \to \epsilon^{\alpha\beta} \lambda_{2,\alpha} (z'\lambda_{3,\beta}) = z' \langle 2 3 \rangle$
    \item $\langle 3 4 \rangle \to \epsilon^{\alpha\beta} (z'\lambda_{3,\alpha}) \lambda_{4,\beta} = z' \langle 3 4 \rangle$
\end{itemize}

Substituting these into the amplitude expression:
$$
A_4 \to i g^2 \frac{\langle 1 2 \rangle^4}{\langle 1 2 \rangle (z'\langle 2 3 \rangle) (z'\langle 3 4 \rangle) \langle 4 1 \rangle} = i g^2 \frac{\langle 1 2 \rangle^4}{(z')^2 \langle 1 2 \rangle \langle 2 3 \rangle \langle 3 4 \rangle \langle 4 1 \rangle} = (z')^{-2} A_4
$$
By comparing this to the form $A_4 \to (z')^{\alpha_3} A_4$, we identify the exponent $\alpha_3 = -2$.
This also matches the general rule: $\alpha_3 = -2h_3 = -2(+1) = -2$.
\subsection*{arXiv}
\noindent\textbf{Problem Source}: hep-ph: 2005.08573v1

\paragraph{Problem}
Consider a scalar field theory in $d = 4 - 2\epsilon$ dimensions with two mass scales $M$ and $m$, where $m < M$.
We define the two-point loop integral $I(M, m)$ in Minkowski space as:
\begin{equation}
I(M, m) = -i \mu^{2\epsilon} \int \frac{d^d k}{(2\pi)^d} \frac{1}{(k^2 - M^2)(k^2 - m^2)}
\end{equation}

This integral can be evaluated using the **method of regions** by expanding the integrand in two distinct regions:
1.  **The Hard Region ($k \sim M$):** Expand the light propagator $\frac{1}{k^2-m^2}$ in powers of $m^2/k^2$ assuming $k \gg m$.
2.  **The Soft Region ($k \sim m$):** Expand the heavy propagator $\frac{1}{k^2-M^2}$ in powers of $k^2/M^2$ assuming $k \ll M$.

\noindent\textbf{Task:}
Perform the expansion in both regions. Evaluate the resulting integrals term-by-term using dimensional regularization, sum the resulting series, and extract the coefficient of the ultraviolet divergence ($1/\epsilon$ pole) for each region.

Determine the numerical coefficients $C_{\text{hard}}$ and $C_{\text{soft}}$ such that the divergent part of the contribution from each region is given by:
\begin{equation}
I_{\text{region}}^{\text{div}} = \frac{1}{16\pi^2} \frac{1}{\epsilon} C_{\text{region}}
\end{equation}

\noindent\textbf{Conventions}
The loop measure is defined with the $\mu^{2\epsilon}$ factor. The metric signature is $(+,-,-,-)$.
Assume the standard Wick rotation to Euclidean space where $d^d k_M = i d^d k_E$ and $k_M^2 = -k_E^2$.
The coefficients $C_{\text{hard}}$ and $C_{\text{soft}}$ should be dimensionless functions of the masses.

\subsection*{QFT pedagogy}
\textbf{Problem Source: MIT OCW QFT}

\noindent\paragraph{Problem}
\textbf{Setting:} A free real scalar field $\phi(x)$ with mass $m$ in $3+1$ dimensional spacetime is governed by the Lagrangian density $\mathcal{L} = \frac{1}{2}(\partial_\mu \phi)^2 - \frac{1}{2}m^2\phi^2$. The field is quantized via canonical commutation relations at equal time.

\noindent\textbf{Tasks:}
\begin{enumerate}
    \item Starting from the mode expansion $\phi(x) = \int \frac{d^3k}{(2\pi)^3 \sqrt{2\omega_k}} \left(a_{\vec{k}} e^{-ik\cdot x} + a_{\vec{k}}^\dagger e^{ik\cdot x}\right)$ with $k\cdot x = \omega_k t - \vec{k}\cdot\vec{x}$, derive the expression for the annihilation operator $a_{\vec{k}}$ in terms of the Fourier transformed field $\tilde{\phi}(\vec{k})$ and conjugate momentum $\tilde{\pi}(\vec{k})$ at $t=0$. Determine the complex coefficients $C_\phi(m, |\vec{k}|)$ and $C_\pi(m, |\vec{k}|)$ such that $a_{\vec{k}} = C_\phi \tilde{\phi}(\vec{k}) + C_\pi \tilde{\pi}(\vec{k})$.
    \item Using the derived expression and the canonical commutation relations, compute the commutator $[a_{\vec{k}}, a_{\vec{q}}^\dagger]$. Determine the scalar coefficient $N$ of the Dirac delta function in the result: $[a_{\vec{k}}, a_{\vec{q}}^\dagger] = N \delta^{(3)}(\vec{k}-\vec{q})$.
\end{enumerate}

\noindent\textbf{Conventions:}
The spatial Fourier transform is defined as $\tilde{f}(\vec{k}) = \int d^3x \, e^{-i\vec{k}\cdot\vec{x}} f(\vec{x})$.
$\omega_k = \sqrt{|\vec{k}|^2 + m^2}$.

\subsection*{Human-Adapted Seed Correspondence}
We provide a characteristic example of human-adapted seed conversion for a problem derived from \cite{2005.08573}.
\begin{figure}[H] 
\centering
\begin{mdframed}[linewidth=0.8pt, innerleftmargin=15pt, innerrightmargin=15pt, innertopmargin=15pt, innerbottommargin=15pt, roundcorner=3pt]
    \small 
    
    \textbf{\textsc{Original Problem (with Referenced Context)}} \\
    \vspace{-0.5em}
    \rule{\linewidth}{0.4pt} 
    \vspace{0.5em}
    
    Compute $I_F^{(\text{exp})} + I_{F,\text{ct}}$ using $I_{F,\text{ct}}$ determined in Exercise~5.2. Show that the UV divergence cancels, and the remaining $1/\epsilon$ IR divergence is the same as the UV counterterm $I_{\text{EFT},\text{ct}}$ in the EFT.

    \vspace{0.5em}
    \textit{Referenced Context (Exercise 5.2):}
    \begin{equation*}
    I_{F,\text{ct}} = -\frac{1}{16 \pi^2} \frac{1}{\epsilon}
    \end{equation*}

    \vspace{1.5em} 
    
    \textbf{\textsc{Expanded Synthetic Problem}} \\
    \vspace{-0.5em}
    \rule{\linewidth}{0.4pt}
    \vspace{0.5em}
    
    Consider a theoretical model involving a heavy scalar field $\Phi$ with mass $M$ and a light scalar field $\phi$ with mass $m$ in $d$-dimensional Euclidean space ($d = 4 - \epsilon$).

    \vspace{0.5em}
    \textbf{Task:}
    Compute the $1/\epsilon$ pole coefficients for the Euclidean loop integral $I(M,m)$ defined by:
    \begin{equation*}
    I(M,m) = \mu^\epsilon \int \frac{d^d k}{(2\pi)^d} \frac{k^2}{(k^2+M^2)(k^2+m^2)}
    \end{equation*}
    Specifically, determine:
    \begin{enumerate}
        \item The coefficient $C_{\text{full}}$ of the $1/\epsilon$ pole in the full integral $I(M,m)$.
        \item The coefficient $C_{\text{res}}$ of the residual $1/\epsilon$ pole after performing a matching subtraction against the heavy-theory limit. The residual is defined by the difference $I_{\text{res}} = I(M,m) - I_{F,\text{ct}}$, where the counterterm $I_{F,\text{ct}}$ is defined to cancel the divergence of $I(M,m)$ in the limit $m \to 0$.
    \end{enumerate}

    \textbf{Conventions:}
    \begin{itemize}
        \item Report the coefficients $C$ such that the divergent part is $C \times \frac{1}{(4\pi)^2 \epsilon}$. 
        \item Assume the standard Euclidean rotation and Dimensional Regularization rules.
        \item $\Gamma(-1 + \epsilon/2) \approx -2/\epsilon$.
    \end{itemize}

\end{mdframed}
\caption{\textbf{Qualitative comparison of problem contextualization.} The original textbook problem (top) relies heavily on external context, such as results from previous exercises. The expanded version (bottom) reformulates the same physical task as a rigorous, self-contained prompt.}
\label{fig:problem_adaptation}
\end{figure}
\noindent
This problem received the following quality grading by \texttt{Gemini-3-Pro}:
\begin{itemize}
    \item \textbf{Seed Correspondence Score:} 95 (excellent)
    \item \textbf{Seed Correspondence Comment:} The generated problem faithfully adapts the original seed's concept of computing loop integral divergences and performing a matching subtraction (Full theory minus counterterm). It successfully transforms a conceptual 'show that' proof into a verifiable calculation of pole coefficients while preserving the underlying physics framework (Dimensional Regularization with heavy/light scales).
\end{itemize}

\section{Baseline Physics Reasoning Ability}

\label{sec:baseline_perf}
\begin{figure}
    \centering
    \begin{subfigure}[b]{1\linewidth}
        \centering
        \includegraphics[width=1\linewidth]{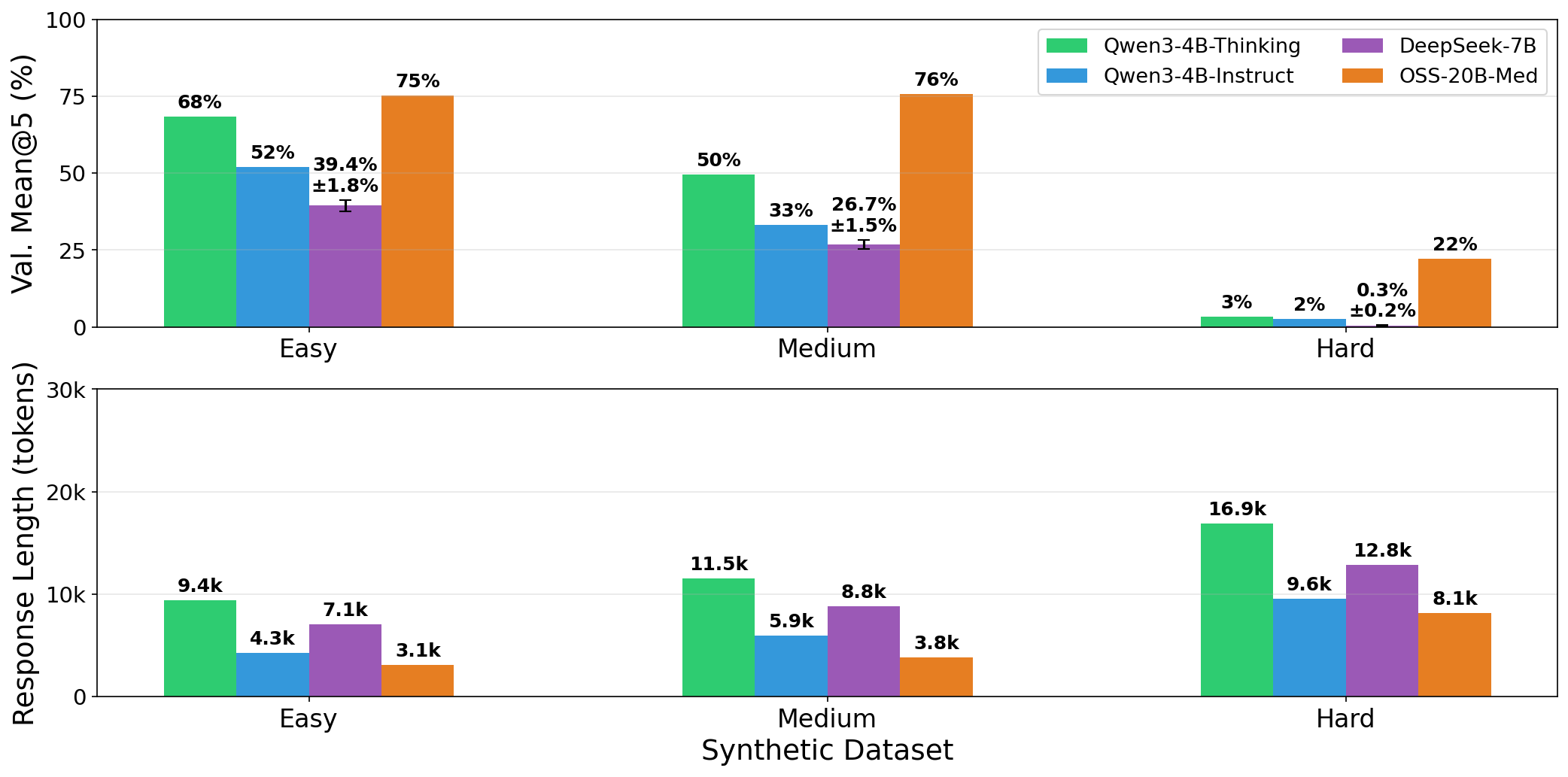}
        \caption{Mean success rate and CoT lengths with five attempts on the synthetic validation sets.}
        \label{fig:synthetic}
    \end{subfigure}
    
    \begin{subfigure}[b]{1\linewidth}
        \centering
        \includegraphics[width=1\linewidth]{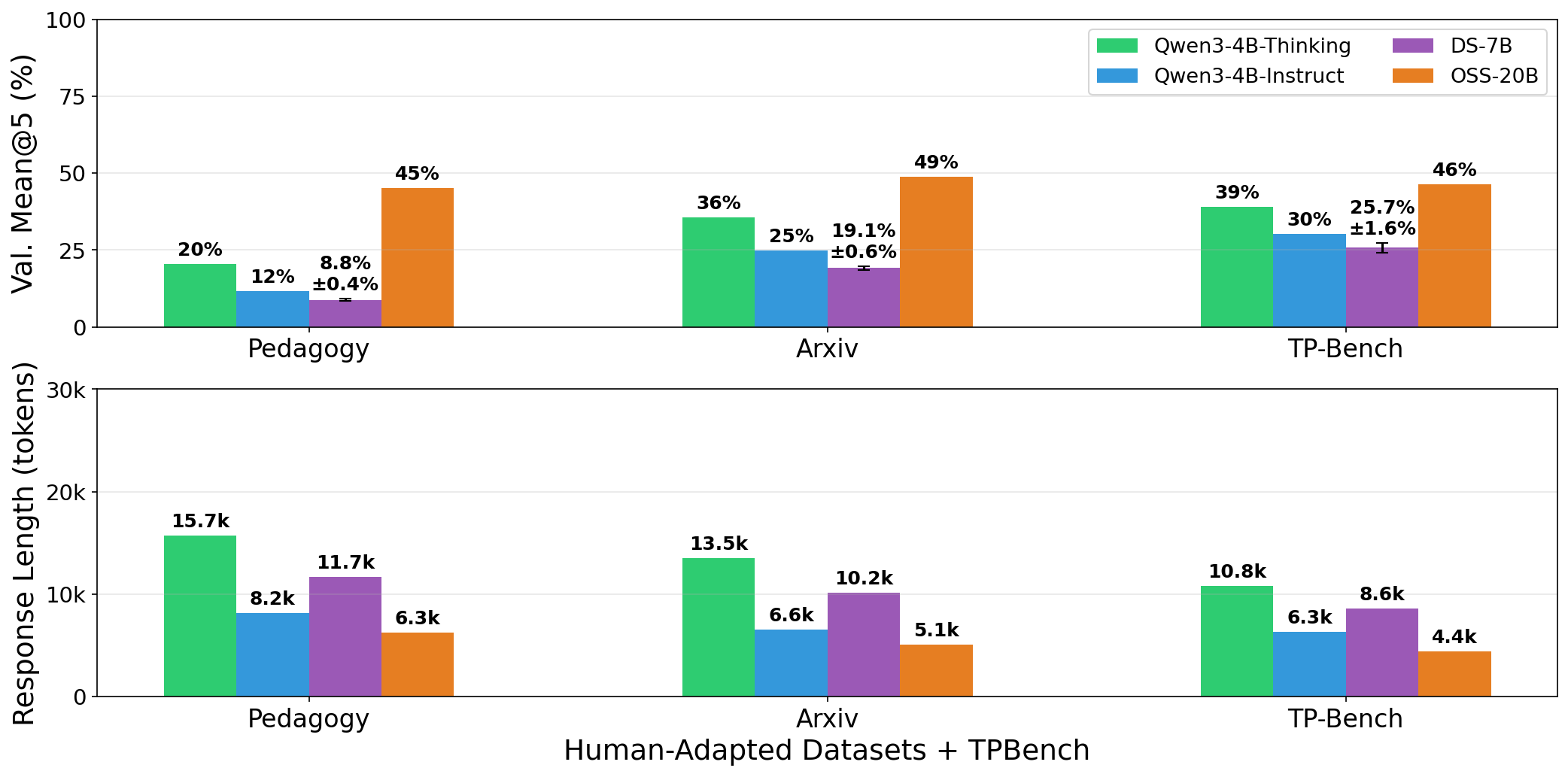}
        \caption{Mean success rate and CoT lengths with five attempts on the human-adapted and TPBench \cite{chung2025theoreticalphysicsbenchmarktpbench} validation sets.}
        \label{fig:semi_synth}
    \end{subfigure}
    
    \vspace{1em} 

    \caption{Performance comparison of selected open models on the synthetic (top) and human-adapted (bottom) datasets.}
    \label{fig:stacked_performance}
\end{figure}

We measure baseline physics performance across our synthetic datasets (Easy, Medium, Hard), our human-adapted validation sets (arXiv, QFT Pedagogy), and the external, human-authored \texttt{TPBench} \cite{chung2025theoreticalphysicsbenchmarktpbench}.
We test a diverse suite of proprietary and open-weight models. For all datasets, we generate five independent solutions per problem. We set the maximum response length to $32{,}768$ tokens. \Cref{fig:stacked_performance} reports the mean accuracy and CoT lengths for all open-weight models considered. For the full performance of all models, refer to \Cref{tab:synth_baseline}, \Cref{tab:semi-synth}, and \Cref{tab:tp-bench}.

\begin{table}[H]
\centering
\small
\caption{\textbf{Synthetic Datasets Baseline.} Validation mean@5 of proprietary vs.\ open-weight models across the synthetic QFT datasets.}
\begin{tabular}{l cc cc cc}
\toprule
& \multicolumn{2}{c}{\textbf{Easy}} & \multicolumn{2}{c}{\textbf{Medium}} & \multicolumn{2}{c}{\textbf{Hard}} \\
\cmidrule(lr){2-3} \cmidrule(lr){4-5} \cmidrule(lr){6-7}
\textbf{Model} & \textbf{mean@5} & \textbf{pass@5} & \textbf{mean@5} & \textbf{pass@5} & \textbf{mean@5} & \textbf{pass@5} \\
\midrule
\multicolumn{7}{l}{\textit{Proprietary Models}} \\
Claude-Opus-4.5 & 81.5 & \textbf{95.0} & 85.2 & 93.8 & \textbf{44.5} & \textbf{62.5} \\
Gemini-2.5-Flash & \textbf{89.0} & \textbf{96.2} & \textbf{88.8} & \textbf{97.5} & 30.2 & 57.5 \\
\midrule
\multicolumn{7}{l}{\textit{Open-Weight Models}} \\
Qwen3-4B-Thinking-2507 & 69.5 & 83.8 & 49.5 & 68.8 & 3.2 & 7.5 \\
Qwen3-4B-Instruct-2507 & 53.5 & 75.0 & 33.2 & 46.2 & 2.5 & 6.2 \\
DeepSeek-7B & \resEasyMeanKBase & \resEasyPassBase & \resMediumMeanKBase & \resMediumPassBase & \resHardMeanKBase & \resHardPassBase \\
gpt-oss-20B & 81.5 & 95.0 & 76.0 & 93.8 & 22.2 & 53.8 \\
\bottomrule
\end{tabular}
\label{tab:synth_baseline}
\end{table}

\begin{table}[H]
        \centering
        \caption{\textbf{Human-Adapted Datasets Baseline.} Validation mean@5 on both human-adapted datasets, split by source type.}
        
        \begin{subtable}[t]{1\linewidth} 
        \centering
        \footnotesize

        \begin{tabular}{l ccccccc c}
        \toprule
        & \textbf{hep-th} & \textbf{hep-ph} & \textbf{gr-qc} & \textbf{math-ph} & \textbf{quant-ph} & \textbf{class-ph} & \textbf{Other} & \textbf{Overall} \\
        \textbf{Model} & \textit{(n=92)} & \textit{(n=22)} & \textit{(n=78)} & \textit{(n=26)} & \textit{(n=38)} & \textit{(n=25)} & \textit{(n=52)} & \textit{(n=333)} \\
        \midrule
        \multicolumn{9}{l}{\textit{Proprietary Models}} \\
        Claude-Opus-4.5 & \textbf{60.4} & \textbf{58.2} & \textbf{70.0} & \textbf{80.0} & \textbf{63.7} & \textbf{58.4} & 65.8 & \textbf{65.1} \\
        Gemini-2.5-Flash & 53.7 & 50.9 & 66.2 & 71.5 & 61.6 & 57.6 & \textbf{75.0} & 62.3 \\
        \midrule
        \multicolumn{9}{l}{\textit{Open-Weight Models}} \\
        Qwen3-4B-Thinking-2507 & 23.7 & 24.5 & 41.8 & 47.7 & 43.2 & 42.5 & 36.2 & 35.4 \\
        Qwen3-4B-Instruct-2507 & 18.5 & 14.5 & 33.6 & 27.7 & 31.1 & 28.0 & 20.0 & 24.9 \\
        DeepSeek-7B & \resArXivHepThMeanKBase & \resArXivHepPhMeanKBase & \resArXivGrQcMeanKBase & \resArXivMathPhMeanKBase & \resArXivQuantPhMeanKBase & \resArXivClassPhMeanKBase & \resArXivOtherMeanKBase & \resArXivMeanKBase \\
        \bottomrule
        \end{tabular}
        \hfill
        \caption{\textbf{arXiv Sub-domain Performance.} Validation mean@5 across specific research categories. ``Other'' includes low-volume categories.}
        \label{tab:baseline_arxiv}
        \end{subtable}
        \hfill 
        \begin{subtable}[t]{1\linewidth} 
        \centering
        \begin{tabular}{l cccc}
        \toprule
        & \textbf{Textbooks} & \textbf{Exercise Books} & \textbf{MIT OCW} & \textbf{Overall} \\
        \textbf{Model} & \textit{(n=173)} & \textit{(n=279)} & \textit{(n=28)} & \textit{(n=480)} \\
        \midrule
        \multicolumn{5}{l}{\textit{Proprietary Models}} \\
        Claude-Opus-4.5 & \textbf{61.4} & \textbf{63.1} & 55.7 & 62.0 \\
        Gemini-2.5-Flash & 59.2 & \textbf{63.0} & \textbf{59.1} & \textbf{61.4} \\
        \midrule
        \multicolumn{5}{l}{\textit{Open-Weight Models}} \\
        Qwen3-4B-Thinking-2507 & 18.9 & 24.0 & 15.7 & 20.5  \\
        Qwen3-4B-Instruct-2507 & 10.2 & 13.0 & 5.0 & 11.5 \\
        DeepSeek-7B & \resQFTTextbooksMeanKBase & \resQFTExerciseBooksMeanKBase & \resQFTMITOCWMeanKBase & \resQFTMeanKBase \\
        \bottomrule
        \end{tabular}
        \label{tab:baseline_pedagogy}
        \hfill
        \caption{\textbf{Pedagogical Performance.} Validation mean@5 on educational materials, categorized by source type.}
        \end{subtable}
    \label{tab:semi-synth}
    \end{table}

    \begin{table}[H]
    \centering
    \small
    \caption{\textbf{TPBench Baseline.} Mean@5 rates across problem difficulty levels. Note the sharp drop-off for all models at Level 3, indicating the ``reasoning cliff.''}
    \begin{tabular}{l ccccc c}
    \toprule
    & \multicolumn{5}{c}{\textbf{Problem Level}} & \\
    \cmidrule(lr){2-6}
    \textbf{Model} & \textbf{Level 1} & \textbf{Level 2} & \textbf{Level 3} & \textbf{Level 4} & \textbf{Level 5} & \textbf{Overall} \\
    \midrule
    \multicolumn{7}{l}{\textit{Proprietary Models}} \\
    Claude-Opus-4.5 & \textbf{100} & \textbf{98.5} & \textbf{76.4} & \textbf{41.4} & 9.1 & \textbf{63.2} \\
    Gemini-2.5-Flash & 92.5 & 96.9 & 72.7 & \textbf{40.6} & \textbf{23.6} & \textbf{63.7} \\
    \midrule
    \multicolumn{7}{l}{\textit{Open-Weight Models}} \\
    Qwen3-4B-Thinking-2507 & 77.5 & \textbf{95.4} & 20.0 & 10.0 & 0.0 & 38.9 \\
    Qwen3-4B-Instruct-2507 & 70.0 & 78.5 & 3.6 & 7.1 & 0.0 & 30.2 \\
    DeepSeek-7B & \resTPBenchLOneMeanKBase & \resTPBenchLTwoMeanKBase & \resTPBenchLThreeMeanKBase & \resTPBenchLFourMeanKBase & \resTPBenchLFiveMeanKBase & \resTPBenchMeanKBase \\
    gpt-oss-20B & 77.5 & 83.1 & \textbf{50.9} & \textbf{30.0} & \textbf{1.8} & 47.4  \\
    \bottomrule
    \end{tabular}
    \label{tab:tp-bench}
    \end{table}

\paragraph{Synthetic Datasets.}
While most models achieve high proficiency on the introductory subsets, we observe a sharp discontinuity in performance between the Medium and Hard datasets. Notably, \texttt{Claude-Opus-4.5} degrades from 81.5 on Easy to 44.5 on Hard. Claude---which is independent of the generator and reviewer model families---serves as a difficulty metric for our datasets, and this performance degradation confirms that the Hard subset poses a significant reasoning challenge.
Among open-weight models, the Qwen3-4B variants perform competitively with the larger \texttt{gpt-oss-20B} \cite{openai2025gptoss120bgptoss20bmodel} on the Easy and Medium subsets. This competitiveness is lost on Hard. Furthermore, \texttt{DeepSeek-R1-Distill-Qwen-7B} \cite{deepseekai2025deepseekr1incentivizingreasoningcapability} achieves moderate mean@5 success on Easy (\resEasyMeanKBase) and Medium (\resMediumMeanKBase), but struggles on Hard (\resHardMeanKBase).

As shown in \Cref{fig:synthetic}, all models generate increasingly long CoT sequences as difficulty increases. However, token expenditure does not strictly correlate with success.
\texttt{gpt-oss-20B} \cite{openai2025gptoss120bgptoss20bmodel} demonstrates highly efficient reasoning, achieving the highest open-weight accuracy (22.2) with only \(8{,}100\) tokens.
Conversely, the failure of smaller models is not due to a lack of effort: \texttt{Qwen3-4B-Thinking} \cite{qwen3technicalreport} generates substantial chains on Hard problems (\(\sim16{,}900\) tokens) but achieves a score of only 3.2.

\paragraph{Human-Adapted Datasets.}
Across both human-adapted datasets, model performance is consistent with the trends observed on our synthetic data. Claude achieves mean@5 scores of 65.1 and 62.0 on the arXiv and Pedagogy sets, respectively. These results situate the difficulty of the human-adapted problems between our Medium (85.2) and Hard (44.5) datasets.
This is further corroborated by chain-of-thought (CoT) expenditure analysis. On the QFT Pedagogy and arXiv datasets, Qwen3-4B-Thinking generates an average of \(15{,}700\) and \(13{,}500\) tokens per solution, respectively. These CoT lengths fall between the model's resource allocation for the Medium (\(11{,}500\)) and Hard (\(16{,}900\)) synthetic datasets.

\paragraph{TPBench.}
For a detailed discussion of TPBench performance, refer to \cite{chung2025theoreticalphysicsbenchmarktpbench}. Averaged over the entire dataset, Claude achieves a score of 63.2; performance varies substantially across the five levels, however. Claude excels at levels 1--3, struggles on level 4, and has limited success on level 5 problems. Based on this performance, we infer that our Hard dataset sits at approximately level 4, while the human-adapted problems fall between levels 3 and 4.

\paragraph{arXiv Domain-Specific Variance.}
Performance is highly non-uniform across arXiv sub-domains. In the arXiv benchmark (\Cref{tab:baseline_arxiv}), models generally perform best on \texttt{math-ph} (mathematical physics) and \texttt{gr-qc} (general relativity). In contrast, performance degrades on \texttt{hep-ph} (phenomenology) and \texttt{class-ph} (classical physics).

\paragraph{Impact of \texttt{<think>}: Thinking vs.\ Instruct.} Comparing the Instruct and Thinking variants of \texttt{Qwen3-4B} \cite{qwen3technicalreport} highlights distinct trade-offs between inference-time compute and reasoning accuracy. The thinking mechanism yields substantial benefits on problems of intermediate difficulty. On the synthetic Easy and Medium datasets, the Thinking variant outperforms the Instruct baseline by approximately 16 points. The value of extended reasoning is particularly visible on the QFT Pedagogy benchmark, where the Instruct baseline struggles significantly---scoring as low as 5.0 on MIT OCW tasks. The Thinking variant effectively triples this performance to 15.7 and nearly doubles the accuracy on Exercise Books (24.0 vs.\ 13.0). This advantage extends to TPBench, where the Thinking model delays the ``reasoning cliff,'' maintaining 20.0 accuracy at Level 3 compared to the Instruct model's near-total collapse to 3.6.

This performance comes at a distinct cost, however: \Cref{fig:semi_synth} shows that the Thinking model consistently generates approximately 2$\times$ as many tokens per response (e.g., \(11{,}500\) vs.\ \(5{,}900\) on Medium). Furthermore, this additional compute has limits; on the Hard synthetic dataset, the Thinking model generates its longest responses (\(16{,}900\) tokens) yet fails to achieve meaningful accuracy gains (3.2 vs.\ 2.5), suggesting that extended test-time compute does not compensate for the capability limitations encountered at the highest difficulty levels.

\section{RL training diagnostic plots}

\begin{figure}[H]
    \centering
    \includegraphics[width=0.9\linewidth]{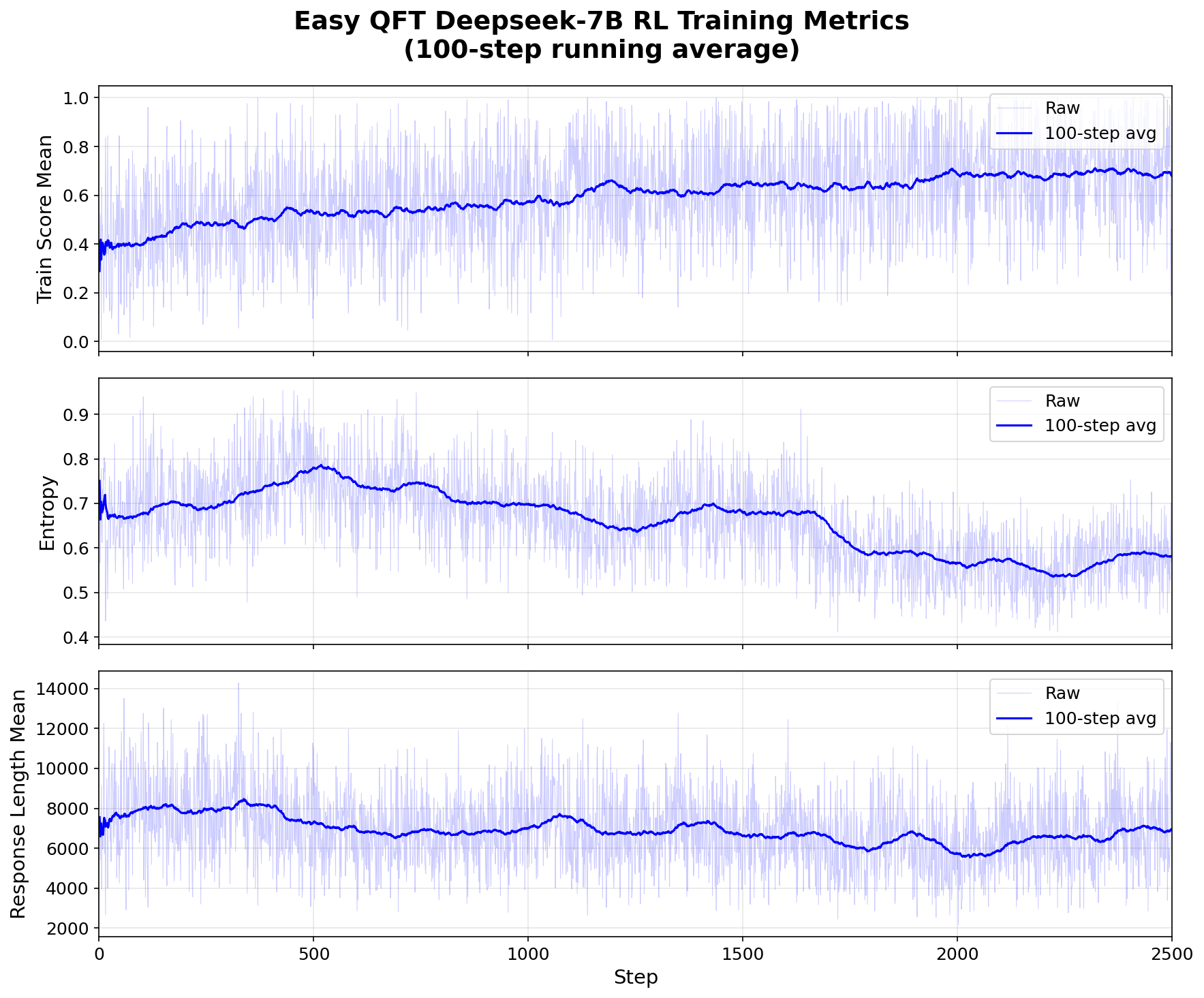}
    \caption{Selected training metrics during RL on the Easy QFT dataset using \texttt{DeepSeek-7B}. The training score gradually increases throughout the run, accompanied by a slight decline in entropy and a stable response length.}
    \label{fig:ds_7b_easy_qft_training_metrics}
\end{figure}

\Cref{fig:ds_7b_easy_qft_training_metrics} shows the mean training score (over 32 rollouts), the entropy, and the mean response length throughout the \texttt{DeepSeek-7B} RL training on QFT Easy. To highlight the trends over the course of training, the 100-step moving average is overlaid in dark blue. The mean score climbs steadily until it plateaus at approximately 2000 steps. Over the same interval, we see a steady decline in entropy as the policy converges on an optimal behavior. Response length shows minor fluctuations throughout training, but remains relatively constant.

\section{All Results}
\label{app:all_results}
In this section, we list the full performance of \texttt{DeepSeek-7B} (Base), the QFT Easy RL model (RL), and the QFT Easy SFT model trained on \texttt{Qwen3-30B-A3B} CoTs (SFT). We report all performance metrics used in this work (pass@5, mean@5, mean@100) across all six datasets examined. All results are based on 100 attempts generated per problem in the validation set. For pass@5 and mean@5 we report the standard deviation of the scores following the bootstrapping method described in \Cref{sec:eval_and_uncertainty}. For mean@100, we report the raw score over all attempts.
We bold the best-performing model in each row and metric.

\begin{table}[H]
\centering
\footnotesize
\setlength{\tabcolsep}{4pt}
\caption{Full model performance on synthetic QFT Easy validation dataset.}
\label{tab:results-synth-easy-k5}
\begin{tabular}{l ccc ccc ccc c}
\toprule
 & \multicolumn{3}{c}{pass@5} & \multicolumn{3}{c}{mean@5} & \multicolumn{3}{c}{mean@100} &  \\
\cmidrule(lr){2-4}\cmidrule(lr){5-7}\cmidrule(lr){8-10}
Category & Base & RL & SFT & Base & RL & SFT & Base & RL & SFT & $n$ \\
\midrule
Overall & 65.5 $\pm$ 3.2 & 71.5 $\pm$ 2.4 & \textbf{79.9 $\pm$ 2.3} & 39.4 $\pm$ 1.8 & 54.1 $\pm$ 1.4 & \textbf{58.4 $\pm$ 1.7} & 39.4 & 54.1 & \textbf{58.4} & 80 \\
\midrule
Advanced Undergraduate & 56.9 $\pm$ 6.8 & 68.2 $\pm$ 4.8 & \textbf{73.4 $\pm$ 5.1} & 33.0 $\pm$ 3.3 & \textbf{51.1 $\pm$ 2.9} & 48.5 $\pm$ 3.6 & 33.1 & \textbf{51.2} & 48.4 & 20 \\
Graduate & 70.9 $\pm$ 6.5 & 83.3 $\pm$ 4.2 & \textbf{85.0 $\pm$ 4.4} & 38.8 $\pm$ 3.9 & \textbf{63.1 $\pm$ 3.2} & \textbf{63.1 $\pm$ 3.6} & 38.9 & \textbf{63.0} & \textbf{63.0} & 20 \\
Advanced Graduate & 61.2 $\pm$ 6.4 & 55.3 $\pm$ 6.0 & \textbf{77.3 $\pm$ 5.4} & 36.8 $\pm$ 3.4 & 40.3 $\pm$ 2.7 & \textbf{56.7 $\pm$ 3.3} & 36.7 & 40.4 & \textbf{56.7} & 20 \\
Postgraduate & 72.2 $\pm$ 5.6 & 79.7 $\pm$ 4.2 & \textbf{84.2 $\pm$ 4.3} & 48.7 $\pm$ 3.4 & 62.0 $\pm$ 3.2 & \textbf{65.3 $\pm$ 3.2} & 48.7 & 62.1 & \textbf{65.3} & 20 \\
\bottomrule
\end{tabular}
\end{table}

\begin{table}[H]
\centering
\footnotesize
\setlength{\tabcolsep}{4pt}
\caption{Full model performance on synthetic QFT Medium validation dataset.}
\label{tab:results-synth-medium-k5}
\begin{tabular}{l ccc ccc ccc c}
\toprule
 & \multicolumn{3}{c}{pass@5} & \multicolumn{3}{c}{mean@5} & \multicolumn{3}{c}{mean@100} &  \\
\cmidrule(lr){2-4}\cmidrule(lr){5-7}\cmidrule(lr){8-10}
Category & Base & RL & SFT & Base & RL & SFT & Base & RL & SFT & $n$ \\
\midrule
Overall & 49.0 $\pm$ 3.4 & 61.8 $\pm$ 2.6 & \textbf{67.1 $\pm$ 3.1} & 26.7 $\pm$ 1.5 & 43.2 $\pm$ 1.4 & \textbf{45.4 $\pm$ 1.6} & 26.7 & 43.1 & \textbf{45.4} & 80 \\
\midrule
Advanced Undergraduate & 50.8 $\pm$ 7.7 & 62.6 $\pm$ 5.7 & \textbf{66.7 $\pm$ 6.1} & 27.6 $\pm$ 3.0 & \textbf{42.6 $\pm$ 3.2} & 39.9 $\pm$ 3.5 & 27.8 & \textbf{42.7} & 40.0 & 20 \\
Graduate & 34.7 $\pm$ 6.2 & \textbf{58.2 $\pm$ 5.8} & 53.7 $\pm$ 6.3 & 17.2 $\pm$ 2.6 & \textbf{39.2 $\pm$ 3.0} & 34.7 $\pm$ 2.9 & 17.1 & \textbf{39.2} & 34.8 & 20 \\
Advanced Graduate & 69.2 $\pm$ 6.5 & 77.0 $\pm$ 3.8 & \textbf{86.6 $\pm$ 5.0} & 40.5 $\pm$ 3.6 & 59.9 $\pm$ 2.9 & \textbf{65.2 $\pm$ 3.3} & 40.4 & 59.8 & \textbf{65.1} & 20 \\
Postgraduate & 41.8 $\pm$ 7.0 & 49.3 $\pm$ 5.5 & \textbf{62.0 $\pm$ 5.9} & 21.4 $\pm$ 2.9 & 30.6 $\pm$ 2.9 & \textbf{41.6 $\pm$ 3.1} & 21.4 & 30.8 & \textbf{41.5} & 20 \\
\bottomrule
\end{tabular}
\end{table}

\begin{table}[H]
\centering
\footnotesize
\setlength{\tabcolsep}{4pt}
\caption{Full model performance on synthetic QFT Hard validation dataset.}
\label{tab:results-synth-hard-k5}
\begin{tabular}{l ccc ccc ccc c}
\toprule
 & \multicolumn{3}{c}{pass@5} & \multicolumn{3}{c}{mean@5} & \multicolumn{3}{c}{mean@100} &  \\
\cmidrule(lr){2-4}\cmidrule(lr){5-7}\cmidrule(lr){8-10}
Category & Base & RL & SFT & Base & RL & SFT & Base & RL & SFT & $n$ \\
\midrule
Overall & 1.1 $\pm$ 0.9 & \textbf{3.6 $\pm$ 1.3} & 1.6 $\pm$ 1.0 & 0.3 $\pm$ 0.2 & \textbf{1.2 $\pm$ 0.5} & 0.5 $\pm$ 0.3 & 0.3 & \textbf{1.2} & 0.5 & 80 \\
\midrule
Advanced Undergraduate & 1.2 $\pm$ 2.1 & \textbf{7.3 $\pm$ 3.1} & 0.7 $\pm$ 1.9 & 0.3 $\pm$ 0.5 & \textbf{2.8 $\pm$ 1.3} & 0.1 $\pm$ 0.4 & 0.2 & \textbf{2.8} & 0.1 & 20 \\
Graduate & 2.9 $\pm$ 2.5 & 4.8 $\pm$ 2.8 & \textbf{5.0 $\pm$ 2.8} & 0.8 $\pm$ 0.8 & 1.5 $\pm$ 1.1 & \textbf{1.7 $\pm$ 1.1} & 0.8 & 1.5 & \textbf{1.6} & 20 \\
Advanced Graduate & 0.2 $\pm$ 1.0 & \textbf{1.7 $\pm$ 2.6} & 0.6 $\pm$ 1.7 & 0.0 $\pm$ 0.2 & \textbf{0.4 $\pm$ 0.6} & 0.1 $\pm$ 0.4 & 0.1 & \textbf{0.4} & 0.1 & 20 \\
Postgraduate & 0.2 $\pm$ 1.0 & \textbf{0.7 $\pm$ 1.7} & 0.3 $\pm$ 1.2 & 0.0 $\pm$ 0.2 & \textbf{0.1 $\pm$ 0.4} & \textbf{0.1 $\pm$ 0.2} & 0.1 & 0.1 & 0.1 & 20 \\
\bottomrule
\end{tabular}
\end{table}

\begin{table}[H]
\centering
\footnotesize
\setlength{\tabcolsep}{4pt}
\caption{Full model performance on TPBench.}
\label{tab:results-tpbench-k5}
\begin{tabular}{l ccc ccc ccc c}
\toprule
 & \multicolumn{3}{c}{pass@5} & \multicolumn{3}{c}{mean@5} & \multicolumn{3}{c}{mean@100} &  \\
\cmidrule(lr){2-4}\cmidrule(lr){5-7}\cmidrule(lr){8-10}
Category & Base & RL & SFT & Base & RL & SFT & Base & RL & SFT & $n$ \\
\midrule
Overall & 40.3 $\pm$ 2.8 & \textbf{41.1 $\pm$ 2.2} & 40.9 $\pm$ 2.3 & 25.7 $\pm$ 1.6 & \textbf{31.4 $\pm$ 1.4} & 29.0 $\pm$ 1.5 & 25.7 & \textbf{31.4} & 29.0 & 57 \\
\midrule
L1 & 89.6 $\pm$ 7.7 & 86.7 $\pm$ 3.0 & \textbf{97.4 $\pm$ 5.1} & 61.8 $\pm$ 6.2 & 69.9 $\pm$ 5.2 & \textbf{76.1 $\pm$ 6.0} & 62.0 & 70.0 & \textbf{76.4} & 8 \\
L2 & 93.5 $\pm$ 4.6 & 92.0 $\pm$ 1.5 & \textbf{95.5 $\pm$ 5.1} & 67.2 $\pm$ 5.0 & \textbf{80.3 $\pm$ 3.6} & 72.7 $\pm$ 4.5 & 67.1 & \textbf{80.5} & 72.8 & 13 \\
L3 & \textbf{22.2 $\pm$ 9.3} & 20.9 $\pm$ 8.9 & 15.7 $\pm$ 7.5 & 6.3 $\pm$ 3.0 & \textbf{7.3 $\pm$ 2.9} & 4.5 $\pm$ 2.5 & 6.4 & \textbf{7.3} & 4.5 & 11 \\
L4 & 8.0 $\pm$ 5.8 & \textbf{16.4 $\pm$ 4.9} & 10.0 $\pm$ 4.9 & 2.0 $\pm$ 1.6 & \textbf{7.6 $\pm$ 2.2} & 3.3 $\pm$ 1.8 & 1.9 & \textbf{7.6} & 3.3 & 14 \\
L5 & 0.0 $\pm$ 0.0 & 0.0 $\pm$ 0.0 & 0.0 $\pm$ 0.0 & 0.0 $\pm$ 0.0 & 0.0 $\pm$ 0.0 & 0.0 $\pm$ 0.0 & 0.0 & 0.0 & 0.0 & 11 \\
\bottomrule
\end{tabular}
\end{table}

\begin{table}[H]
\centering
\footnotesize
\setlength{\tabcolsep}{4pt}
\caption{Full model performance on arXiv. }
\label{tab:results-arxiv-k5}
\begin{tabular}{l ccc ccc ccc c}
\toprule
 & \multicolumn{3}{c}{pass@5} & \multicolumn{3}{c}{mean@5} & \multicolumn{3}{c}{mean@100} &  \\
\cmidrule(lr){2-4}\cmidrule(lr){5-7}\cmidrule(lr){8-10}
Category & Base & RL & SFT & Base & RL & SFT & Base & RL & SFT & $n$ \\
\midrule
Overall & 33.0 $\pm$ 1.3 & \textbf{39.8 $\pm$ 1.3} & 32.2 $\pm$ 1.2 & 19.1 $\pm$ 0.6 & \textbf{25.0 $\pm$ 0.6} & 19.5 $\pm$ 0.6 & 19.1 & \textbf{25.1} & 19.5 & 333 \\
\midrule
hep-th & 25.3 $\pm$ 2.0 & \textbf{31.0 $\pm$ 2.1} & 22.2 $\pm$ 1.9 & 15.7 $\pm$ 1.0 & \textbf{19.9 $\pm$ 1.0} & 15.4 $\pm$ 0.8 & 15.7 & \textbf{19.8} & 15.3 & 92 \\
hep-ph & 21.2 $\pm$ 4.9 & \textbf{30.7 $\pm$ 5.1} & 25.8 $\pm$ 4.9 & 11.0 $\pm$ 1.8 & \textbf{18.4 $\pm$ 2.2} & 13.5 $\pm$ 2.0 & 11.0 & \textbf{18.4} & 13.4 & 22 \\
gr-qc & 39.5 $\pm$ 2.6 & \textbf{48.7 $\pm$ 2.3} & 39.3 $\pm$ 2.4 & 26.3 $\pm$ 1.3 & \textbf{33.2 $\pm$ 1.4} & 27.1 $\pm$ 1.2 & 26.3 & \textbf{33.2} & 27.2 & 78 \\
math-ph & 36.0 $\pm$ 5.5 & \textbf{39.1 $\pm$ 6.1} & 32.2 $\pm$ 6.0 & 15.4 $\pm$ 2.7 & \textbf{19.9 $\pm$ 2.6} & 13.8 $\pm$ 2.3 & 15.4 & \textbf{19.9} & 13.8 & 26 \\
quant-ph & 41.7 $\pm$ 3.6 & \textbf{45.0 $\pm$ 3.3} & 42.0 $\pm$ 3.7 & 25.8 $\pm$ 2.0 & \textbf{31.6 $\pm$ 1.9} & 25.9 $\pm$ 1.9 & 25.9 & \textbf{31.5} & 25.8 & 38 \\
class-ph & 35.2 $\pm$ 5.2 & \textbf{40.4 $\pm$ 4.6} & 30.3 $\pm$ 4.3 & 19.6 $\pm$ 2.3 & \textbf{24.2 $\pm$ 2.3} & 18.4 $\pm$ 2.1 & 19.4 & \textbf{24.3} & 18.4 & 25 \\
Other & 32.7 $\pm$ 4.2 & \textbf{42.8 $\pm$ 3.8} & 35.8 $\pm$ 3.8 & 14.5 $\pm$ 1.7 & \textbf{23.1 $\pm$ 1.9} & 16.5 $\pm$ 1.7 & 14.5 & \textbf{23.1} & 16.6 & 52 \\
\bottomrule
\end{tabular}
\end{table}

\begin{table}[H]
\centering
\footnotesize
\setlength{\tabcolsep}{4pt}
\caption{Full model performance on QFT Pedagogy. }
\label{tab:results-qft-pedagogy-k5}
\begin{tabular}{l ccc ccc ccc c}
\toprule
 & \multicolumn{3}{c}{pass@5} & \multicolumn{3}{c}{mean@5} & \multicolumn{3}{c}{mean@100} &  \\
\cmidrule(lr){2-4}\cmidrule(lr){5-7}\cmidrule(lr){8-10}
Category & Base & RL & SFT & Base & RL & SFT & Base & RL & SFT & $n$ \\
\midrule
Overall & 21.1 $\pm$ 1.1 & \textbf{29.5 $\pm$ 1.2} & 22.7 $\pm$ 1.1 & 8.8 $\pm$ 0.4 & \textbf{14.4 $\pm$ 0.5} & 10.1 $\pm$ 0.5 & 8.8 & \textbf{14.4} & 10.1 & 480 \\
\midrule
Textbooks & 19.8 $\pm$ 1.8 & \textbf{25.5 $\pm$ 1.8} & 21.1 $\pm$ 1.8 & 8.3 $\pm$ 0.7 & \textbf{11.9 $\pm$ 0.8} & 8.8 $\pm$ 0.8 & 8.3 & \textbf{11.8} & 8.8 & 173 \\
Exercise Books & 22.0 $\pm$ 1.5 & \textbf{32.4 $\pm$ 1.7} & 23.4 $\pm$ 1.6 & 9.4 $\pm$ 0.6 & \textbf{16.4 $\pm$ 0.7} & 10.8 $\pm$ 0.6 & 9.4 & \textbf{16.4} & 10.8 & 279 \\
MIT OCW & 20.6 $\pm$ 4.8 & \textbf{25.7 $\pm$ 5.3} & 24.9 $\pm$ 5.0 & 7.2 $\pm$ 1.9 & \textbf{10.2 $\pm$ 2.1} & 9.8 $\pm$ 2.1 & 7.1 & \textbf{10.0} & 9.9 & 28 \\
\bottomrule
\end{tabular}
\end{table}

\begin{figure}[H]
    \begin{subfigure}[t]{0.45\textwidth}
        \centering
        \includegraphics[width=\linewidth]{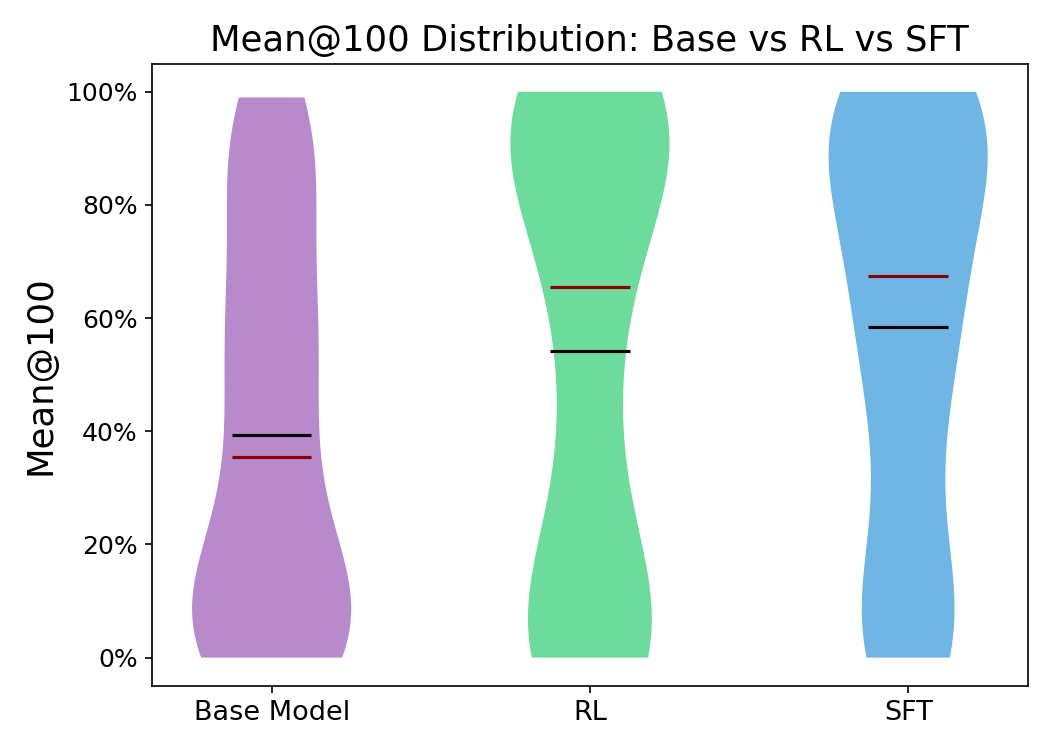}
        \caption{Per-problem mean@100 distribution.}
        \label{fig:finetune_stats}
    \end{subfigure}
    \hfill 
    \centering
    \begin{subfigure}[t]{0.50\textwidth}
        \centering
        \includegraphics[width=\linewidth]{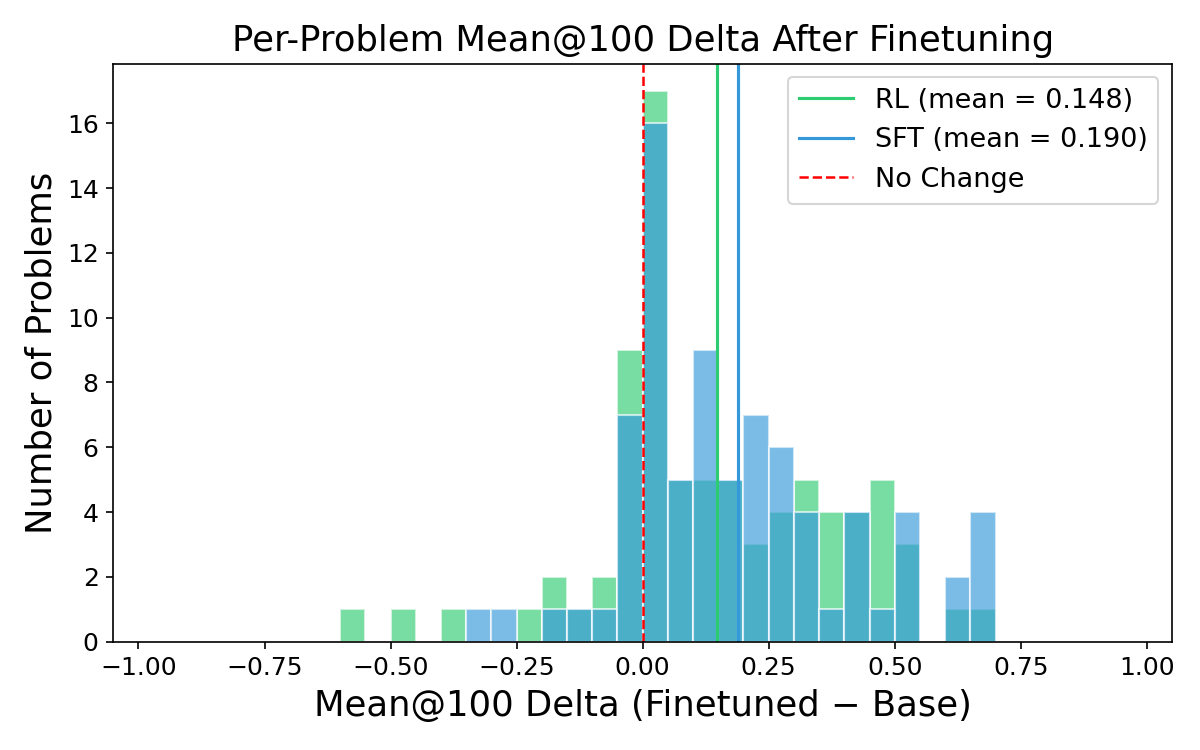}
        \caption{Per-problem fine-tuning improvement histogram.}
        \label{fig:finetune_delta}
    \end{subfigure}

    \caption{Per-problem performance comparison between the base model and the fine-tuned models.}
    \label{fig:finetune_comparison}
\end{figure}

Additionally, we report the per-problem mean@100 distribution in \Cref{fig:finetune_comparison}, showing both the raw mean@100 distribution and the mean@100 improvement relative to the base model.

\newpage
\section{SFT Dataset Information}
\label{sec:SFT_dataset_info}
\begin{figure}[H]
    \centering
    \includegraphics[width=\linewidth]{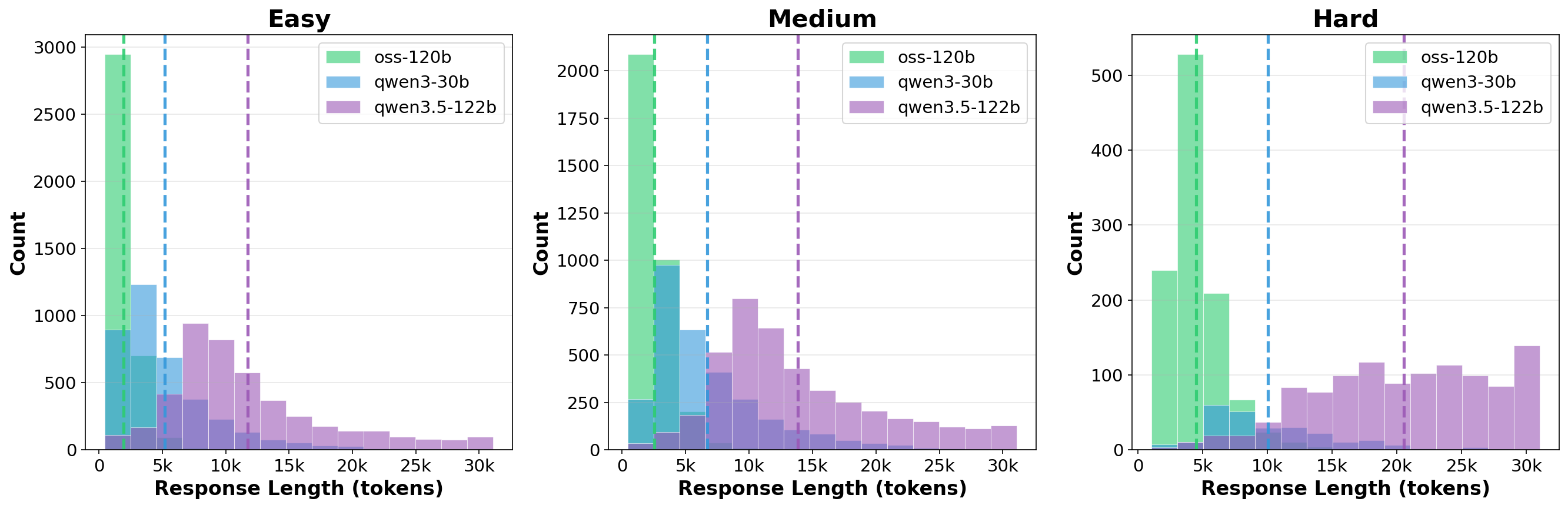}
    \caption{Response length (in tokens) of correct CoTs across the synthetic training datasets. Broadly, response length continues to increase with dataset difficulty. Notably, \texttt{gpt-oss-120B} and \texttt{Qwen3.5-122B-A10B} achieve similar performance while exhibiting substantially different CoT lengths.}
    \label{fig:sft_resp_length}
\end{figure}

The SFT datasets are summarized in \Cref{tab:sft_data}. Specifically, \Cref{tab:sft_train_data} and \Cref{tab:sft_val_data} detail the training and validation sets, which consist of chain-of-thought (CoT) responses to their respective problems. \Cref{fig:sft_resp_length} illustrates the response length distributions, while \Cref{tab:sft_tokens} provides the total training token counts.

\begin{table}[H]
   
    \small
    \centering
    \caption{\textbf{SFT Dataset Statistics:} Numbers of CoT examples, problems solved, and training tokens.}
    \label{tab:sft_data}
    \begin{subtable}{\textwidth}
        \centering
        \caption{\textbf{SFT Training Dataset}}
        \label{tab:sft_train_data}
        \begin{tabular}{l c c c c | c c c}
        \toprule
        & \multicolumn{4}{c}{\textbf{Total CoTs}} & \multicolumn{3}{c}{\textbf{Problems Solved \& Success Rate}} \\
        \cmidrule(lr){2-5} \cmidrule(lr){6-8}
        \textbf{Model} & \textbf{Easy} & \textbf{Medium} & \textbf{Hard} & \textbf{Total} & \textbf{Easy (1,026)} & \textbf{Medium (1,012)} & \textbf{Hard (551)} \\
        \midrule
        gpt-oss-120B & 3,762 & 3,342 & 1,083 & \textbf{8,187} & 938 (91.4\%) & 884 (87.4\%) & 379 (68.8\%) \\
        Qwen3-30B-A3B & 3,761 & 3,037 & 245 & \textbf{7,043} & 900 (87.7\%) & 807 (79.8\%) & 113 (20.5\%) \\
        Qwen3.5-122B-A10B & 4,456 & 4,141 & 1,091 & \textbf{9,688} & 985 (96.0\%) & 956 (94.6\%) & 360 (65.3\%) \\
        \bottomrule
        \end{tabular}
    
    \end{subtable}
    
    \vspace{1em}
    
    \begin{subtable}{\textwidth}
        \centering
        \caption{\textbf{SFT Validation Dataset}}
        \label{tab:sft_val_data}
        
        \begin{tabular}{l c c c c | c c c}
        \toprule
        & \multicolumn{4}{c}{\textbf{Examples per Dataset}} & \multicolumn{3}{c}{\textbf{Problems Solved (out of 80 each)}} \\
        \cmidrule(lr){2-5} \cmidrule(lr){6-8}
        \textbf{Model} & \textbf{Easy} & \textbf{Medium} & \textbf{Hard} & \textbf{Total} & \textbf{Easy} & \textbf{Medium} & \textbf{Hard} \\
        \midrule
        gpt-oss-120B & 279 & 291 & 149 & \textbf{719} & 70 (87.5\%) & 73 (91.3\%) & 50 (62.5\%) \\
        Qwen3-30B-A3B & 267 & 240 & 28 & \textbf{535} & 67 (83.8\%) & 61 (76.3\%) & 11 (13.8\%) \\
        Qwen3.5-122B-A10B & 336 & 323 & 137 & \textbf{796} & 76 (95.0\%) & 73 (91.3\%) & 49 (61.3\%) \\
        \bottomrule
        \end{tabular}
    \end{subtable}
    
    \vspace{1em}
    
    \begin{subtable}{\textwidth}
        \centering
        \caption{\textbf{Total Training Tokens}}
        \label{tab:sft_tokens}
                
        \begin{tabular}{l c c c c}
        \toprule
        & \multicolumn{3}{c}{\textbf{Tokens by Level}} & \textbf{Overall} \\
        \cmidrule(lr){2-4} \cmidrule(lr){5-5}
        \textbf{Model} & \textbf{Easy} & \textbf{Medium} & \textbf{Hard} & \textbf{Total} \\
        \midrule
        gpt-oss-120B & 7.4M & 8.4M & 4.9M & 20.7M \\
        Qwen3-30B-A3B & 19.5M & 20.4M & 2.5M & 42.4M \\
        Qwen3.5-122B-A10B & 52.3M & 57.3M & 22.4M & 132.0M \\
        \midrule
        \textbf{All Models} & \textbf{79.2M} & \textbf{86.1M} & \textbf{29.8M} & \textbf{195.0M} \\
        \bottomrule
        \end{tabular}
    \end{subtable}

\end{table}

\section{Experiment 3: Narrow-Domain RL---Fermions and Spinors}
\label{sec:exp3}
As an additional experiment, we report our brief exploration of parameter-efficient fine-tuning (PEFT) using LoRA \cite{hu2021loralowrankadaptationlarge}.

Building on the broad fine-tuning studies in \Cref{sec:exp1} and \Cref{sec:exp2}, we investigate the limits of specializing a model on a single topic under a constrained data budget. Narrow-domain fine-tuning could in principle be of interest to researchers working on specific topics where the base model fails. As a concrete example, we focus on the graduate-level \emph{fermions and spinors} topic, where the base model exhibits comparatively weak performance.

For this experiment, we construct separate in-distribution (ID) and out-of-distribution (OOD) evaluation sets. The ID sets contain only fermion/spinor problems, whereas the OOD sets contain only non-fermion/spinor problems from the corresponding validation set. 
For Medium, we resample the 44 available fermion/spinor problems into 35 training problems and 9 held-out validation problems, with no overlap. For the Easy and Hard ID evaluations, we pool the fermion/spinor problems from their original training and validation sets; none of these problems is used in the Medium LoRA training run. Each OOD set contains only the non-fermion/spinor problems from the corresponding original validation set.

\subsection{Results}

\paragraph{Training Parameters}
We find full fine-tuning to be unstable with a training set this small, and instead use LoRA \cite{hu2021loralowrankadaptationlarge} fine-tuning (all linear layers, \(r = 64\), \(\alpha = 128\)) to reduce the number of trainable parameters and mitigate overfitting.
Even with fewer trainable parameters, RL training on LoRA models has been shown to yield performance gains on reasoning tasks \cite{wang2025tinatinyreasoningmodels,schulman2025lora}.
Following standard practice for LoRA fine-tuning, we use a learning rate of $10^{-5}$ ($10\times$ the full fine-tuning setting).
We keep the effective batch size at 128, implemented as a batch size of 8 with 16 rollouts. With this setup, one pass over the training set corresponds to roughly 5 RL steps. We evaluate every 5 RL steps and select the best-performing checkpoint for analysis. All other parameters remain consistent with our earlier RL configuration.

\begin{tcolorbox}[colback=blue!5!white, colframe=blue!50!black, title=\textbf{Key Takeaways: Targeted Specialization (Fermions \& Spinors)}, arc=4pt, boxrule=0.5pt, left=6pt, right=6pt, top=4pt, bottom=4pt]
\begin{itemize}
    \item LoRA fine-tuning on a small fermion/spinor dataset (35 problems) yields targeted improvements within the same domain.
    \item Mean performance on Easy problems increases from $14.5\% \to 25.5\%$ and on Medium problems from $8.9\% \to 20.0\%$, though we observe no gains on Hard problems.
    \item The model exhibits no systematic change in performance on out-of-distribution tasks.
\end{itemize}
\end{tcolorbox}
Performance peaks at step 40, after which signs of overfitting to the training set appear. On the small in-distribution (ID) Medium split (9 problems), fermion/spinor mean@5 improves from 8.9\% $\rightarrow$ 20.0\%, with pass@5 increasing from 22.2\% $\rightarrow$ 55.6\%.
For additional results, we evaluate on the combined set of training and validation fermion and spinor problems in both ID Easy and ID Hard---neither set is used for training. On ID Easy (65 problems), mean@5 increases from 14.5\% $\rightarrow$ 25.5\% with no change in pass@5. On ID Hard (27 problems), we find no change in performance.
On out-of-distribution (OOD) tasks, we find no systematic performance regression: on average, mean@5 changes only marginally on Easy (39.5\% $\rightarrow$ 39.2\%) and decreases slightly on Medium (27.6\% $\rightarrow$ 26.1\%). OOD pass@5 fluctuates (67.1\% $\rightarrow$ 65.8\% on Easy; 52.6\% $\rightarrow$ 46.1\% on Medium), which we interpret as statistical variation. On OOD Hard, mean@5 increases from 0.0\% to 0.5\%, while pass@5 increases from 0.0\% to 2.6\%.

In this experiment, we observed no systematic evidence of forgetting at the available sample size. We had initially anticipated that fine-tuning strictly on graduate-level fermions and spinors might degrade out-of-distribution performance or, conversely, induce positive transfer to related topics. Instead, our results indicate that in this case the model compartmentalizes the targeted learning: it improves on in-distribution tasks without compromising its pre-existing knowledge base, but fails to generalize these gains to broader physics domains.
\begin{table}[H]
    \centering
    \small
    \caption{\textbf{Narrow-domain fine-tuning results.} \texttt{DeepSeek-7B} performance on the fermion/spinor in-distribution (ID) vs.\ out-of-distribution (OOD) subsets across the Easy/Medium/Hard tiers.}
    
    \begin{subtable}[t]{1\linewidth}
        \centering
        \small
        \caption{\textbf{ID (fermion/spinor).}}
        \begin{tabular}{l cc cc cc}
            \toprule
            & \multicolumn{2}{c}{\textbf{Easy} (65)} & \multicolumn{2}{c}{\textbf{Medium} (9)} & \multicolumn{2}{c}{\textbf{Hard} (27)} \\
            \cmidrule(lr){2-3} \cmidrule(lr){4-5} \cmidrule(lr){6-7}
            \textbf{Model}
            & \textbf{mean@5} & \textbf{pass@5}
            & \textbf{mean@5} & \textbf{pass@5}
            & \textbf{mean@5} & \textbf{pass@5} \\
            \midrule
            \texttt{DeepSeek-7B}
            & 14.5 & 54.5
            & 8.9 & 22.2
            & 0.0 & 0.0 \\
            \texttt{DeepSeek-7B} + fermion/spinor RL
            & 25.5 & 54.5
            & 20.0 & 55.6
            & 0.0 & 0.0 \\
            $\Delta$ & \textcolor{green!60!black}{+11.0} & +0.0 & \textcolor{green!60!black}{+11.1} & \textcolor{green!60!black}{+33.4} & +0.0 & +0.0 \\
            \bottomrule
        \end{tabular}
    \end{subtable}
    
    \vspace{0.75em}
    
    \begin{subtable}[t]{1\linewidth}
        \centering
        \small
        \caption{\textbf{OOD (excluding fermion/spinor).}}
        \begin{tabular}{l cc cc cc}
            \toprule
            & \multicolumn{2}{c}{\textbf{Easy} (76)} & \multicolumn{2}{c}{\textbf{Medium} (75)} & \multicolumn{2}{c}{\textbf{Hard} (76)} \\
            \cmidrule(lr){2-3} \cmidrule(lr){4-5} \cmidrule(lr){6-7}
            \textbf{Model}
            & \textbf{mean@5} & \textbf{pass@5}
            & \textbf{mean@5} & \textbf{pass@5}
            & \textbf{mean@5} & \textbf{pass@5} \\
            \midrule
            \texttt{DeepSeek-7B}
            & 39.5 & 67.1
            & 27.6 & 52.6
            & 0.0 & 0.0 \\
            \texttt{DeepSeek-7B} + RL on fermion/spinor problems
            & 39.2  & 65.8
            & 26.1 & 46.1
            & 0.5 & 2.6     \\
            $\Delta$ & \textcolor{red!60!black}{-0.3} & \textcolor{red!60!black}{-1.3} & \textcolor{red!60!black}{-1.5}  & \textcolor{red!60!black}{-6.5}  & \textcolor{green!60!black}{+0.5}  & \textcolor{green!60!black}{+2.6}  \\
            \bottomrule
        \end{tabular}
    \end{subtable}
    
    \label{tab:fermion_synth_baseline}
\end{table}

\section{LLM-driven CoT Analysis Consistency Checks}
\normalsize
\label{app:consistency}
Chain-of-thought analysis is an inherently non-trivial task. Unlike human-produced answers, chains of thought are unnecessarily verbose, often circular, and frequently incoherent. The model jumps between ideas, leaving them underdeveloped and switching back and forth without apparent reason. Because of the large number of chains of thought (30 unique problems with 100 attempts per problem) and their length (up to \(20{,}000\) tokens) in \Cref{sec:error_analysis}, we performed the CoT analysis with \texttt{gpt-oss-120B}, which offers a compromise between expressivity and cost. We extensively validated the CoT analysis pipeline against a stronger model, \texttt{Claude-Sonnet-4.6} (via the Claude Code CLI).

We define three experimental configurations by crossing two Stage 2 (distillation) analyzers with two Stage 3 (error classification) analyzers:

\begin{center}
\begin{tabular}{c c c c}
\hline
\textbf{Configuration} & \textbf{Stage 2 Analyzer} & \textbf{Stage 3 Analyzer} & \textbf{Abbreviation} \\
\hline
\textbf{A} & Claude Sonnet & Claude Sonnet & Claude-on-Claude \\
\textbf{B} & gpt-oss-120B & Claude Sonnet & Claude-on-oss \\
\textbf{C} & gpt-oss-120B & gpt-oss-120B & oss-on-oss \\
\hline
\end{tabular}
\end{center}

Comparing \textbf{A vs.~B} isolates the effect of Stage 2 distillation on the final classification (same classifier, different inputs).
Comparing \textbf{B vs.~C} isolates the effect of the Stage 3 classifier (same inputs, different classifiers).
Comparing \textbf{A vs.~C} measures end-to-end agreement when both stages differ.
We study seven problems drawn from the most improved QFT Easy problems:

\begin{center}
\begin{tabular}{c l c}
\hline
\textbf{Problem} & \textbf{Topic} & \textbf{Incorrect Attempts (of 10)} \\
\hline
\textbf{p1064} & QCD running coupling with heavy quark thresholds (RGE) & 10 \\
\textbf{p16} & Chern--Simons level quantization (topological) & 3 \\
\textbf{p48} & Compton scattering Ward identity (QED) & 4 \\
\textbf{p129} & Massive vector field (Proca) propagator and polarizations & 7 \\
\textbf{p220} & Complex scalar field symmetry breaking and Goldstone theorem & 10 \\
\textbf{p413} & Proca field propagator sum over polarizations & 9 \\
\textbf{p722} & $\mathcal{N}=2$ SYM BPS mass spectrum and duality & 9 \\
\hline
\end{tabular}
\end{center}

All rollouts were generated by \texttt{DeepSeek-R1-Distill-Qwen-7B} (pre-RL). We analyze the first 10 attempts (indices 0--9) for each problem, yielding \textbf{52 incorrect attempts} in total across the seven problems. The golden solution decompositions (Stage~1) were produced by \texttt{gpt-oss-120B} and are shared across all configurations. Total error counts (rounded to the nearest percent) by category across all 52 rollouts are:

\begin{center}
\begin{tabular}{l c c c}
\hline
\textbf{Category} & \textbf{A (Claude-on-Claude)} & \textbf{B (Claude-on-oss)} & \textbf{C (oss-on-oss)} \\
\hline
mathematical & 70 (43\%) & 83 (50\%) & 80 (52\%) \\
logical & 41 (25\%) & 32 (19\%) & 26 (17\%) \\
executional & 35 (21\%) & 36 (22\%) & 23 (15\%) \\
factual & 17 (10\%) & 16 (10\%) & 26 (17\%) \\
\hline
\textbf{Total} & 163 & 167 & 155 \\
\hline
\end{tabular}
\end{center}

We consider this level of agreement satisfactory, and we note that the main source of disagreement is an inherent ambiguity in the error categories, which different models interpret differently. For example, a statement such as ``We have $y=A/x$, then $x=Ay$'' can be classified as either a ``mathematical error'' or a ``logical error''. Similarly, many attempts misstate the Proca propagator numerator structure $-g_{\mu\nu} + k_\mu k_\nu/m^2$, an error that sits at the intersection of factual recall, mathematical derivation, and logical reasoning about transversality constraints. Claude distributes these across the logical and factual categories, while \texttt{gpt-oss-120B} consistently labels them factual. While per-problem classification is noisy, the aggregated statistics are more reliable and offer valuable insight into the effect of RL on error reduction.

\end{document}